\documentclass[11pt]{article}

\usepackage{times}
\usepackage{latexsym}

\usepackage{ccg}
\usepackage{caption,subcaption}
\usepackage{amsmath}
\usepackage{amssymb}
\usepackage[hidelinks]{hyperref}

\usepackage{tikz}
\usetikzlibrary{positioning, backgrounds,tikzmark, calc}
\usepackage{tikz-dependency}
\usepackage[linguistics]{forest}

\newcommand{\tkzmn}[2]{\tikzmarknode{#1}{#2}}

\pgfdeclarelayer{background}
\pgfsetlayers{background, main}

\usepackage{linguex} %

\usepackage{graphicx}

\usepackage{appendix}

\usepackage{ntheorem}
\theoremseparator{:}

\usepackage{color}

\definecolor{ltgray}{rgb}{0.5,0.5,0.5}

\definecolor{brown}{rgb}{0.54,0.27,0.07}
\definecolor{orange}{rgb}{0.95,0.74,0.00}

\definecolor{darkgreen}{rgb}{0.10,0.99,0.10}
\definecolor{pink}{rgb}{1.0,0.75,0.79}

\usepackage[T1]{fontenc}

\usepackage{natbib}

\usepackage{bigfoot} %

\newcommand{\reveal}{\textsc{reveal}}
\newcommand{\chinchilla}{\textit{Chinchilla}}

\newcommand{\rmse}[1][\null]{$RMSE^{#1}$}
\newcommand{\deltarmse}[1][\null]{$\Delta RMSE^{#1}$}

\newcommand{\cibeta}{CI$^{\beta}_{99\%}$}
\newcommand{\cirmse}{CI$^{\Delta RMSE}_{99\%}$}

\newcommand{\rms}{\texttt{rms}}
\newcommand{\fzero}{\texttt{f0}}
\newcommand{\wordrate}{\texttt{wordrate}}
\newcommand{\freq}{\texttt{freq}}
\newcommand{\surprisal}{\texttt{chinchilla\_surprisal}}
\newcommand{\bottomup}{\texttt{cfg\_bottomup}}
\newcommand{\topdown}{\texttt{cfg\_topdown}}
\newcommand{\leftcorner}{\texttt{cfg\_leftcorner}}
\newcommand{\ccgleft}{\texttt{ccg\_left}}
\newcommand{\ccgreveal}{\texttt{ccg\_revealing}}

\title{
Modeling structure-building in the brain with CCG parsing and large language models 
}

\author{
Milo\v{s} Stanojevi\'{c}\footnote{co-first author} \\
\and
Jonathan R. Brennan\footnotemark[\value{footnote}]~\thanks{corresponding author: \url{jobrenn@umich.edu}} \\
\and 
Donald Dunagan \\
\and
Mark Steedman \\
\and
John T. Hale
}

\definecolor{ltblue}{rgb}{0.63, 0.79, 0.95}
\definecolor{ltgreen}{rgb}{0.67, 0.88, 0.69}

\begin{document}

\maketitle

\clearpage

\begin{abstract}

To model behavioral and neural correlates of language comprehension in naturalistic environments researchers have turned to broad-coverage tools from natural-language processing and machine learning. 
Where syntactic structure is explicitly modeled, prior work has relied predominantly on context-free grammars (CFG), yet such formalisms are not sufficiently expressive for human languages. 
Combinatory Categorial Grammars (CCGs) are sufficiently expressive directly compositional models of grammar with flexible constituency that affords incremental interpretation.
In this work we evaluate whether a more expressive CCG provides a better model than a CFG for human neural signals collected with fMRI while participants listen to an audiobook story.
We further test between variants of CCG that differ in how they handle optional adjuncts.
These evaluations are carried out against a baseline that includes estimates of next-word predictability from a Transformer neural network language model.
Such a comparison reveals unique contributions of CCG structure-building predominantly in the left posterior temporal lobe: CCG-derived measures offer a superior fit to neural signals compared to those derived from a CFG.
These effects are spatially distinct from bilateral superior temporal effects that are unique to predictability.
Neural effects for structure-building are thus separable from predictability during naturalistic listening, and those effects are best characterized by a grammar whose expressive power is motivated on independent linguistic grounds.

\end{abstract}

\clearpage

\section{Introduction} \label{sec:intro}

At the sentence~level, there remain many unanswered~questions regarding language~comprehension.
What algorithm best describes this cognitive~process?  The interactive and dynamic character of this sort of cognition \citep{Marslen-Wilson:1975fk,Tanenhaus:1995lq} has
 motivated a turn to the brain. 
 The hope is that neural~data with more granular spatial and/or temporal resolution can help to tease key~pieces apart.
This brain-informed strategy is promising, but faces a number of challenges related to scale.
One issue is how insights from experimental~designs that probe isolated~phrases \citep{Bemis:2011fk, zaccarella2017building, murphyMinimalPhraseComposition2022, Matchin:2018rw} and sentences \citep{Pallier:2011fk, Nelson:2017kq, Zaccarella:2017sy} might scale to more natural instances of language~processing \citep{Hasson:2015qq}.
A promising approach to this scale~problem deploys broad-coverage tools from natural-language processing (NLP) to operationalize cognitive~models for language understanding and to annotate more naturalistic instances of language. 
Statistical alignment between annotations and neural~signals is then used to localize specific cognitive~processes and to adjudicate between alternative models \citep[e.g.][]{COGS:COGS12445, Brennan:2016yu, Brennan:2020ku, Wehbe:2014la,reddyCanFMRIReveal2021,Shain:2020qq,Nelson:2017kq,Bhattasali:2018fk}. 
Such research has revealed a compelling alignment in the neural~bases of different~aspects of sentence~comprehension.
Over the past few years, results with both experimentally-controlled stimuli as well as more naturalistic~materials
have converged upon a view of posterior temporal regions \citep{Wilson:2011lq, murphyMinimalPhraseComposition2022, Zaccarella:2017sy, Brennan:2020ku}
as working in~tandem with anterior temporal and inferior~frontal areas to subserve the apperception of linguistic structure.
Of the latter, anterior~temporal brain~areas have been linked with semantic~combinatorics \citep[e.g.][]{liDisentanglingSemanticComposition2021} while inferior~frontral regions may subserve working memory operations \citep[e.g.][]{Amici:2007yu, Matchin:2014la}; 
\citet{matchinCorticalOrganizationSyntax2020} offer a integrated~framework for these findings.

The body of work cited above rests~on two key~assumptions that have yet to be put to the test.
The first assumption has to do with the~grammar: existing~studies have for the most~part modeled syntax using context-free grammars (CFG) that capture some constituency facts, but are not expressive enough  to capture the full~range of natural language dependencies \citep{Joshi:1985fk, Stabler:2013zp, Steedman:2000yq}.%
\footnote{Some important nuance to this generalization is discussed on page~\pageref{sec:prevwork}.}
The second assumption concerns the trade-off between complexity and predictability. 
It has long been recognized that language use might be modulated by
structural~complexity \citep{frazier1985syntactic, Hawkins:2004yf, Miller:1963ix}, perhaps giving rise to a ``bottleneck'' whereby processing~costs are primarily driven by usage-based factors like predictability which may, in turn, reflect factors including structural complexity \citetext{see \citealp{Levy:2008vh} but also \citealp{hawkins94,Hawkins:2004yf,Hawkins2014}}.
While previous~work has sought to tease these apart, the newest generation of transformer-based Large Language Models (LLMs) \cite[et seq.]{vaswaniAttentionAllYou2017} offers an unprecedented tool to capture
these usage-related factors.
Indeed, the apparent match between the outputs of these models and human neural signals suggest they offer a very strong baseline for isolating neural responses that reflect statistical patterns alone \citep{schrimpfNeuralArchitectureLanguage2021, kumarReconstructingCascadeLanguage2022, caucheteuxBrainsAlgorithmsPartially2022, 
caucheteuxEvidencePredictiveCoding2023,
heilbronHierarchyLinguisticPredictions2022}.

In this paper, we model structure-building using
a Combinatory Categorial Grammar (CCG) with human-like expressiveness  \citep{stanojevicCCGParsingAlgorithm2019, stanojevicMaxMarginIncrementalCCG2020}.
We quantify the number of parsing~steps incrementally, word-by-word, and use that quantity to model whole-brain fMRI time-series recorded while participants listen to an audiobook story.
This neural modeling effort includes state-of-the-art estimates of word-predictability from the \chinchilla{} LLM \citep{chinchilla:arxiv}.
With these materials, we ask three questions:
First, whether CCG parser~steps capture neural~variance better than steps derived from a CFG.
Second, whether such correlations hold above-and-beyond LLM-based estimates of predictability.
And, third, we compare alternative formulations of CCG~parsing to determine which one is most human-like.

\subsection{Incremental parsers as models for neural signals} \label{sec:prevwork}

\citet{Stabler:2013zp} highlights the ``hidden consensus'' among diverse groups of linguists %
such that world's languages surpass the limits of context-free grammar (CFG), but not by very much.
This consensus implicates the ``mildly context-sensitive'' formalisms characerized by \citet{Joshi:1985fk}.
One example of the kind of structures that require such greater expressivity are crossing serial dependencies found in numerous languages. 
Example \ref{ex:dutch} illustrates such a dependency with an expression from Dutch; here, the embedded clauses are ordered such that the paired subjects and verbs are inter-leaved, rather than being nested \citep[p. 25]{Steedman:2000yq}.

\exg. \label{ex:dutch}%
	\ldots omdat ik Cecilia Henk de nijlpaarden zag helpen voeren \\
	\ldots because \tkzmn{s1}{I} \tkzmn{s2}{Cecilia} \tkzmn{s3}{Henk} the hippopotamuses \tkzmn{v1}{saw\vphantom{p}} \tkzmn{v2}{help} \tkzmn{v3}{feed\vphantom{p}} \\
	\vskip 2em ``\ldots because I saw Cecilia help Henk feed the hippopotamuses'' \\
	\begin{tikzpicture}[remember picture, overlay, node distance = 1em]
		\draw[-, rounded corners=5pt, shorten >= 2pt] (v1.south) --  +(0,-.3)  -| (s1.south);
		\draw[-, rounded corners=5pt, shorten >= 2pt] (v2.south) --  +(0,-.5)  -| (s2.south);
		\draw[-, rounded corners=5pt, shorten >= 2pt] (v3.south) --  +(0,-.7) -| (s3.south);
	\end{tikzpicture}

Prior efforts to model naturalistic neural signals associated with structure building have not, on the whole, reflected this consensus. 
Structural complexity estimates based on less-expressive CFGs have been found to correlate with regions in the left temporal lobe  \citep{Brennan:2010ab,Nelson:2017kq, Brennan:2016yu,reddyCanFMRIReveal2021}. 
Attempts to extend this to more expressive grammars have been extremely limited.
\citet{Shain:2020qq} start from a Generalized Categorial Grammar, which may be quite expressive, but ultimately proceed to compile that down to a CFG. 
\citet{Brennan:2020ku} use a Recurrent Neural Network Grammar (RNNG; \citealp{N16-1024}) and estimate structural complexity via the number of parser transitions attempted across a parallel ``beam'' of partial analyses.
Yet, the phrase~structures parsed by this particular RNNG were comparatively na\"ive.
They reflected only the constituency~annotations in the Penn~Treebank~\citep{marcus93}, without explicitly treating long-distance dependencies such as filler-gap constructions and WH~questions \citep[see \S4 of][]{bies95:guidelines}.
Finally, \citet{Brennan:2016yu} estimate structure using a CFG as well as a more expressive Minimalist Grammar (MG); the latter captures neural variance robustly in posterior temporal regions and more modestly in anterior regions.
However, the MG deployed there was hand-built for the stimulus text; it was not broad-coverage in a way that would generalize to other instances of natural language.

Despite these limitations, prior studies paring structural annotations with naturalistic data have revealed left-temporal correlates for structural complexity that are broadly consistent with results
from more constrained experimental designs.
However, these studies show individual~differences in more detailed terms of the relative contribution of anterior versus posterior middle and superior temporal regions.
Such discrepancies could reflect the limitations of context-free grammar per~se,
differences in parsing~strategy,
or other analytical differences such~as the selection of regions of interest.  The present~effort aims to address each of these limits.

\begin{figure*}[ht!]
    \centering
      \subcaptionbox{\label{fig:comparison:constituency}}{
          \scalebox{0.7}{
              \begin{forest}
              [S,circle,draw,scale=1.2,
                  [NP,circle,draw,
                      [{Mary},draw,align=center,tier=words,name=Mary,]]
                  [VP,circle,draw,
                          [V,draw,circle,scale=1.1
                              [{reads},draw,align=center,tier=words,name=reads,]
                          ]
                          [NP,circle,draw,
                             [{papers},draw,align=center,tier=words,name=papers,]
                          ]
                  ]
              ]
              \end{forest}
          }
      }
    
      \subcaptionbox{\label{fig:comparison:ccg}}{
         \scalebox{0.7}{\deriv{3}{
           \rm Mary &\rm reads &\rm papers \\
           \uline{1}&\uline{1}&\uline{1}\\
           \it NP&\it (S{\bs}NP)/NP &\it NP\\
           \textcolor{red}{\textit{mary}'} & \textcolor{red}{\lambda{} x.\lambda y. \textit{reads}'(y, x)} & \textcolor{red}{\textit{papers}'} \\
            &\fapply{2} \\
            &\mc{2}{\it S{\bs}NP}\\
            & \mc{2}{\textcolor{red}{
                      \lambda{} y.\textit{reads}'(y,\ \textit{papers}')
                      }} \\
            \bapply{3}\\
            \mc{3}{\it S} \\
            \mc{3}{\textcolor{red}{
                      \textit{reads}'(\textit{mary}',\ \textit{papers}')
                      }} \\
                      & \\
                      & \\
                      & \\
                      & \\
                      & \\
         }}
      }
      \subcaptionbox{\label{fig:comparison:ccgB}}{
         \scalebox{0.7}{
			\deriv{3}{
			    \rm Mary &\rm reads &\rm papers \\
			    \uline{1}&\uline{1}&\uline{1}\\
			    \it NP&\it (S{\bs}NP)/NP &\it NP\\
			    \textcolor{red}{\textit{mary}'} & \textcolor{red}{\lambda{} x.\lambda y. \textit{reads}'(y, x)} & \textcolor{red}{\textit{papers}'} \\
			    \ftype{1} & &  \\
			              & & \  \\
			    \it S/(S{\bs}NP)&                &      \\
			    \textcolor{red}{\lambda{} p. p\ \textit{mary}'} & & \\
			    \fcomp{2} \\
			    \mc{2}{\it S/NP}\\
			    \mc{2}{\textcolor{red}{\lambda{} x. \textit{reads}'(\textit{mary}', x)}} & \\
			    \fapply{3}\\
			    \mc{3}{\it S} \\
			    \mc{3}{\textcolor{red}{
			              \textit{reads}'(\textit{mary}',\ \textit{papers}')
			              }} \\
			}}
      }
     \caption{Comparison of different syntactic representations for the sentence ``Mary reads papers''. 
     (A) shows a phrase-structure constituency tree that follows a context-free grammar. 
     (B--C) Left and right-branching  combinatoric categorical grammar derivations; both yield the same final semantic interpretation yet differ in the order by which phrases are composed. }
     \label{fig:comparison}
\end{figure*}
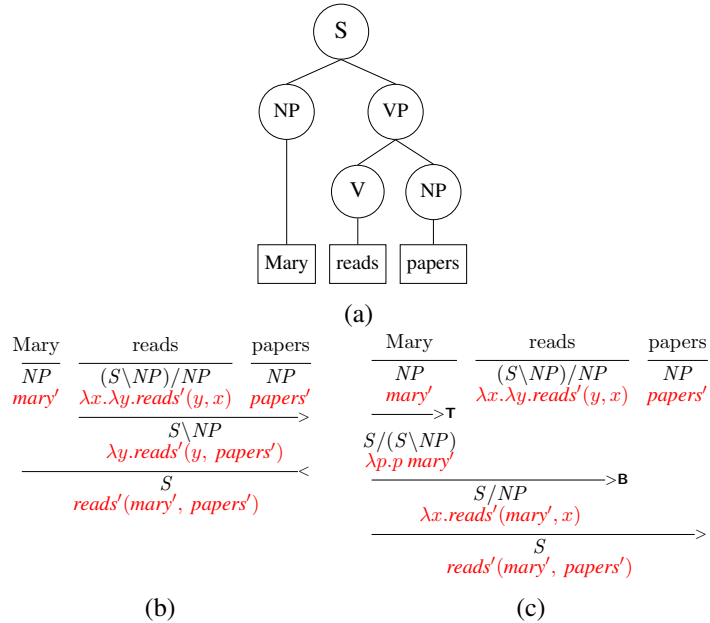

\subsubsection*{Combinatory Categorial Grammar} \label{sec:ccg}

Combinatory Categorial Grammar (CCG) is a mildly context-sensitive formalism
that
fits into realistic processing~models \citep{Steedman:2000yq}. 
Its flexible~constituency allows for a very high degree of incrementality even with
simple analysis methods like shift-reduce parsing.
In addition, it is directly~compositional in the sense of \citet{barker:volume};
each syntactic~constituent is assigned a corresponding semantic~interpretation in the form of a lambda~term that has explicit truth~conditions.
Together, these properties of the grammar offer a good~match to human sentence processing, where comprehension operates immediately and incrementally.%
\footnote{Human sentence processing
can proceed in a way that is closely time-locked to the incoming stream of words \citep{Marslen-Wilson:1975fk,Altmann:1988pd,Tanenhaus:1995lq}.
\citet{sagwasow:pccg} distill this and related considerations into a short list of requirements on performance-plausible grammars.
In subsequent years, Stabler's \citeyearpar{stabler97} formalization emerged as a version of transformational~grammar that meets Sag and Wasow's requirements \citep[see e.g.][]{hale06,GrafEtAl17JLM,stanojevic:stabler:2018:cogacll,hunter-etal-2019-active,chen21}.}
Thanks to the treebanking efforts of Julia~Hockenmaier, CCG is broad-coverage in exactly the manner required
for matching naturalistic experimental stimuli \citep{hockenmaierCCGbankCorpusCCG2007}.
Panels (b) and (c) of Figure \ref{fig:comparison}
present two different CCG analyses that provide the same semantic interpretation.
These contrast with the na\"ive phrase structure in panel (a) that typifies context-free approaches to sentence structure. %

As suggested above, these existence of two alternative analyses grant a CCG-based parser the power to operate incrementally, word-by-word.
These alternatives differ in terms of the \textit{eagerness} of a corresponding parsing~process:
    the ``left-branching'' derivation  \ref{fig:comparison:ccgB} eagerly composes structure as soon as it is grammatically possible.
    The ``right-branching'' derivation \ref{fig:comparison:ccg} is less eager; there are points in the derivation where multiple words must be recognized before a composition~step can be taken. 
    For instance the Subject ``Mary'' is not composed with its main~verb ``reads'' until the third and final word. Both analyses are perfectly well-formed and are associated with exactly the same semantic~interpretation as shown in red.

Although context-free grammar lacks flexible constituency, there exists a universe of alternative parsing~strategies for these grammars that likewise can be viewed as varying in eagerness.
Top-down parsing is maximally eager, whereas bottom-up parsing is minimally eager. 
This generalized perspective is elaborated in the Methods section, which includes Figures \ref{fig:cfg:top:down}--\ref{fig:cfg:left:corner} illustrating all of these strategies.
\citet[][chapter 3]{Hale:2012ve} offers a pedagogical treatment. 
Crucially, to model naturalistic behavioral and neural signals from language-users, one must commit both to the grammatical structures that the language-user is using, and also to the strategy (more or less eager) by which structures are composed \citep{brennanNaturalisticSentenceComprehension2016}.

A traditional, if Anglo-centric, view is that human sentence~processing operates according to the ``left-corner'' strategy \citep{johnson-laird83,Abney:1991rp}.
This traditional~view is supported by patterns of expected memory-use during comprehension~\citep{resnik-1992-left}.
Indeed, \citet{COGS:COGS12445} present evidence that neural~signals from the anterior temporal lobe recorded with magnetoencephalography during reading are consistent with the left-corner, not bottom-up, strategy.
\citet{Nelson:2017kq}, on the other hand, report electrophysiological~data recorded from intra-cranial recordings in frontal and temporal regions that are most consistent with either a bottom-up or left-corner strategy but not, in their analysis, with an eager top-down strategy.
Among their differences, this latter neurolinguistic study involves isolated sentences, while the former uses more natural story-book reading; this highlights the potential tension between different literatures mentioned above.

Bottom-up parsing is thus a relatively a simple strategy which enjoys some empirical support.
It can be applied together with CCG or na\"ive phrase~structure.
Regardless of which grammar is used, bottom-up parsing always faces a prima~facie problem
capturing the incremental interpretation that humans appear to show during real-world comprehension, especially regarding \textit{right adjunction}.
The issue can be illustrated with the simple sentence ``Mary reads papers daily.'' %
From an incremental perspective, a bottom-up analysis would yield a complete sentence after just the words ``Mary reads papers'' -- that is indeed a grammatically acceptable string.
When confronted with the modifier ``daily'', the parser would need an extra sequence of processing steps to reject the existing analysis and re-compose the modifier ``daily'' with the verb phrase ``reads papers''.
The extra processing steps seems to be at odds with human processing evidence which shows little to no difficulties with right-adjunction of this sort.
\citet[ch. 2]{Lewis:1993aa} reviews a host of what he terms ``unproblematic ambiguities'' in this vein which together place constraints on the kind of flexibility needed to account for human sentence processing patterns.

\begin{figure}
\centering
\resizebox{\textwidth}{!}{
    \begin{forest}
    for tree={
        if n children=0{
            tier=terminal
        }{},
    }
    [S 
        [S/VP, edge={blue, very thick}
            [NP, tikz={\node [fill=blue, opacity=0.1, rounded corners, inner sep=0, blue, very thick, fit to=tree]{};}, edge={blue, very thick} 
                [the pilot, roof] 
            ]
        ]
        [VP, edge={blue, very thick, dashed}
            [VP 
                [embarrassed Mary, roof] 
            ]
            [CONJ 
                [and] 
            ]
            [VP, edge={blue, very thick, dashed} 
                [VP/PP, edge={blue, very thick, dashed} 
                    [VP/PP/NN [put] ]
                    [NP, tikz={\node [fill=blue, opacity=0.1, rounded corners, inner sep=0, blue, fit to=tree]{};}, edge={blue, very thick} 
                        [herself ] 
                    ]
                [PP, edge={dashed, thick}
                    [in a very awkward \colorbox{red!20}{situation}, roof ] 
                ]
                ]
            ]
        ]
    ]
    \end{forest}
}
    \caption{Fully bottom-up parsers face limits as models of incremental comprehension in humans. 
    In this example, co-reference between \textit{the pilot} and \textit{herself} requires a fully connected path between the two noun-phrases (blue), yet such a path is not complete under bottom-up strategies until the word \textit{situation} (red) is parsed. 
    Dashed lines indicate the path that remains un-connected until the final word.
    See \citet{stanojevicMaxMarginIncrementalCCG2020} for more discussion. (Figure adapted from \citealp{sturt2005processing}.)}
    \label{fig:sturt}
\end{figure}
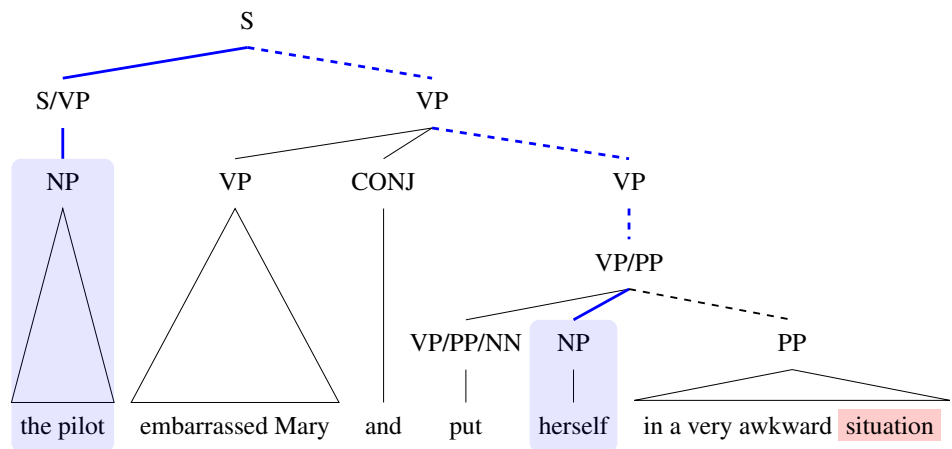

Experimental evidence sharpens the challenge for bottom-up analyses and CCG.
\citet{sturt2005processing} examine patterns of eye-fixations while participants read sentences combining reflexive anaphora and conjunction, such as ``The pilot embarrassed Mary and put herself an a very awkward situation.'' 
Here, co-reference between ``the pilot'' and ``herself'' requires a connected path (``c-command within a local domain'' in the terminology of \citealp{Cho81}).
Yet, bottom-up strategies do not make such a path available until the second coordinated phrase has been fully parsed, as schematically illustrated in Figure~\ref{fig:sturt}.
Despite this lack of syntactic connection, eye-fixation data reveals that comprehenders
in fact do resolve the relevant co-reference relationship immediately, without waiting for the clause-final word ``situation.''
Evidence for immediate interpretation of co-reference in sentences such as this is incompatible with bottom-up parsing strategies discussed thus far.

\citet{stanojevicCCGParsingAlgorithm2019,stanojevicMaxMarginIncrementalCCG2020} present a CCG parser with components designed to address this challenge.%
\footnote{More precisely, \citeauthor{sturt:and:lombardo}'s challenge is addressed by combining revealing parsing operation with a predictive step. For more details about this construction see \citet{CUNY2020}.}
They do so by implementing an incremental tree-rotation algorithm via a {\reveal} operation added to the parser along-side \textsc{shift} and \textsc{reduce}.
Under this revised strategy, the parser incrementally converts left-branching to right-branching structures that are suitable for composing with the modifier.

With this backdrop, we aim to test which of a family of broad-coverage syntactic parsers offers the best fit to human neural signals.  
We compare models that differ in the grammar used to build structures (less expressive CFG or human-like CCG), as well as the parsing~strategy that is applied incrementally. 
For CCG, we test between a model that includes the {\reveal} operation, and one which does not. For CFG, we evaluate both a top-down and a bottom-up strategy (as discussed in the Methods section below, the top-down and left-corner strategies are not distinguishable with the fMRI data used in this study.)

Comparing models along these dimensions alone, however, is not sufficient.
Indeed, language processing is highly sensitive to the predictability and frequency of expressions; we turn next to psycholinguistic and computational models of predictability that we might probe in comparison to parsing models discussed thus far.

\subsection{Predictability, processing, and large language models}

Language users are exquisitely sensitive to the statistical properties of their linguistic experience \citep[e.g.][]{bybeeUsageGrammarMind2006}.
Indeed, \citet{Levy:2008vh} proposes that linguistic expectations might serve as a ``causal~bottleneck between the linguistic~representations constructed during sentence~comprehension and the processing~difficulty incurred at a given~word within a sentence.''~(page 1128).
The leading~idea is that language~processing is modulated, perhaps in large~part, by expectations grounded in usage and, in~turn, usage is affected by linguistic -- including syntactic -- complexity. 
This idea builds on a rich tradition of research at the intersection of psycholinguistics, typology, and linguistic theory, including observations that language processing is facilitated when input is syntactically predictable \citep[e.g][]{hale2001pep,Hale:2010aa},
    studies showing preferences to disambiguate towards structurally simpler expressions \citep{frazier1985syntactic}, 
    and evidence that such preferences shape the typological distribution of languages as they exert performance pressures as languages develop and change (\citealp{Hawkins:2004yf}, see also \citealp{Futrell:2015fp}).
The well-documented link between structural complexity, usage, and linguistic expectations demands careful attention in any effort to tease out neural signatures of structure-processing.

In NLP, neural-network language models, especially LLMs based on the transformer architecture of \citet{vaswaniAttentionAllYou2017}, have shown
tremendous capacity to capture and reproduce certain statistical characteristics of huge amounts of text.
These models are usage-based in the sense that their architecture is not tuned to the structural properties of human language, and their initial state is random; the linguistic patterns that they induce are due solely to input they receive.
The inputs to the models are strings of text and they are trained via gradient descent to predict a masked word given some amount of surrounding  (``bidirectional'') or just preceding (``unidirectional'') context.
Left to right unidirectional, or ``causal'' language~models operate incrementally, word by word, in the same~order as human language processing.

Next-word prediction, the task optimized by LLMs, seems at odds with the comparatively rich task of language comprehension. 
Yet these models have offered an unprecedented window into the rich information latent in the statistical patterns of language use, showing evidence of inducing aspects of syntactic structure \citep{manningEmergentLinguisticStructure2020} with some notable limitations \citep{Ettinger:2020jy}; see \citet{linzenSyntacticStructureDeep2021} for a review.
Such structure may come to bear on behavioral and neural correlates of processing via the ``causal bottleneck'' mentioned above.
Indeed, probing for evidence of structure through the lens of how it might constrain expectations has been precisely the focus of prior work examining behavioral and neural indices of syntactic structure in naturalistic settings \citep{Frank:2011qf,Frank:2015kq,Brennan:2017db,Shain:2020qq,Henderson:2016yq}.
Those efforts complement the path we take here to probe structural complexity more directly and independently of any mediation between structure and predictability.

Given that both LLMs and humans are highly sensitive to patterns of language use, it is perhaps not surprising that LLMs have been found to reliably correlate with a variety of human neural signals.
\citet{heilbronHierarchyLinguisticPredictions2022} evaluates the fit between next-word predictions furnished by a LLM and neural signals in a variety of ways \citep[see also][]{caucheteuxEvidencePredictiveCoding2023}.
In that work, model outputs are quantified in terms of \textit{surprisal}, which is a transformation of conditional probability. 
These quantities, as defined by a language~model, can be binned into syntactic~categories or classes of defined by shared phonological~form(s).
Previous work has reported robust~correlations between model-derived prediction at these various~levels and distinct neural~signals recorded with both electroencephalography (EEG) and magnetoencephalography (MEG).

Other groups have focused on testing for alignment between the internal states of LLMs and human neural signals.
\citet{schrimpfNeuralArchitectureLanguage2021} evaluate the fit between the neural network activations of a range of language models, including transformer-based LLMs, and human neural signals recorded with fMRI as well as electrocorticography (ECoG).
They find statistically reliable fits between patterns of activation in the models and human neural signals recorded while participants comprehend sentences and more natural stories. 
Moreover, they find the next-word prediction performance of the models is the single best predictor of the degree of match between model activations and neural signals.
\citet{caucheteuxBrainsAlgorithmsPartially2022} also find that LLMs with better missing-word prediction performance on text show a better statistical match with fMRI and MEG neural signals recorded while participants read isolated sentences.
These effects span broad swatchs of left frontal and temporal cortices. 
Probing the nature of these fits further, they find the best model-to-brain match obtains with the middle layers of the LLMs (e.g. layers 8--9 of a 12-layer feed-forward transformer network) rather than input or output layers. 
\citet{kumarReconstructingCascadeLanguage2022} report similar results for fMRI data collected while participants listen to narratives.

A persistent challenge for the above-mentioned work is that the LLMs whose activation patterns show a high degree of match to brain signals are themselves ``black box'' models whose internal states are not directly interpretable.
\citeauthor{kumarReconstructingCascadeLanguage2022} propose an interesting strategy to confront this limitation by focusing specifically on how different ``attention heads'' in the transformer network, which serve in this architecture to mediate the spread of feed-forward activation as a function of context, drive neural performance in different brain regions.
Analyzing the distribution of activation across attention heads is one of several strategies that have been pursued with some success in rendering LLMs interpretable in terms of linguistic structure \citep[e.g.][and others]{DBLP:journals/corr/KuncoroBKDNS16, voita-etal-2019-analyzing,manningEmergentLinguisticStructure2020, yuAssessingPhrasalRepresentation2020}.

We take a different, parallel, approach in this project by beginning with parsing models whose internal states are directly interpretable in terms of linguistic representations and computations.
Along-side measures from those models, we add estimates of next-word predictability from a state-of-the-art LLM in order to tease apart structural processing from usage-based expectations. 
To this end, we use the \chinchilla{} LLM introduced by \citet{chinchilla:arxiv}.
This particular model was developed with a focus on finding the optimal tradeoff between model size and training data.
It is a top-performer of the current generation of transformer-based language models on a range of tasks, including next-word prediction, outperforming DeepMind's \textit{Gopher} \citep{raeScalingLanguageModels2021}, OpenAI's \textit{GPT-3} \citep{brownLanguageModelsAre2020}, and other current state-of-the-art LLMs.

In sum, we test for alignment between neural signals and processing steps from a parser with human-level expressive~power (CCG) along-side the strongest-to-date baseline model for quantifying next-word expectations. 
To preview our results: we find that the CCG parser correlates with left-localized fMRI signals above-and-beyond simpler contex-free estimates of parsing, these correlations are made stronger when CCG is augmented with the {\reveal} operation that better handles right-adjunction.
These effects for structure-building localized to left posterior temporal cortices (and elsewhere) are linearly independent of neural effects for next-word predictability, which separately show high correlations with neural activity along the left superior temporal gyrus.

\section{Methods} \label{sec:methods}

\subsection{Computational modeling} \label{sec:computationalmodeling}

Our target predictors come from three types of models: 
    (1) constituency tree parser steps, 
    (2) CCG parser steps and 
    (3) Chinchilla large-language model surprisals. We describe each of these in the following sections.

\subsubsection*{Predictors from Constituency Parsers}

Most constituency parsers fall into three different categories that differ by how much they speculate about the future words that have not been observed yet.%
\footnote{The theory of Generalized Left Corner Parsing makes even more fine-grained distinctions between strategies; see \citet[][chapter 3]{Hale:2012ve} for an introduction.}
A bottom-up parser, sometimes also called Shift-Reduce parser, is the least speculative and forms tree nodes only over the words that have been observed. Figure~\ref{fig:cfg:bottom:up} shows an example parsing trace of this strategy for the sentence ``Mary reads papers daily''. 
Because it speculates very little, this strategy is not very incremental. 
For instance, words \textit{Mary} and \textit{reads} not get connected until the last step when all words ``\textit{Mary reads papers daily}'' are observed. 
This clearly does not fit with the incremental nature of human sentence processing.

\newcommand{\scalingBU}[1]{\scalebox{0.5}{#1}}

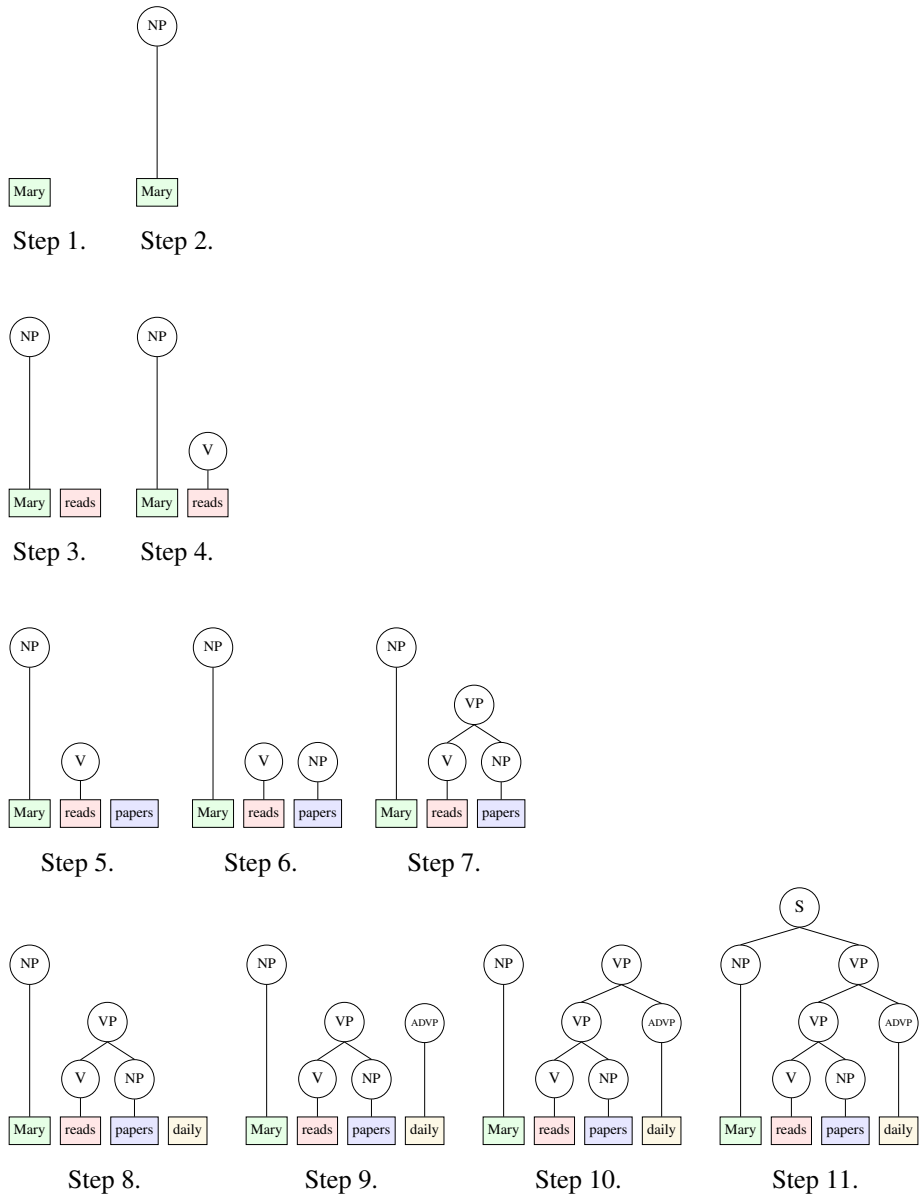
\begin{figure*}
    \captionsetup[subfigure]{labelformat=empty}
    \subcaptionbox{Step 1.\label{cfg:bottom:up:step:one}}{
          \scalingBU{
              \begin{forest}
              [S,phantom,circle,draw,scale=1.2,tikz={\node [gray, fit=()(Mary)(daily)]{};}
                  [NP,circle,draw,phantom
                      [{Mary},draw,fill=green,fill opacity=0.1,align=center,tier=words,name=Mary]]
                  [VP,circle,draw,phantom
                      [VP,circle,draw,phantom
                          [V,draw,circle,scale=1.1,phantom
                              [{reads},draw,fill=red,fill opacity=0.1,align=center,tier=words,name=reads,phantom]
                          ]
                          [NP,circle,draw,phantom
                             [{papers},draw,fill=blue,fill opacity=0.1,align=center,tier=words,name=NPtwo,phantom]
                          ]
                      ]
                      [ADVP,circle,draw,scale=0.7,phantom
                         [{daily},draw,fill=orange,fill opacity=0.1,align=center,tier=words,name=daily,phantom]
                      ]
                  ]
              ]
              \end{forest}
          }
      }
      \subcaptionbox{Step 2.\label{cfg:bottom:up:step:two}}{
          \scalingBU{
              \begin{forest}
              [S,phantom,circle,draw,scale=1.2,tikz={\node [red, fit=()(Mary)(daily)]{};}
                  [NP,circle,draw
                      [{Mary},draw,fill=green,fill opacity=0.1,align=center,tier=words,name=Mary]]
                  [VP,circle,draw,phantom
                      [VP,circle,draw,phantom
                          [V,draw,circle,scale=1.1,phantom
                              [{reads},draw,fill=red,fill opacity=0.1,align=center,tier=words,name=reads,phantom]
                          ]
                          [NP,circle,draw,phantom
                             [{papers},draw,fill=blue,fill opacity=0.1,align=center,tier=words,name=NPtwo,phantom]
                          ]
                      ]
                      [ADVP,circle,draw,scale=0.7,phantom
                         [{daily},draw,fill=orange,fill opacity=0.1,align=center,tier=words,name=daily,phantom]
                      ]
                  ]
              ]
              \end{forest}
          }
      }

      \subcaptionbox{Step 3.\label{cfg:bottom:up:step:three}}{
          \scalingBU{
              \begin{forest}
              [S,phantom,circle,draw,scale=1.2,tikz={\node [gray, fit=()(Mary)(daily)]{};}
                  [NP,circle,draw
                      [{Mary},draw,fill=green,fill opacity=0.1,align=center,tier=words,name=Mary]]
                  [VP,circle,draw,phantom
                      [VP,circle,draw,phantom
                          [V,draw,circle,scale=1.1,phantom
                              [{reads},draw,fill=red,fill opacity=0.1,align=center,tier=words,name=reads]
                          ]
                          [NP,circle,draw,phantom
                             [{papers},draw,fill=blue,fill opacity=0.1,align=center,tier=words,name=NPtwo,phantom]
                          ]
                      ]
                      [ADVP,circle,draw,scale=0.7,phantom
                         [{daily},draw,fill=orange,fill opacity=0.1,align=center,tier=words,name=daily,phantom]
                      ]
                  ]
              ]
              \end{forest}
          }
      }
      \subcaptionbox{Step 4.\label{cfg:bottom:up:step:four}}{
          \scalingBU{
              \begin{forest}
              [S,phantom,circle,draw,scale=1.2,tikz={\node [blue, fit=()(Mary)(daily)]{};}
                  [NP,circle,draw
                      [{Mary},draw,fill=green,fill opacity=0.1,align=center,tier=words,name=Mary]]
                  [VP,circle,draw,phantom
                      [VP,circle,draw,phantom
                          [V,draw,circle,scale=1.1
                              [{reads},draw,fill=red,fill opacity=0.1,align=center,tier=words,name=reads]
                          ]
                          [NP,circle,draw,phantom
                             [{papers},draw,fill=blue,fill opacity=0.1,align=center,tier=words,name=NPtwo,phantom]
                          ]
                      ]
                      [ADVP,circle,draw,scale=0.7,phantom
                         [{daily},draw,fill=orange,fill opacity=0.1,align=center,tier=words,name=daily,phantom]
                      ]
                  ]
              ]
              \end{forest}
          }
      }

      \subcaptionbox{Step 5.\label{cfg:bottom:up:step:five}}{
          \scalingBU{
              \begin{forest}
              [S,phantom,circle,draw,scale=1.2,tikz={\node [gray, fit=()(Mary)(daily)]{};}
                  [NP,circle,draw
                      [{Mary},draw,fill=green,fill opacity=0.1,align=center,tier=words,name=Mary]]
                  [VP,circle,draw,phantom
                      [VP,circle,draw,phantom
                          [V,draw,circle,scale=1.1
                              [{reads},draw,fill=red,fill opacity=0.1,align=center,tier=words,name=reads]
                          ]
                          [NP,circle,draw,phantom
                             [{papers},draw,fill=blue,fill opacity=0.1,align=center,tier=words,name=NPtwo]
                          ]
                      ]
                      [ADVP,circle,draw,scale=0.7,phantom
                         [{daily},draw,fill=orange,fill opacity=0.1,align=center,tier=words,name=daily,phantom]
                      ]
                  ]
              ]
              \end{forest}
          }
      }
      \subcaptionbox{Step 6.\label{cfg:bottom:up:step:six}}{
          \scalingBU{
              \begin{forest}
              [S,phantom,circle,draw,scale=1.2,tikz={\node [green, fit=()(Mary)(daily)]{};}
                  [NP,circle,draw
                      [{Mary},draw,fill=green,fill opacity=0.1,align=center,tier=words,name=Mary]]
                  [VP,circle,draw,phantom
                      [VP,circle,draw,phantom
                          [V,draw,circle,scale=1.1
                              [{reads},draw,fill=red,fill opacity=0.1,align=center,tier=words,name=reads]
                          ]
                          [NP,circle,draw
                             [{papers},draw,fill=blue,fill opacity=0.1,align=center,tier=words,name=NPtwo]
                          ]
                      ]
                      [ADVP,circle,draw,scale=0.7,phantom
                         [{daily},draw,fill=orange,fill opacity=0.1,align=center,tier=words,name=daily,phantom]
                      ]
                  ]
              ]
              \end{forest}
          }
      }
      \subcaptionbox{Step 7.\label{cfg:bottom:up:step:seven}}{
          \scalingBU{
              \begin{forest}
              [S,phantom,circle,draw,scale=1.2,tikz={\node [green, fit=()(Mary)(daily)]{};}
                  [NP,circle,draw
                      [{Mary},draw,fill=green,fill opacity=0.1,align=center,tier=words,name=Mary]]
                  [VP,circle,draw,phantom
                      [VP,circle,draw
                          [V,draw,circle,scale=1.1
                              [{reads},draw,fill=red,fill opacity=0.1,align=center,tier=words,name=reads]
                          ]
                          [NP,circle,draw
                             [{papers},draw,fill=blue,fill opacity=0.1,align=center,tier=words,name=NPtwo]
                          ]
                      ]
                      [ADVP,circle,draw,scale=0.7,phantom
                         [{daily},draw,fill=orange,fill opacity=0.1,align=center,tier=words,name=daily,phantom]
                      ]
                  ]
              ]
              \end{forest}
          }
      }

      \subcaptionbox{Step 8.\label{cfg:bottom:up:step:eight}}{
          \scalingBU{
              \begin{forest}
              [S,phantom,circle,draw,scale=1.2,tikz={\node [gray, fit=()(Mary)(daily)]{};}
                  [NP,circle,draw
                      [{Mary},draw,fill=green,fill opacity=0.1,align=center,tier=words,name=Mary]]
                  [VP,circle,draw,phantom
                      [VP,circle,draw
                          [V,draw,circle,scale=1.1
                              [{reads},draw,fill=red,fill opacity=0.1,align=center,tier=words,name=reads]
                          ]
                          [NP,circle,draw
                             [{papers},draw,fill=blue,fill opacity=0.1,align=center,tier=words,name=NPtwo]
                          ]
                      ]
                      [ADVP,circle,draw,scale=0.7,phantom
                         [{daily},draw,fill=orange,fill opacity=0.1,align=center,tier=words,name=daily]
                      ]
                  ]
              ]
              \end{forest}
          }
      }
      \subcaptionbox{Step 9.\label{cfg:bottom:up:step:nine}}{
          \scalingBU{
              \begin{forest}
              [S,phantom,circle,draw,scale=1.2,tikz={\node [purple, fit=()(Mary)(daily)]{};}
                  [NP,circle,draw
                      [{Mary},draw,fill=green,fill opacity=0.1,align=center,tier=words,name=Mary]]
                  [VP,circle,draw,phantom
                      [VP,circle,draw
                          [V,draw,circle,scale=1.1
                              [{reads},draw,fill=red,fill opacity=0.1,align=center,tier=words,name=reads]
                          ]
                          [NP,circle,draw
                             [{papers},draw,fill=blue,fill opacity=0.1,align=center,tier=words,name=NPtwo]
                          ]
                      ]
                      [ADVP,circle,draw,scale=0.7
                         [{daily},draw,fill=orange,fill opacity=0.1,align=center,tier=words,name=daily]
                      ]
                  ]
              ]
              \end{forest}
          }
      }
      \subcaptionbox{Step 10.\label{cfg:bottom:up:step:ten}}{
          \scalingBU{
              \begin{forest}
              [S,phantom,circle,draw,scale=1.2,tikz={\node [purple, fit=()(Mary)(daily)]{};}
                  [NP,circle,draw
                      [{Mary},draw,fill=green,fill opacity=0.1,align=center,tier=words,name=Mary]]
                  [VP,circle,draw
                      [VP,circle,draw
                          [V,draw,circle,scale=1.1
                              [{reads},draw,fill=red,fill opacity=0.1,align=center,tier=words,name=reads]
                          ]
                          [NP,circle,draw
                             [{papers},draw,fill=blue,fill opacity=0.1,align=center,tier=words,name=NPtwo]
                          ]
                      ]
                      [ADVP,circle,draw,scale=0.7
                         [{daily},draw,fill=orange,fill opacity=0.1,align=center,tier=words,name=daily]
                      ]
                  ]
              ]
              \end{forest}
          }
      }
      \subcaptionbox{Step 11.\label{cfg:bottom:up:step:eleven}}{
          \scalingBU{
              \begin{forest}
              [S,circle,draw,scale=1.2,tikz={\node [purple, fit=()(Mary)(daily)]{};}
                  [NP,circle,draw
                      [{Mary},draw,fill=green,fill opacity=0.1,align=center,tier=words,name=Mary]]
                  [VP,circle,draw
                      [VP,circle,draw
                          [V,draw,circle,scale=1.1
                              [{reads},draw,fill=red,fill opacity=0.1,align=center,tier=words,name=reads]
                          ]
                          [NP,circle,draw
                             [{papers},draw,fill=blue,fill opacity=0.1,align=center,tier=words,name=NPtwo]
                          ]
                      ]
                      [ADVP,circle,draw,scale=0.7
                         [{daily},draw,fill=orange,fill opacity=0.1,align=center,tier=words,name=daily]
                      ]
                  ]
              ]
              \end{forest}
          }
      }
      \caption{Bottom-Up parsing strategy for Context-Free Grammars.
      Each row indicates parser steps associated with a single word.}
      \label{fig:cfg:bottom:up}
\end{figure*}

In the other extreme is a top-down parser that is maximally speculative. 
Figure~\ref{fig:cfg:top:down} shows the trace of top-down parser for the same example sentence. 
Here the parser speculates the whole path from the root node \textit{S} to the observed words, even though this path may not be consistent with future words that are not yet observed. 
This parser is as incremental as possible: all words are connected as soon as they are observed. 
For instance, the top-down parser can establish that \textit{Mary} is a subject of \textit{reads} as soon as it observes ``\textit{Mary reads}''.

\newcommand{\scalingTD}[1]{\scalebox{0.5}{#1}}

\begin{figure*}
    \captionsetup[subfigure]{labelformat=empty}
    \subcaptionbox{Step 1.\label{cfg:top:down:step:one}}{
          \scalingTD{
              \begin{forest}
              [S,phantom,circle,draw,scale=1.2,tikz={\node [rounded corners, fit=()(Mary)(daily)]{};}
                  [NP,circle,draw,phantom,
                      [{Mary},draw,fill=green,fill opacity=0.1,align=center,tier=words,name=Mary]]
                  [VP,circle,draw,phantom
                      [VP,circle,draw,phantom
                          [V,draw,circle,scale=1.1,phantom
                              [{reads},draw,fill=red,fill opacity=0.1,align=center,tier=words,name=reads,phantom]
                          ]
                          [NP,circle,draw,phantom
                             [{papers},draw,fill=blue,fill opacity=0.1,align=center,tier=words,name=NPtwo,phantom]
                          ]
                      ]
                      [ADVP,circle,draw,scale=0.7,phantom
                         [{daily},draw,fill=orange,fill opacity=0.1,align=center,tier=words,name=daily,phantom]
                      ]
                  ]
              ]
              \end{forest}
          }
      }
      \subcaptionbox{Step 2.\label{cfg:top:down:step:two}}{
          \scalingTD{
              \begin{forest}
              [S,circle, draw,scale=1.2,tikz={\node [rounded corners, fit=()(Mary)(VPone)]{};}
                  [\phantom{NP},circle,draw,
                      [{Mary},draw,fill=green,fill opacity=0.1,align=center,tier=words,name=Mary, no edge]]
                  [\phantom{VP},circle,draw,name=VPone
                      [VP,circle,draw,phantom
                          [V,draw,circle,scale=1.1,phantom
                              [{reads},draw,fill=red,fill opacity=0.1,align=center,tier=words,name=reads,phantom]
                          ]
                          [NP,circle,draw,phantom
                             [{papers},draw,fill=blue,fill opacity=0.1,align=center,tier=words,name=NPtwo,phantom]
                          ]
                      ]
                      [ADVP,circle,draw,scale=0.7,phantom
                         [{daily},draw,fill=orange,fill opacity=0.1,align=center,tier=words,name=daily,phantom]
                      ]
                  ]
              ]
              \end{forest}
          }
      }
      \subcaptionbox{Step 3.\label{cfg:top:down:step:three}}{
          \scalingTD{
              \begin{forest}
              [S,circle,draw,scale=1.2,tikz={\node [red, fit=()(Mary)(VPone)]{};}
                  [NP,circle,draw
                      [{Mary},draw,fill=green,fill opacity=0.1,align=center,tier=words,name=Mary]]
                  [\phantom{VP},circle,draw,name=VPone
                      [VP,circle,draw,phantom
                          [V,draw,circle,scale=1.1,phantom
                              [{reads},draw,fill=red,fill opacity=0.1,align=center,tier=words,name=reads,phantom]
                          ]
                          [NP,circle,draw,phantom
                             [{papers},draw,fill=blue,fill opacity=0.1,align=center,tier=words,name=NPtwo,phantom]
                          ]
                      ]
                      [ADVP,circle,draw,scale=0.7,phantom
                         [{daily},draw,fill=orange,fill opacity=0.1,align=center,tier=words,name=daily,phantom]
                      ]
                  ]
              ]
              \end{forest}
          }
      } 

      \subcaptionbox{Step 4.\label{cfg:top:down:step:four}}{
          \scalingTD{
              \begin{forest}
              [S,circle,draw,scale=1.2,tikz={\node [gray, fit=()(Mary)(VPone)]{};}
                  [NP,circle,draw
                      [{Mary},draw,fill=green,fill opacity=0.1,align=center,tier=words,name=Mary]]
                  [\phantom{VP},circle,draw,name=VPone
                      [VP,circle,draw,phantom
                          [V,draw,circle,scale=1.1,phantom
                              [{reads},draw,fill=red,fill opacity=0.1,align=center,tier=words,name=reads]
                          ]
                          [NP,circle,draw,phantom
                             [{papers},draw,fill=blue,fill opacity=0.1,align=center,tier=words,name=NPtwo,phantom]
                          ]
                      ]
                      [ADVP,circle,draw,scale=0.7,phantom
                         [{daily},draw,fill=orange,fill opacity=0.1,align=center,tier=words,name=daily,phantom]
                      ]
                  ]
              ]
              \end{forest}
          }
      }
      \subcaptionbox{Step 5.\label{cfg:top:down:step:five}}{
          \scalingTD{
              \begin{forest}
              [S,circle,draw,scale=1.2,tikz={\node [blue, fit=()(Mary)(ADVP)]{};}
                  [NP,circle,draw
                      [{Mary},draw,fill=green,fill opacity=0.1,align=center,tier=words,name=Mary]]
                  [VP,circle,draw,name=VPone
                      [\phantom{VP},circle,draw,name=VPtwo
                          [V,draw,circle,scale=1.1,phantom
                              [{reads},draw,fill=red,fill opacity=0.1,align=center,tier=words,name=reads]
                          ]
                          [NP,circle,draw,phantom
                             [{papers},draw,fill=blue,fill opacity=0.1,align=center,tier=words,name=NPtwo,phantom]
                          ]
                      ]
                      [\phantom{ADVP},circle,draw,scale=0.7,name=ADVP
                         [{daily},draw,fill=orange,fill opacity=0.1,align=center,tier=words,name=daily,phantom]
                      ]
                  ]
              ]
              \end{forest}
          }
      }
      \subcaptionbox{Step 6.\label{cfg:top:down:step:six}}{
          \scalingTD{
              \begin{forest}
              [S,circle,draw,scale=1.2,tikz={\node [blue, fit=()(Mary)(ADVP)]{};}
                  [NP,circle,draw
                      [{Mary},draw,fill=green,fill opacity=0.1,align=center,tier=words,name=Mary]]
                  [VP,circle,draw,name=VPone
                      [VP,circle,draw,name=VPtwo
                          [\phantom{V},draw,circle,scale=1.1
                              [{reads},draw,fill=red,fill opacity=0.1,align=center,tier=words,name=reads,edge=white]
                          ]
                          [\phantom{NP},circle,draw
                             [{papers},draw,fill=blue,fill opacity=0.1,align=center,tier=words,name=NPtwo,phantom]
                          ]
                      ]
                      [\phantom{ADVP},circle,draw,scale=0.7,name=ADVP
                         [{daily},draw,fill=orange,fill opacity=0.1,align=center,tier=words,name=daily,phantom]
                      ]
                  ]
              ]
              \end{forest}
          }
      }
      \subcaptionbox{Step 7.\label{cfg:top:down:step:seven}}{
          \scalingTD{
              \begin{forest}
              [S,circle,draw,scale=1.2,tikz={\node [blue, fit=()(Mary)(ADVP)]{};}
                  [NP,circle,draw
                      [{Mary},draw,fill=green,fill opacity=0.1,align=center,tier=words,name=Mary]]
                  [VP,circle,draw,name=VPone
                      [VP,circle,draw
                          [V,draw,circle,scale=1.1
                              [{reads},draw,fill=red,fill opacity=0.1,align=center,tier=words,name=reads]
                          ]
                          [\phantom{NP},circle,draw
                             [{papers},draw,fill=blue,fill opacity=0.1,align=center,tier=words,name=NPtwo,phantom]
                          ]
                      ]
                      [\phantom{ADVP},circle,draw,scale=0.7,name=ADVP
                         [{daily},draw,fill=orange,fill opacity=0.1,align=center,tier=words,name=daily,phantom]
                      ]
                  ]
              ]
              \end{forest}
          }
      } 

      \subcaptionbox{Step 8.\label{cfg:top:down:step:eight}}{
          \scalingTD{
              \begin{forest}
              [S,circle,draw,scale=1.2,tikz={\node [gray, fit=()(Mary)(ADVP)]{};}
                  [NP,circle,draw
                      [{Mary},draw,fill=green,fill opacity=0.1,align=center,tier=words,name=Mary]]
                  [VP,circle,draw,name=VPone
                      [VP,circle,draw
                          [V,draw,circle,scale=1.1
                              [{reads},draw,fill=red,fill opacity=0.1,align=center,tier=words,name=reads]
                          ]
                          [\phantom{NP},circle,draw
                             [{papers},draw,fill=blue,fill opacity=0.1,align=center,tier=words,name=NPtwo,edge=white]
                          ]
                      ]
                      [\phantom{ADVP},circle,draw,scale=0.7,name=ADVP
                         [{daily},draw,fill=orange,fill opacity=0.1,align=center,tier=words,name=daily,phantom]
                      ]
                  ]
              ]
              \end{forest}
          }
      }
      \subcaptionbox{Step 9.\label{cfg:top:down:step:nine}}{
          \scalingTD{
              \begin{forest}
              [S,circle,draw,scale=1.2,tikz={\node [green, fit=()(Mary)(ADVP)]{};}
                  [NP,circle,draw
                      [{Mary},draw,fill=green,fill opacity=0.1,align=center,tier=words,name=Mary]]
                  [VP,circle,draw,name=VPone
                      [VP,circle,draw
                          [V,draw,circle,scale=1.1
                              [{reads},draw,fill=red,fill opacity=0.1,align=center,tier=words,name=reads]
                          ]
                          [NP,circle,draw
                             [{papers},draw,fill=blue,fill opacity=0.1,align=center,tier=words,name=NPtwo]
                          ]
                      ]
                      [\phantom{ADVP},circle,draw,scale=0.7,name=ADVP
                         [{daily},draw,fill=orange,fill opacity=0.1,align=center,tier=words,name=daily,phantom]
                      ]
                  ]
              ]
              \end{forest}
          }
      }
      
      \subcaptionbox{Step 10.\label{cfg:top:down:step:ten}}{
          \scalingTD{
              \begin{forest}
              [S,circle,draw,scale=1.2,tikz={\node [gray, fit=()(Mary)(daily)]{};}
                  [NP,circle,draw
                      [{Mary},draw,fill=green,fill opacity=0.1,align=center,tier=words,name=Mary]]
                  [VP,circle,draw,name=VPone
                      [VP,circle,draw
                          [V,draw,circle,scale=1.1
                              [{reads},draw,fill=red,fill opacity=0.1,align=center,tier=words,name=reads]
                          ]
                          [NP,circle,draw
                             [{papers},draw,fill=blue,fill opacity=0.1,align=center,tier=words,name=NPtwo]
                          ]
                      ]
                      [\phantom{ADVP},circle,draw,scale=0.7
                         [{daily},draw,fill=orange,fill opacity=0.1,align=center,tier=words,name=daily,edge=white]
                      ]
                  ]
              ]
              \end{forest}
          }
      }
      \subcaptionbox{Step 11.\label{cfg:top:down:step:eleven}}{
          \scalingTD{
              \begin{forest}
              [S,circle,draw,scale=1.2,tikz={\node [purple, fit=()(Mary)(daily)]{};}
                  [NP,circle,draw
                      [{Mary},draw,fill=green,fill opacity=0.1,align=center,tier=words,name=Mary]]
                  [VP,circle,draw,name=VPone
                      [VP,circle,draw
                          [V,draw,circle,scale=1.1
                              [{reads},draw,fill=red,fill opacity=0.1,align=center,tier=words,name=reads]
                          ]
                          [NP,circle,draw
                             [{papers},draw,fill=blue,fill opacity=0.1,align=center,tier=words,name=NPtwo]
                          ]
                      ]
                      [ADVP,circle,draw,scale=0.7
                         [{daily},draw,fill=orange,fill opacity=0.1,align=center,tier=words,name=daily]
                      ]
                  ]
              ]
              \end{forest}
          }
      }
      \caption{Top-Down parsing strategy for Context-Free Grammars.
      Each row indicates parser steps associated with a single word.
      Empty circles indicates postulated nodes where the category has not yet been fixed.}
      \label{fig:cfg:top:down}
\end{figure*}

Left-corner parsing is a third type of constituency parser that fills the space between extreme speculation (top-down parsing) and extreme non-incrementality (bottom-up parsing).%
A left-corner parser builds all constituents bottom-up, just like the bottom-up parser, but it also predicts the parent node that is not fully complete and of whom the current constituent is a left child (``left-corner''). 
For instance, in Figure~\ref{fig:cfg:left:corner} Step~2 builds the same constituent as the bottom-up parser, but in Step~3 it also builds (i.e. speculates) the parent that is not fully complete. 
This partial speculation makes the left-corner parser more incremental than the bottom-up strategy, but still less incremental than top-down.
Left-corner parsing establishes relation between \textit{Mary} and \textit{reads} after consuming words ``\textit{Mary reads papers}''.

\newcommand{\scalingLC}[1]{\scalebox{0.5}{#1}}

\begin{figure*}
    \captionsetup[subfigure]{labelformat=empty}
      \subcaptionbox{Step 1.\label{cfg:left:corner:step:one}}{
          \scalingLC{
              \begin{forest}
              [S,phantom,circle,draw,scale=1.2,tikz={\node [gray, fit=()(Mary)(daily)]{};}
                  [NP,circle,draw,phantom
                      [{Mary},draw,fill=green,fill opacity=0.1,align=center,tier=words,name=Mary]]
                  [VP,circle,draw,phantom
                      [VP,circle,draw,phantom
                          [V,draw,circle,scale=1.1,phantom
                              [{reads},draw,fill=red,fill opacity=0.1,align=center,tier=words,name=reads,phantom]
                          ]
                          [NP,circle,draw,phantom
                             [{papers},draw,fill=blue,fill opacity=0.1,align=center,tier=words,name=NPtwo,phantom]
                          ]
                      ]
                      [ADVP,circle,draw,scale=0.7,phantom
                         [{daily},draw,fill=orange,fill opacity=0.1,align=center,tier=words,name=daily,phantom]
                      ]
                  ]
              ]
              \end{forest}
          }
      }
      \subcaptionbox{Step 2.\label{cfg:left:corner:step:two}}{
          \scalingLC{
              \begin{forest}
              [S,phantom,circle,draw,scale=1.2,tikz={\node [red, fit=()(Mary)(daily)]{};}
                  [NP,circle,draw
                      [{Mary},draw,fill=green,fill opacity=0.1,align=center,tier=words,name=Mary]]
                  [VP,circle,draw,phantom
                      [VP,circle,draw,phantom
                          [V,draw,circle,scale=1.1,phantom
                              [{reads},draw,fill=red,fill opacity=0.1,align=center,tier=words,name=reads,phantom]
                          ]
                          [NP,circle,draw,phantom
                             [{papers},draw,fill=blue,fill opacity=0.1,align=center,tier=words,name=NPtwo,phantom]
                          ]
                      ]
                      [ADVP,circle,draw,scale=0.7,phantom
                         [{daily},draw,fill=orange,fill opacity=0.1,align=center,tier=words,name=daily,phantom]
                      ]
                  ]
              ]
              \end{forest}
          }
      }
      \subcaptionbox{Step 3.\label{cfg:left:corner:step:three}}{
          \scalingLC{
              \begin{forest}
              [S,circle,draw,scale=1.2,tikz={\node [red, fit=()(Mary)(daily)]{};}
                  [NP,circle,draw
                      [{Mary},draw,fill=green,fill opacity=0.1,align=center,tier=words,name=Mary]]
                  [\phantom{VP},circle,draw
                      [VP,circle,draw,phantom
                          [V,draw,circle,scale=1.1,phantom
                              [{reads},draw,fill=red,fill opacity=0.1,align=center,tier=words,name=reads,phantom]
                          ]
                          [NP,circle,draw,phantom
                             [{papers},draw,fill=blue,fill opacity=0.1,align=center,tier=words,name=NPtwo,phantom]
                          ]
                      ]
                      [ADVP,circle,draw,scale=0.7,phantom
                         [{daily},draw,fill=orange,fill opacity=0.1,align=center,tier=words,name=daily,phantom]
                      ]
                  ]
              ]
              \end{forest}
          }
      }

      \subcaptionbox{Step 4.\label{cfg:left:corner:step:four}}{
          \scalingLC{
              \begin{forest}
              [S,circle,draw,scale=1.2,tikz={\node [gray, fit=()(Mary)(daily)]{};}
                  [NP,circle,draw
                      [{Mary},draw,fill=green,fill opacity=0.1,align=center,tier=words,name=Mary]]
                  [\phantom{VP},circle,draw
                      [VP,circle,draw,phantom
                          [V,draw,circle,scale=1.1,phantom
                              [{reads},draw,fill=red,fill opacity=0.1,align=center,tier=words,name=reads]
                          ]
                          [NP,circle,draw,phantom
                             [{papers},draw,fill=blue,fill opacity=0.1,align=center,tier=words,name=NPtwo,phantom]
                          ]
                      ]
                      [ADVP,circle,draw,scale=0.7,phantom
                         [{daily},draw,fill=orange,fill opacity=0.1,align=center,tier=words,name=daily,phantom]
                      ]
                  ]
              ]
              \end{forest}
          }
      }
      \subcaptionbox{Step 5.\label{cfg:left:corner:step:five}}{
          \scalingLC{
              \begin{forest}
              [S,circle,draw,scale=1.2,tikz={\node [blue, fit=()(Mary)(daily)]{};}
                  [NP,circle,draw
                      [{Mary},draw,fill=green,fill opacity=0.1,align=center,tier=words,name=Mary]]
                  [\phantom{VP},circle,draw
                      [VP,circle,draw,phantom
                          [V,draw,circle,scale=1.1
                              [{reads},draw,fill=red,fill opacity=0.1,align=center,tier=words,name=reads]
                          ]
                          [NP,circle,draw,phantom
                             [{papers},draw,fill=blue,fill opacity=0.1,align=center,tier=words,name=NPtwo,phantom]
                          ]
                      ]
                      [ADVP,circle,draw,scale=0.7,phantom
                         [{daily},draw,fill=orange,fill opacity=0.1,align=center,tier=words,name=daily,phantom]
                      ]
                  ]
              ]
              \end{forest}
          }
      }
      \subcaptionbox{Step 6.\label{cfg:left:corner:step:six}}{
          \scalingLC{
              \begin{forest}
              [S,circle,draw,scale=1.2,tikz={\node [blue, fit=()(Mary)(daily)]{};}
                  [NP,circle,draw
                      [{Mary},draw,fill=green,fill opacity=0.1,align=center,tier=words,name=Mary]]
                  [\phantom{VP},circle,draw
                      [VP,circle,draw,edge=white
                          [V,draw,circle,scale=1.1
                              [{reads},draw,fill=red,fill opacity=0.1,align=center,tier=words,name=reads]
                          ]
                          [\phantom{NP},circle,draw
                             [{papers},draw,fill=blue,fill opacity=0.1,align=center,tier=words,name=NPtwo,phantom]
                          ]
                      ]
                      [ADVP,circle,draw,scale=0.7,phantom
                         [{daily},draw,fill=orange,fill opacity=0.1,align=center,tier=words,name=daily,phantom]
                      ]
                  ]
              ]
              \end{forest}
          }
      }

      \subcaptionbox{Step 7.\label{cfg:left:corner:step:seven}}{
          \scalingLC{
              \begin{forest}
              [S,circle,draw,scale=1.2,tikz={\node [gray, fit=()(Mary)(daily)]{};}
                  [NP,circle,draw
                      [{Mary},draw,fill=green,fill opacity=0.1,align=center,tier=words,name=Mary]]
                  [\phantom{VP},circle,draw
                      [VP,circle,draw,edge=white
                          [V,draw,circle,scale=1.1
                              [{reads},draw,fill=red,fill opacity=0.1,align=center,tier=words,name=reads]
                          ]
                          [\phantom{NP},circle,draw
                             [{papers},draw,fill=blue,fill opacity=0.1,align=center,tier=words,name=NPtwo,edge=white]
                          ]
                      ]
                      [ADVP,circle,draw,scale=0.7,phantom
                         [{daily},draw,fill=orange,fill opacity=0.1,align=center,tier=words,name=daily,phantom]
                      ]
                  ]
              ]
              \end{forest}
          }
      }
      \subcaptionbox{Step 8.\label{cfg:left:corner:step:eight}}{
          \scalingLC{
              \begin{forest}
              [S,circle,draw,scale=1.2,tikz={\node [green, fit=()(Mary)(daily)]{};}
                  [NP,circle,draw
                      [{Mary},draw,fill=green,fill opacity=0.1,align=center,tier=words,name=Mary]]
                  [\phantom{VP},circle,draw
                      [VP,circle,draw,edge=white
                          [V,draw,circle,scale=1.1
                              [{reads},draw,fill=red,fill opacity=0.1,align=center,tier=words,name=reads]
                          ]
                          [NP,circle,draw
                             [{papers},draw,fill=blue,fill opacity=0.1,align=center,tier=words,name=NPtwo]
                          ]
                      ]
                      [ADVP,circle,draw,scale=0.7,phantom
                         [{daily},draw,fill=orange,fill opacity=0.1,align=center,tier=words,name=daily,phantom]
                      ]
                  ]
              ]
              \end{forest}
          }
      }
      \subcaptionbox{Step 9.\label{cfg:left:corner:step:nine}}{
          \scalingLC{
              \begin{forest}
              [S,circle,draw,scale=1.2,tikz={\node [green, fit=()(Mary)(daily)]{};}
                  [NP,circle,draw
                      [{Mary},draw,fill=green,fill opacity=0.1,align=center,tier=words,name=Mary]]
                  [VP,circle,draw
                      [VP,circle,draw
                          [V,draw,circle,scale=1.1
                              [{reads},draw,fill=red,fill opacity=0.1,align=center,tier=words,name=reads]
                          ]
                          [NP,circle,draw
                             [{papers},draw,fill=blue,fill opacity=0.1,align=center,tier=words,name=NPtwo]
                          ]
                      ]
                      [\phantom{ADVP},circle,draw,scale=0.7
                         [{daily},draw,fill=orange,fill opacity=0.1,align=center,tier=words,name=daily,phantom]
                      ]
                  ]
              ]
              \end{forest}
          }
      }

      \subcaptionbox{Step 10.\label{cfg:left:corner:step:ten}}{
          \scalingLC{
              \begin{forest}
              [S,circle,draw,scale=1.2,tikz={\node [gray, fit=()(Mary)(daily)]{};}
                  [NP,circle,draw
                      [{Mary},draw,fill=green,fill opacity=0.1,align=center,tier=words,name=Mary]]
                  [VP,circle,draw
                      [VP,circle,draw
                          [V,draw,circle,scale=1.1
                              [{reads},draw,fill=red,fill opacity=0.1,align=center,tier=words,name=reads]
                          ]
                          [NP,circle,draw
                             [{papers},draw,fill=blue,fill opacity=0.1,align=center,tier=words,name=NPtwo]
                          ]
                      ]
                      [\phantom{ADVP},circle,draw,scale=0.7
                         [{daily},draw,fill=orange,fill opacity=0.1,align=center,tier=words,name=daily,edge=white]
                      ]
                  ]
              ]
              \end{forest}
          }
      }
      \subcaptionbox{Step 11.\label{cfg:left:corner:step:eleven}}{
          \scalingLC{
              \begin{forest}
              [S,circle,draw,scale=1.2,tikz={\node [purple, fit=()(Mary)(daily)]{};}
                  [NP,circle,draw
                      [{Mary},draw,fill=green,fill opacity=0.1,align=center,tier=words,name=Mary]]
                  [VP,circle,draw
                      [VP,circle,draw
                          [V,draw,circle,scale=1.1
                              [{reads},draw,fill=red,fill opacity=0.1,align=center,tier=words,name=reads]
                          ]
                          [NP,circle,draw
                             [{papers},draw,fill=blue,fill opacity=0.1,align=center,tier=words,name=NPtwo]
                          ]
                      ]
                      [ADVP,circle,draw,scale=0.7
                         [{daily},draw,fill=orange,fill opacity=0.1,align=center,tier=words,name=daily]
                      ]
                  ]
              ]
              \end{forest}
          }
      }
      \caption{Left-Corner parsing strategy for Context-Free Grammars.
      Each row indicates parser steps associated with a single word.
      Empty circles indicates postulated nodes where the category has not yet been fixed.}
      \label{fig:cfg:left:corner}
\end{figure*}

\subsubsection*{Measuring the effort of constituency parsing} \label{sec:effort}

We apply a notion of computational~work that is inspired by Kaplan's Number of Transitions Made or Attempted~\citep{kaplan72}. 
This is a count of basic operations in a parsing~mechanism such~as an Augmented Transition Network \citep[for an overview of computational~psycholinguistics that introduces this mechanism see][]{hale17}. 
In this work with naturalistic texts as opposed to known garden-path sentences, we set~aside ambiguity~resolution and simply count derivation tree nodes in the manner of \citet{frazier1985syntactic}. 
To arrive at an incremental complexity metric, we sum up the number of nodes that would be visited (on a given parsing strategy) between one derivation-tree leaf node and its successor in the linear word-string. 
This summation is schematized as the number of panels per row in the diagrams shown in Figures \ref{fig:cfg:bottom:up}--\ref{fig:cfg:left:corner}.
It is this count that quantifies the amount of work predicted for each word. 
We have one indicator of processing effort for each constituency parsing strategy: \bottomup{} and \topdown{} (\leftcorner{} is excluded for methodological reasons discussed in \ref{sec:analysis}.). The parsing strategies are computed over the constituency trees provided by the \emph{Benepar} parser \citep{kitaev-etal-2019-multilingual} that is a state-of-the-art constituency parser with a accuracy of 
$95.9\%$ in recovering labelled constituents in English. %

\subsubsection*{Predictors from Combinatory Categorial Grammar (CCG)}

As introduced earlier on page~\pageref{sec:ccg} 
Combinatory Categorial Grammar (CCG) has multiple advantages over context-free grammar.
These arise from CCG's flexible notion of a constituency and its flexibility in building~up this
constituency structure.
The first~advantage is that CCG can generate some constructions that appear in human language
whose analysis within context-free grammar would be linguistically-inadequate \citep{stanojevic:steedman:cl:2021}.
The second advantage %
is that CCG can be very incremental without resorting to a
speculative top-down parsing strategy.

We illustrate the contrast between CFG and CCG with an example repeated from Figure~\ref{fig:comparison} for the sentence ``Mary reads papers''.
In that sentence, a CFG offers no way of building a node for subject-verb combination ``Mary reads'' while CCG can represent it as $\textit{S/NP}$;
this term may be glossed as \guillemotleft a sentence that is missing a noun-phrase on its right\guillemotright.
Such flexibility allows CCG to derive the same semantic representation in several different ways. 
As shown
in Figure~\ref{fig:comparison} CCG can form both left- and right-branching structures for sentence ``Mary reads papers''. 
By contrast, CFG can form only a right-branching structure. 
To accomplish this flexibility CCG makes use of a set of combinators with type-raising and function composition that we do not review here. 
For a compact introduction to CCG combinators see \citet{steedman-ccg-intro}.

The CCG right-branching tree is quite similar to the CFG tree and processing it with a bottom-up parser would be equally non-incremental as shown by the trace in Figure~\ref{fig:ccg:right}. 
However, the left-branching tree has completely different properties; the key advantage of the left-branching structure is that now even the simplest bottom-up parser is very incremental. 
Figure~\ref{fig:ccg:left} illustrates this with the trace of a left-branching bottom-up CCG parser.

\newcommand{\scalingCCGRight}[1]{\scalebox{0.65}{#1}}

\begin{figure*}
    \captionsetup[subfigure]{labelformat=empty}
      \subcaptionbox{Step 1.\label{ccg:right:step:one}}{
          \scalingCCGRight{
              \begin{forest}
              [\textit{S},phantom,circle,draw,scale=1.5,tikz={\node [ gray, fit=()(Mary)(papers)]{};}
                  [{\textit{NP}\\Mary},draw,fill=green,fill opacity=0.1,align=center,tier=words,name=Mary]
                  [\textit{S{\bs}NP},phantom,circle,draw
                      [{\textit{(S{\bs}NP)/NP}\\reads},phantom,draw,fill=red,fill opacity=0.1,align=center,tier=words,name=reads]
                     [{\textit{NP}\\papers},phantom,draw,fill=blue,fill opacity=0.1,align=center,tier=words,name=papers]
                  ]
              ]
              \end{forest}
          }
      }

      \subcaptionbox{Step 2.\label{ccg:right:step:two}}{
          \scalingCCGRight{
              \begin{forest}
              [\textit{S},phantom,circle,draw,scale=1.5,tikz={\node [ gray, fit=()(Mary)(papers)]{};}
                  [{\textit{NP}\\Mary},draw,fill=green,fill opacity=0.1,align=center,tier=words,name=Mary]
                  [\textit{S{\bs}NP},phantom,circle,draw
                      [{\textit{(S{\bs}NP)/NP}\\reads},draw,fill=red,fill opacity=0.1,align=center,tier=words,name=reads]
                     [{\textit{NP}\\papers},phantom,draw,fill=blue,fill opacity=0.1,align=center,tier=words,name=papers]
                  ]
              ]
              \end{forest}
          }
      }

      \subcaptionbox{Step 3.\label{ccg:right:step:three}}{
          \scalingCCGRight{
              \begin{forest}
              [\textit{S},phantom,circle,draw,scale=1.5,tikz={\node [ gray, fit=()(Mary)(papers)]{};}
                  [{\textit{NP}\\Mary},draw,fill=green,fill opacity=0.1,align=center,tier=words,name=Mary]
                  [\textit{S{\bs}NP},phantom,circle,draw
                      [{\textit{(S{\bs}NP)/NP}\\reads},draw,fill=red,fill opacity=0.1,align=center,tier=words,name=reads]
                     [{\textit{NP}\\papers},draw,fill=blue,fill opacity=0.1,align=center,tier=words,name=papers]
                  ]
              ]
              \end{forest}
          }
      }
      \subcaptionbox{Step 4.\label{ccg:right:step:four}}{
          \scalingCCGRight{
              \begin{forest}
              [\textit{S},phantom,circle,draw,scale=1.5,tikz={\node [ green, fit=()(Mary)(papers)]{};}
                  [{\textit{NP}\\Mary},draw,fill=green,fill opacity=0.1,align=center,tier=words,name=Mary]
                  [\textit{S{\bs}NP},circle,draw
                      [{\textit{(S{\bs}NP)/NP}\\reads},draw,fill=red,fill opacity=0.1,align=center,tier=words,name=reads]
                     [{\textit{NP}\\papers},draw,fill=blue,fill opacity=0.1,align=center,tier=words,name=papers]
                  ]
              ]
              \end{forest}
          }
      }
      \subcaptionbox{Step 5.\label{ccg:right:step:five}}{
          \scalingCCGRight{
              \begin{forest}
              [\textit{S},circle,draw,scale=1.5,tikz={\node [ green, fit=()(Mary)(papers)]{};}
                  [{\textit{NP}\\Mary},draw,fill=green,fill opacity=0.1,align=center,tier=words,name=Mary]
                  [\textit{S{\bs}NP},circle,draw
                      [{\textit{(S{\bs}NP)/NP}\\reads},draw,fill=red,fill opacity=0.1,align=center,tier=words,name=reads]
                     [{\textit{NP}\\papers},draw,fill=blue,fill opacity=0.1,align=center,tier=words,name=papers]
                  ]
              ]
              \end{forest}
          }
      }
      \caption{Parsing steps for CCG right-branching.
      Each row indicates parser steps associated with a single word.}
      \label{fig:ccg:right}
\end{figure*}
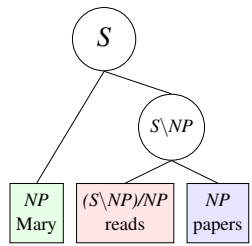

\newcommand{\scalingCCGLeft}[1]{\scalebox{0.65}{#1}}

\begin{figure*}
    \captionsetup[subfigure]{labelformat=empty}
      \subcaptionbox{Step 1.\label{ccg:left:step:one}}{
          \scalingCCGLeft{
              \begin{forest}
              [\textit{S},phantom,circle,draw,scale=1.5,tikz={\node [ gray, fit=()(tr)(Mary)(papers)]{};}
                  [\textit{S/NP},phantom,ellipse,draw,scale=1.1
                      [\textit{S/(S{\bs}NP)},phantom,ellipse,draw,scale=1.1,name=tr
                          [{\textit{NP}\\Mary},draw,fill=green,fill opacity=0.1,align=center,tier=words,name=Mary]
                      ]
                      [{\textit{(S{\bs}NP)/NP}\\reads},phantom,draw,fill=red,fill opacity=0.1,align=center,tier=words,name=reads]
                  ]
                  [{\textit{NP}\\papers},phantom,draw,fill=blue,fill opacity=0.1,align=center,tier=words,name=papers]
              ]
              \end{forest}
          }
      }
      \subcaptionbox{Step 2.\label{ccg:left:step:two}}{
          \scalingCCGLeft{
              \begin{forest}
              [\textit{S},phantom,circle,draw,scale=1.5,tikz={\node [ red, fit=()(tr)(Mary)(papers)]{};}
                  [\textit{S/NP},phantom,ellipse,draw,scale=1.1
                      [\textit{S/(S{\bs}NP)},ellipse,draw,scale=1.1,name=tr
                          [{\textit{NP}\\Mary},draw,fill=green,fill opacity=0.1,align=center,tier=words,name=Mary]
                      ]
                      [{\textit{(S{\bs}NP)/NP}\\reads},phantom,draw,fill=red,fill opacity=0.1,align=center,tier=words,name=reads]
                  ]
                  [{\textit{NP}\\papers},phantom,draw,fill=blue,fill opacity=0.1,align=center,tier=words,name=papers]
              ]
              \end{forest}
          }
      }

      \subcaptionbox{Step 3.\label{ccg:left:step:three}}{
          \scalingCCGLeft{
              \begin{forest}
              [\textit{S},phantom,circle,draw,scale=1.5,tikz={\node [ gray, fit=()(tr)(Mary)(papers)]{};}
                  [\textit{S/NP},phantom,ellipse,draw,scale=1.1
                      [\textit{S/(S{\bs}NP)},ellipse,draw,scale=1.1,name=tr
                          [{\textit{NP}\\Mary},draw,fill=green,fill opacity=0.1,align=center,tier=words,name=Mary]
                      ]
                      [{\textit{(S{\bs}NP)/NP}\\reads},draw,fill=red,fill opacity=0.1,align=center,tier=words,name=reads]
                  ]
                  [{\textit{NP}\\papers},phantom,draw,fill=blue,fill opacity=0.1,align=center,tier=words,name=papers]
              ]
              \end{forest}
          }
      }
      \subcaptionbox{Step 4.\label{ccg:left:step:four}}{
          \scalingCCGLeft{
              \begin{forest}
              [\textit{S},phantom,circle,draw,scale=1.5,tikz={\node [ blue, fit=()(tr)(Mary)(papers)]{};}
                  [\textit{S/NP},ellipse,draw,scale=1.1
                      [\textit{S/(S{\bs}NP)},ellipse,draw,scale=1.1,name=tr
                          [{\textit{NP}\\Mary},fill=green,fill opacity=0.1,draw,align=center,tier=words,name=Mary]
                      ]
                      [{\textit{(S{\bs}NP)/NP}\\reads},draw,fill=red,fill opacity=0.1,align=center,tier=words,name=reads]
                  ]
                  [{\textit{NP}\\papers},phantom,draw,fill=blue,fill opacity=0.1,align=center,tier=words,name=papers]
              ]
              \end{forest}
          }
      }
      
      \subcaptionbox{Step 5.\label{ccg:left:step:five}}{
          \scalingCCGLeft{
              \begin{forest}
              [\textit{S},phantom,circle,draw,scale=1.5,tikz={\node [ gray, fit=()(tr)(Mary)(papers)]{};}
                  [\textit{S/NP},ellipse,draw,scale=1.1
                      [\textit{S/(S{\bs}NP)},ellipse,draw,scale=1.1,name=tr
                          [{\textit{NP}\\Mary},draw,fill=green,fill opacity=0.1,align=center,tier=words,name=Mary]
                      ]
                      [{\textit{(S{\bs}NP)/NP}\\reads},draw,fill=red,fill opacity=0.1,align=center,tier=words,name=reads]
                  ]
                  [{\textit{NP}\\papers},draw,fill=blue,fill opacity=0.1,align=center,tier=words,name=papers]
              ]
              \end{forest}
          }
      }
      \subcaptionbox{Step 6.\label{ccg:left:step:six}}{
          \scalingCCGLeft{
              \begin{forest}
              [\textit{S},circle,draw,scale=1.5,tikz={\node [ green, fit=()(tr)(Mary)(papers)]{};}
                  [\textit{S/NP},ellipse,draw,scale=1.1
                      [\textit{S/(S{\bs}NP)},ellipse,draw,scale=1.1,name=tr
                          [{\textit{NP}\\Mary},draw,fill=green,fill opacity=0.1,align=center,tier=words,name=Mary]
                      ]
                      [{\textit{(S{\bs}NP)/NP}\\reads},draw,fill=red,fill opacity=0.1,align=center,tier=words,name=reads]
                  ]
                  [{\textit{NP}\\papers},draw,fill=blue,fill opacity=0.1,align=center,tier=words,name=papers]
              ]
              \end{forest}
          }
      }
      \caption{Parsing steps for CCG left-branching. 
      Each row indicates parser steps associated with a single word.}
      \label{fig:ccg:left}
\end{figure*}
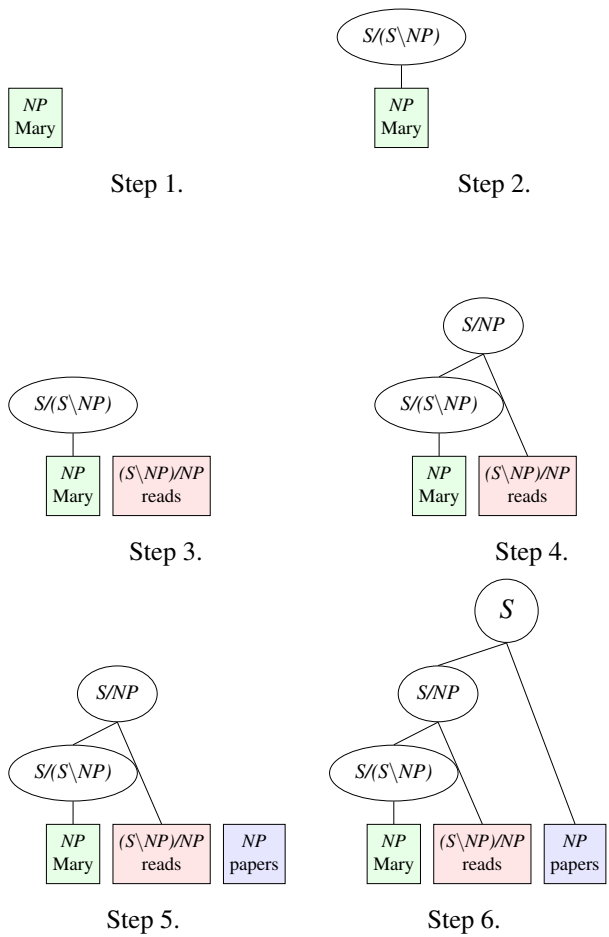

\subsubsection*{Right Adjunction and CCG Parsing with Revealing}

While left-branching trees offer a simple solution for achieving incrementality, they are not sufficient. 
One particular problem is presented by this is \textit{right adjuncts}, or optional modifiers that appear to the right of the constituents they are modifying. 
Their optionality makes these constituents hard to predict across parsers and grammar formalisms. 
\citet{Hale:2012ve} proposed a solution for the CFG case in the form of \textit{generalized} left-corner parsing \citep{demers-1977-generalized-left-corner}. 
This generalized form allows for CFG rules to be annotated with the level of incrementality they should have. 
\citeauthor{Hale:2012ve} proposes to make adjunction rules less~incremental than non-adjunction rules. 
The problem with that approach is that there is no way for an incremental parser to know if it should use the adjunction or non-adjunction rule until after it sees whether there is an adjunct.

Ideally the parser would be able to incrementally process the sentence and assume that there is no right adjunct, but in case the right adjunct appears the parser should be able to incorporate it without costly backtracking. 
This is precisely what the Revealing CCG parser from \citet{stanojevicCCGParsingAlgorithm2019} does by exploiting the flexibility of CCG trees and combinators. 
To understand the Revealing parsing strategy, consider the example shown in Figure~\ref{fig:ccg:rotate}. 
Until Step~7 the parsing steps here are the same as in a left-branching CCG tree (\ref{fig:ccg:left}). 
In Step~7 the Revealing strategy applies a \textsc{rotation} operation that converts the already-built left branching tree into a semantically equivalent right branching tree. 
This is a computationally efficient operation that, crucially, is fully deterministic. 
It does not incur any significant processing cost, but it does provide a representation that will be useful later if a right-adjunct appears. 
In Step~8 the right adjunct has appeared. 
From its type the parser recognizes that it looks to modify a verb phrase (i.e. a constituent with CCG type $\textit{S{\bs}NP}$). 
The parser looks for the candidate for modification in the right edge of the preceding subtree.
It finds the constituent ``reads papers'' to be of the right type and \textsc{reveal}s   it for modification in Step~9.

\newcommand{\scalingCCGRotate}[1]{\scalebox{0.55}{#1}}

\begin{figure*}
    \captionsetup[subfigure]{labelformat=empty}
      \subcaptionbox{Step 1.\label{ccg:rotate:step:one}}{
          \scalingCCGRotate{
              \begin{forest}
              [\textit{S},phantom,circle,draw,scale=1.5,tikz={\node [ gray, fit=()(tr)(Mary)(Mary)]{};}
                  [\textit{S/NP},phantom,ellipse,draw,scale=1.1
                      [\textit{S/(S{\bs}NP)},phantom,ellipse,draw,scale=1.1,name=tr
                          [{\textit{NP}\\Mary},fill=green,fill opacity=0.1,draw,align=center,tier=words,name=Mary]
                      ]
                      [{\textit{(S{\bs}NP)/NP}\\reads},fill=red,fill opacity=0.1,phantom,draw,align=center,tier=words,name=reads]
                  ]
                  [{\textit{NP}\\papers},fill=blue,fill opacity=0.1,phantom,draw,align=center,tier=words,name=papers]
              ]
              \end{forest}
          }
      }
      \subcaptionbox{Step 2.\label{ccg:rotate:step:two}}{
          \scalingCCGRotate{
              \begin{forest}
              [\textit{S},phantom,circle,draw,scale=1.5,tikz={\node [ red, fit=()(tr)(Mary)(Mary)]{};}
                  [\textit{S/NP},phantom,ellipse,draw,scale=1.1
                      [\textit{S/(S{\bs}NP)},ellipse,draw,scale=1.1,name=tr
                          [{\textit{NP}\\Mary},fill=green,fill opacity=0.1,draw,align=center,tier=words,name=Mary]
                      ]
                      [{\textit{(S{\bs}NP)/NP}\\reads},fill=red,fill opacity=0.1,phantom,draw,align=center,tier=words,name=reads]
                  ]
                  [{\textit{NP}\\papers},fill=blue,fill opacity=0.1,phantom,draw,align=center,tier=words,name=papers]
              ]
              \end{forest}
          }
      }

      \subcaptionbox{Step 3.\label{ccg:rotate:step:three}}{
          \scalingCCGRotate{
              \begin{forest}
              [\textit{S},phantom,circle,draw,scale=1.5,tikz={\node [ gray, fit=()(tr)(Mary)(reads)]{};}
                  [\textit{S/NP},phantom,ellipse,draw,scale=1.1
                      [\textit{S/(S{\bs}NP)},ellipse,draw,scale=1.1,name=tr
                          [{\textit{NP}\\Mary},fill=green,fill opacity=0.1,draw,align=center,tier=words,name=Mary]
                      ]
                      [{\textit{(S{\bs}NP)/NP}\\reads},fill=red,fill opacity=0.1,draw,align=center,tier=words,name=reads]
                  ]
                  [{\textit{NP}\\papers},fill=blue,fill opacity=0.1,phantom,draw,align=center,tier=words,name=papers]
              ]
              \end{forest}
          }
      }
      \subcaptionbox{Step 4.\label{ccg:rotate:step:four}}{
          \scalingCCGRotate{
              \begin{forest}
              [\textit{S},phantom,circle,draw,scale=1.5,tikz={\node [ blue, fit=()(tr)(Mary)(reads)]{};}
                  [\textit{S/NP},ellipse,draw,scale=1.1
                      [\textit{S/(S{\bs}NP)},ellipse,draw,scale=1.1,name=tr
                          [{\textit{NP}\\Mary},fill=green,fill opacity=0.1,draw,align=center,tier=words,name=Mary]
                      ]
                      [{\textit{(S{\bs}NP)/NP}\\reads},fill=red,fill opacity=0.1,draw,align=center,tier=words,name=reads]
                  ]
                  [{\textit{NP}\\papers},fill=blue,fill opacity=0.1,phantom,draw,align=center,tier=words,name=papers]
              ]
              \end{forest}
          }
      }

      \subcaptionbox{Step 5.\label{ccg:rotate:step:five}}{
          \scalingCCGRotate{
              \begin{forest}
              [\textit{S},phantom,circle,draw,scale=1.5,tikz={\node [ gray, fit=()(tr)(Mary)(papers)]{};}
                  [\textit{S/NP},ellipse,draw,scale=1.1
                      [\textit{S/(S{\bs}NP)},ellipse,draw,scale=1.1,name=tr
                          [{\textit{NP}\\Mary},fill=green,fill opacity=0.1,draw,align=center,tier=words,name=Mary]
                      ]
                      [{\textit{(S{\bs}NP)/NP}\\reads},fill=red,fill opacity=0.1,draw,align=center,tier=words,name=reads]
                  ]
                  [{\textit{NP}\\papers},fill=blue,fill opacity=0.1,draw,align=center,tier=words,name=papers]
              ]
              \end{forest}
          }
      }
      \subcaptionbox{Step 6.\label{ccg:rotate:step:six}}{
          \scalingCCGRotate{
              \begin{forest}
              [\textit{S},circle,draw,scale=1.5,tikz={\node [ green, fit=()(tr)(Mary)(papers)]{};}
                  [\textit{S/NP},ellipse,draw,scale=1.1
                      [\textit{S/(S{\bs}NP)},ellipse,draw,scale=1.1,name=tr
                          [{\textit{NP}\\Mary},fill=green,fill opacity=0.1,draw,align=center,tier=words,name=Mary]
                      ]
                      [{\textit{(S{\bs}NP)/NP}\\reads},fill=red,fill opacity=0.1,draw,align=center,tier=words,name=reads]
                  ]
                  [{\textit{NP}\\papers},fill=blue,fill opacity=0.1,draw,align=center,tier=words,name=papers]
              ]
              \end{forest}
          }
      }
      \subcaptionbox{Step 7.\label{ccg:rotate:step:seven}}{
          \scalingCCGRotate{
              \begin{forest}
              [\textit{S},circle,draw,scale=1.5,tikz={\node [ green, fit=()(tr)(Mary)(papers)]{};}
                  [\textit{S/(S{\bs}NP)},ellipse,draw,scale=1.1,name=tr
                      [{\textit{NP}\\Mary},fill=green,fill opacity=0.1,draw,align=center,tier=words,name=Mary]
                  ]
                  [\textit{S{\bs}NP},ellipse,draw,scale=1.1
                      [{\textit{(S{\bs}NP)/NP}\\reads},fill=red,fill opacity=0.1,draw,align=center,tier=words,name=reads]
                      [{\textit{NP}\\papers},fill=blue,fill opacity=0.1,draw,align=center,tier=words,name=papers]
                  ]
              ]
              \end{forest}
          }
      }

      \subcaptionbox{Step 8.\label{ccg:rotate:step:eight}}{
          \scalingCCGRotate{
              \begin{forest}
              [\textit{S},circle,draw,scale=1.5,tikz={\node [ gray, fit=()(tr)(Mary)(daily)]{};}
                  [{\textit{(S{\bs}NP){\bs}(S{\bs}NP)}\\daily},fill=orange,fill opacity=0.1,draw,align=center,tier=words,phantom]
                  [\textit{S/(S{\bs}NP)},ellipse,draw,scale=1.1,name=tr
                      [{\textit{NP}\\Mary},fill=green,fill opacity=0.1,draw,align=center,tier=words,name=Mary]
                  ]
                  [\textit{S{\bs}NP},ellipse,draw,scale=1.1
                      [{\textit{(S{\bs}NP)/NP}\\reads},fill=red,fill opacity=0.1,draw,align=center,tier=words,name=reads]
                      [{\textit{NP}\\papers},fill=blue,fill opacity=0.1,draw,align=center,tier=words,name=papers]
                  ]
                  [{\textit{(S{\bs}NP){\bs}(S{\bs}NP)}\\daily},fill=orange,fill opacity=0.1,draw,align=center,tier=words,name=daily,edge=white]
              ]
              \end{forest}
          }
      }
      \subcaptionbox{Step 9.\label{ccg:rotate:step:nine}}{
          \scalingCCGRotate{
              \begin{forest}
              [\textit{S},circle,draw,scale=1.5,tikz={\node [ purple, fit=()(tr)(Mary)(daily)]{};}
                  [\textit{S/(S{\bs}NP)},ellipse,draw,scale=1.1,name=tr
                      [{\textit{NP}\\Mary},fill=green,fill opacity=0.1,draw,align=center,tier=words,name=Mary]
                  ]
                  [\textit{S{\bs}NP},ellipse,draw,scale=1.1
                      [\textit{S{\bs}NP},ellipse,draw,scale=1.1
                          [{\textit{(S{\bs}NP)/NP}\\reads},fill=red,fill opacity=0.1,draw,align=center,tier=words,name=reads]
                          [{\textit{NP}\\papers},fill=blue,fill opacity=0.1,draw,align=center,tier=words,name=papers]
                      ]
                      [{\textit{(S{\bs}NP){\bs}(S{\bs}NP)}\\daily},fill=orange,fill opacity=0.1,draw,align=center,tier=words,name=daily]
                  ]
              ]
              \end{forest}
          }
      }
      \caption{Parsing steps for CCG Revealing. Each row indicates parser steps associated with a single word.}
      \label{fig:ccg:rotate}
\end{figure*}
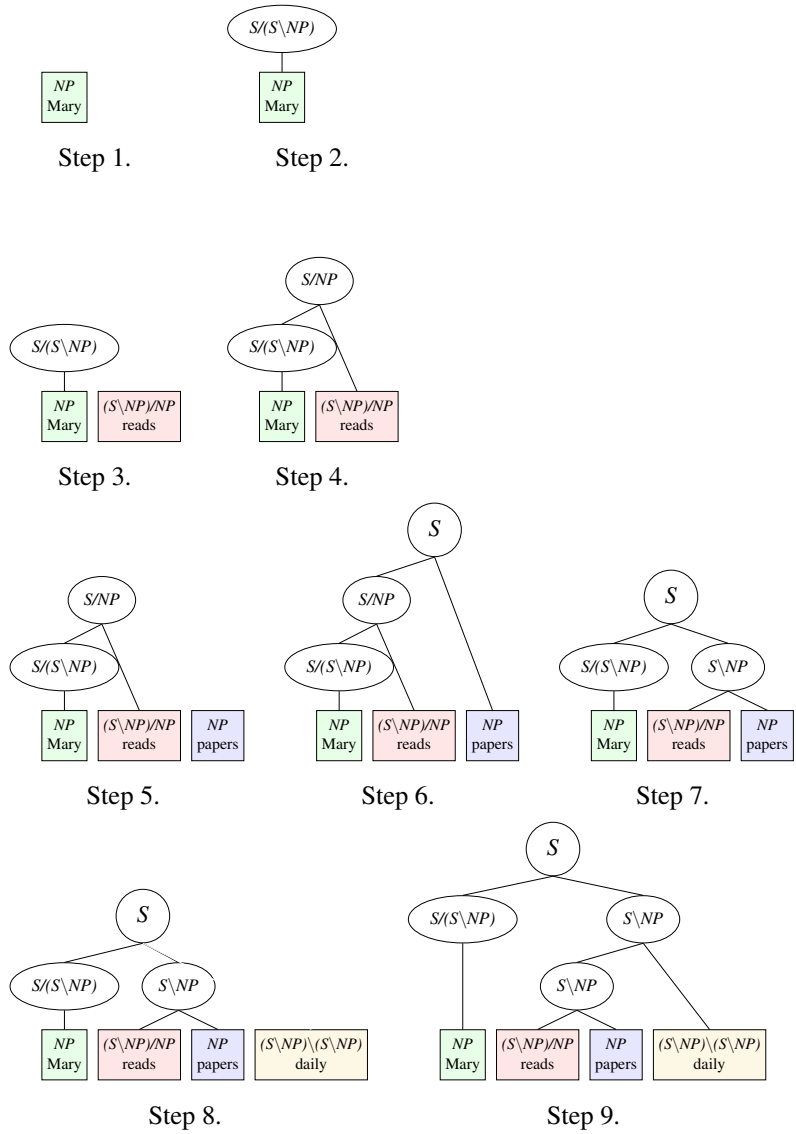

Just as in the CFG case, we measure parsing~effort in CCG by counting the operations that the parser needs to conduct for each word of the input sentence. We compute parsing effort for two different CCG parsing strategies: CCG left-branching (\ccgleft) and CCG-revealing (\ccgreveal). The trees over which we derive these CCG parsing predictors are obtained using \emph{Rotating-CCG} parser by \citet{stanojevicCCGParsingAlgorithm2019} which is a highly accurate parser that can 
recover labelled dependencies in English with $90.8\%$ accuracy. %

\subsubsection*{Modeling predictability with a Large Language Model}

To quantify the predictability of successive~words, we use \textit{Chinchilla} which is DeepMind's state-of-the-art large language model \citep{chinchilla:arxiv}. 
This is Transformer-based neural network with $70$ billion parameters that was trained on $1.4$ trillion tokens of text (close to $10$ TB of data).

As a predictor of processing difficulty of a probabilistic language model we use surprisal \citep{hale2001pep} which is
the negative logarithm of the probability of the next~word, given all the preceding words. 
More formally:

\begin{align}
    \operatorname{surprisal}(\operatorname{word}_i) = - \log_2 p(\operatorname{word}_i | \operatorname{word}_0 \dots \operatorname{word}_{i-1})
\end{align}

\noindent A technical~issue exists in that the probabilities provided by \textit{Chinchilla} are over tokens instead of words. 
Tokens are automatically constructed using SentencePiece \citep{kudo-richardson-2018-sentencepiece} which produces tokens that are not necessarily meaningful in a linguistic sense, i.e. they need not represent morphemes. 
To get word probabilities we 
sum the log-transformed probabilities of all the tokens that are extracted from the given word. 
If the $k$ tokens of word $\operatorname{word}_i$ start from token at position $f(\operatorname{word}_i)$, we compute surprisal for $\operatorname{word}_i$ as:

\begin{align}
    \operatorname{surprisal}(\operatorname{word}_i) = - \sum_{j=f(\operatorname{word}_i)}^{f(\operatorname{word}_i)+k-1} \log_2 p(\operatorname{token}_j | \operatorname{token}_0 \dots \operatorname{token}_{j-1})
\end{align}

\subsubsection*{Control Variables}

In addition to the model-derived predictors described above, we also created a set of control predictors to capture lower-level lexical and acoustic properties of the stimulus. 
For lexical properties, we use include log lexical frequency (\freq{}) derived from the SUBTLEX-US corpus \citep{Brysbaert:2009fk} along-side a simple predictor for word-offsets (\wordrate{}), following \citet{Brennan:2010ab}.
For acoustic properties, we include continuous measures of root-mean-squared power of the audiobook story (\rms{}) and the prosodic contour of narration in terms of the fundamental frequency of speech (\fzero{}) derived using the Praat software package \citep{Boersma:2001vn}.

\subsection{Participants \& Data} \label{sec:participants}

\begin{figure}
    \centering
    \includegraphics[width=\textwidth]{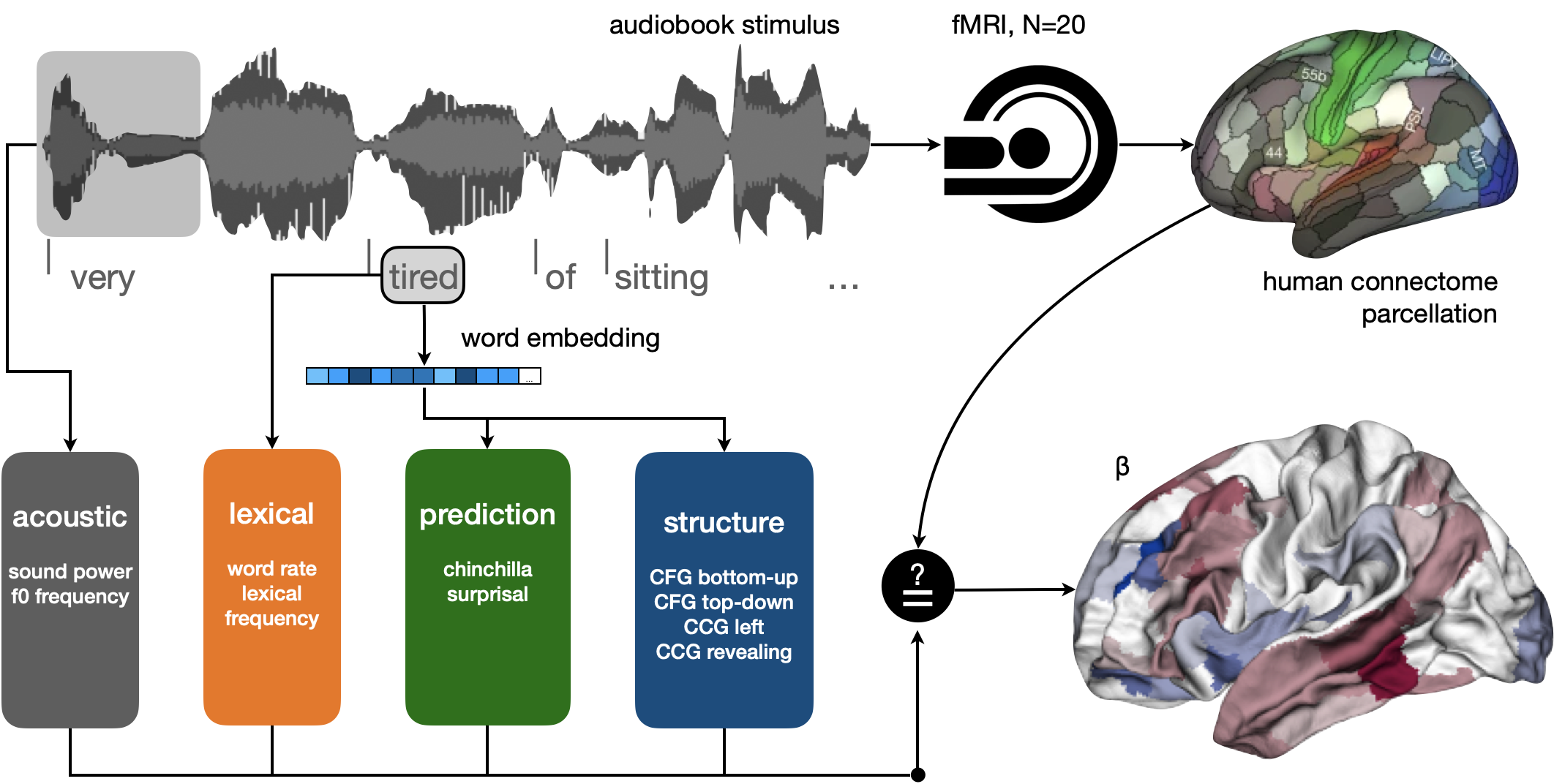}
    \caption{The data comprises N=20 fMRI time-series recorded while participants listened to an audiobook story. 
    Voxels were grouped according to the Human Connectome Project's multi-modal cortical parcellation and time-series were averaged within each region (top-right).
    The audiobook story was annotated for acoustic and lexical control variables along with state-of-the-art estimates of lexical prediction from the {\chinchilla} LLM and a family of structure-based predictors using both a CFG and the more expressive CCG formalism (bottom left).
    Hierarchical linear regression was used to evaluate the fit between linguistic predictors and neural time-series (bottom-right).
    }
    \label{fig:analysis}
\end{figure}

The data for this project come from the LPPC–fMRI corpus \citep{liPetitPrinceMultilingual2021}.
The full corpus comprises 112 fMRI datasets collected while participants listen to the audiobook \textit{The Little Prince} by Antoine de Saint-Exup{\'e}ry in their first language of either English ($N=49$), French ($N=28$), or Mandarin ($N=35$) (Figure \ref{fig:analysis}, top left).
In brief, participants listened on MRI-safe headphones to the audiobook story for about 90 minutes while lying supine for fMRI scanning.
The audiobook was presented across nine runs, each corresponding to a chapter and lasting about 10 minutes.
Images were acquired at 3T (MRI GE Discovery MR750).
Structural scans were acquired with a T1-weighted MPRAGE sequence; functional scans were acquired with a multi-echo planar sequence with parameters: 
    TR = 2 s, 
    TEs = [12.8 27.5 43] ms, 
    FA = $77^{\circ}$,
    matrix size = $72 \times 72$, 
    FOV = $240.0 \times 240.0$ mm,
    image acceleration =  $2 \times$.
The first four volumes of each run were discarded.
Functional data were denoised using multi-echo independent component analysis \citep{kunduDifferentiatingBOLDNonBOLD2012} prior to co-registering functional and structural volumes and normalization to the MNI atlas. 

We use the first 20 English-language datasets here (subject IDs 57--78).
The choice of language reflects the fact that English-language training resources available for the CCG parser \citep{hockenmaierCCGbankCorpusCCG2007}.
This particular subset of English-language dataserts was chosen randomly; the number of datasets seeks to balance balances broad-coverage against computational feasibility as detailed more below.
All of the selected participants showed high accuracy on the comprehension questions (Mean = 94\%, Range = [86\% 100\%]).

For each of these datasets, structural scans were parcellated into 360 cortical regions based on a volumetric preparation of the HCPMM1 atlas \citep{glasserMultimodalParcellationHuman2016} using the Nilearn package \citep{abrahamMachineLearningNeuroimaging2014}.%
\footnote{
    The volumetric parcellation was prepared by Dianne Patterson and it is available at \url{https://neuroimaging-core-docs.readthedocs.io/en/latest/pages/atlases.html}
}
Functional data from voxels within each parcel were averaged together yielding fMRI time-series for each participant for each of 360 regions spanning the whole cortex (Figure \ref{fig:analysis} top right).
These time-series were then converted to $z$-scores separately for each participant and for each scanner run.

These time-series were divided into training and test sets by extracting the first 400 fMRI samples per participant and per ROI for model training (1,404,000 datapoints per hemisphere spanning approximately 13.3 minutes of scan time, excluding the first 10 volumes and the first volume from each subsequent run) and samples 401--800 per participatn per ROI for out-of-sample testing (1,436,400 datapoints per hemisphere excluding the first volume from each run).

\subsection{Statistical Analysis} \label{sec:analysis}

\begin{figure}
    \centering
    \includegraphics[width=\textwidth]{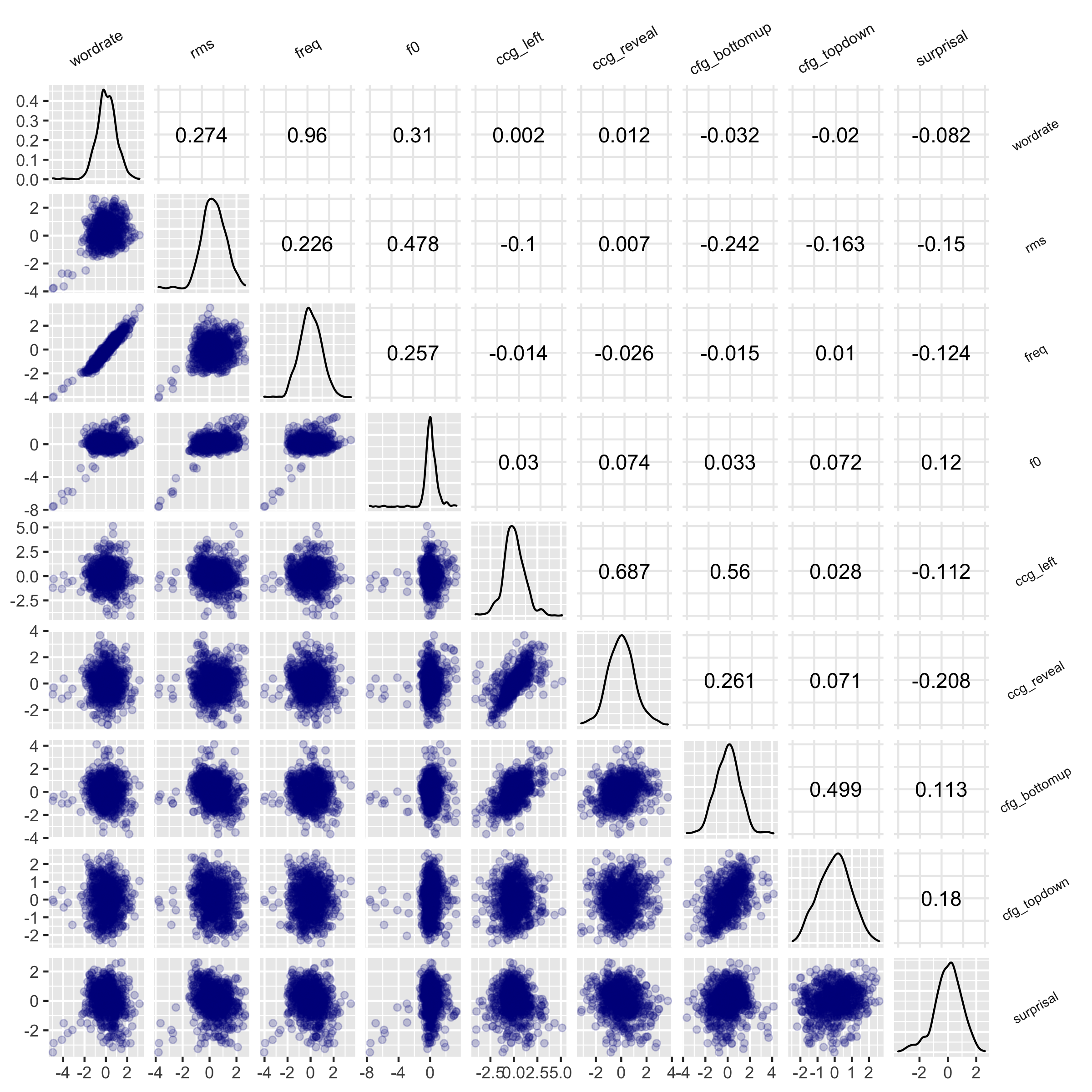}
    \caption{Distributions and bivariate correlates (Pearson's $r$) between all regressors entered into the statistical analysis.}
    \label{fig:predictors}
\end{figure}

To align the target and control predictors with the fMRI time-series, each predictor was convolved with the hemodynamic response function (modeled with the Nilearn library as a mixture of two gamma distributions) and re-sampled to 0.5 Hz to match the sampling rate of the data. 
For this operation, all predictors defined at the word level were represented as unit impulse functions time-aligned to word-offset.
The convolution operation increased colinearity between predictors as they all are influenced by the speech rate of the audiobook stimulus.
For each target regressor we isolated the orthogonal component of that term to the space spanned by the \wordrate{} control regressor and the constant vector 1.
Even after orthogonalization, the correlation between \topdown{} and \leftcorner{} was very high ($r > 0.95$) and so only one term, \topdown{}, was included in our analysis.
Figure \ref{fig:predictors} shows the bivariate distributions and correlation coefficients for all pairwise combinations of regressors that were entered into a statistical model against the fMRI time-series.

\subsubsection*{Fitting procedure} \label{sec:fittingprocedure}

We quantify the fit from target and control regressors to the fMRI time-series using a Bayesian multi-level linear regression fit with the Stan platform for statistical modeling via the \texttt{brms} package in R \citep{burknerAdvancedBayesianMultilevel2018, StanModelingLanguage2022}.
As schematically illustrated on the bottom of Figure \ref{fig:analysis}, this model was fit against the parcellated fMRI data set aside for training described above (first $\approx 13.3$ min per dataset); target and control regressors were included in the model as fixed effects along with random slopes by participant and for brain region \citep[see][]{Shain:2020qq}.
The statistical model was defined as given below using an extension of the Wilkinson-Rogers notation. 
(The term \texttt{fMRI} here indicates a time-series of $\approx 13.3$ m BOLD signal samples from 20 participants from 180 regions per hemisphere.)

\vskip 1em
\texttt{
\begin{tabular}{rcl}
   fMRI & $\sim$ &  \wordrate{} + \rms{} + \freq{} + \fzero{} + \\
        &   & \surprisal{} + \\
        &   & \bottomup{} + \topdown{} + \\          
        &   & \ccgleft{} + \ccgreveal{} + \\[1.5em]
       & & (0 + \wordrate{} + \rms{} + \freq{} + \fzero{} + \\
       & & \surprisal{} + \\
       & & \bottomup{} + \topdown{} + \\
       & & \ccgleft{} + \ccgreveal{} || subject) + \\[1.5em]
       & & (0 + \wordrate{} + \rms{} + \freq{} + \fzero{} + \\
       & & \surprisal{} + \\
       & & \bottomup{} + \topdown{} \\
       & & \ccgleft{} + \ccgreveal{} || region) 
\end{tabular}
}

\vskip 1em

Regressors were mean-centered prior to being entered into the model; note that because the data were $z$-scored, therefor also centered, we simplify the modeling by excluding by-participant and by-region intercepts. 
We also simplify by not estimating the covariances between random terms.
We specified weakly informative priors, including $\mathcal{N}(0,1)$ for the model intercept, $\mathcal{N}(0, 2.5)$ for all other fixed effect parameters; and an exponential distribution with a rate of 1 for the standard deviation of hierarchical terms.
Priors for all other terms conformed to defaults for \texttt{brms} 2.17.0.
The posterior distribution was sampled with four Markov chains, each comprising 2,000 samples (1,000 for warm-up).

We two such models, one per hemisphere, using 32 cores (3.0 GHz Intel Xeon Gold 6154) on the ``Great Lakes'' high-performance computing cluster at the University of Michigan.
The fitting procedure made use of within-chain parallelization across 8 threads per-chain and QR decomposition was applied to the population-level design matrix; fitting took approximately 6 days to complete per hemisphere.
Model diagnostics did not indicate any problems with the procedure: there were no divergent transitions and we observed good mixing between chains (all $\hat{R} < 1.01$).

\subsubsection*{Statistical inferences} \label{sec:inferences}

Inferences were conditioned on the fitted models in two ways.
First, we examined the posterior distribution of the regression coefficients ($\beta$) for each target effect per region (see Figure \ref{fig:analysis} bottom right).
Second, we also evaluate how each term contributes to the \textit{out-of-sample goodness of fit} per region in the following way.%
\footnote{
    The approach to out-of-sample goodness of fit we describe offers two practical advantages in the present case over alternatives methods for comparing terms such as approximate leave-one-out cross-validation \citep{Vehtari:2017mz}.
    First, we can evaluate goodness-of-fit per region by restricting attention to $y_i - \hat{y_i}$ where $i$ spans just datapoints from a specific region.
    Second, our method does not require refitting the model for each ablation which would be computationally costly.    
    }
We define a measure of goodness-of-fit using the root-mean-squared error, $\text{RMSE} = \sqrt{\frac{1}{n} \sum_{i=1}^{n} (y_i - \hat{y_i})^2}$, 
where $y$ comes from the $n$ held-out testing samples of fMRI data per participant and $\hat{y}$ is the estimated value for each held-out out sample using 500 draws of the posterior distribution of coefficients. 
We compute {\rmse[full]} for the model with all terms, and iteratively, {\rmse[term]} where each regressor is ablated by applying a circular shift by $\frac{n}{2}$ to each predictor vector of length $n$ prior to computing $\hat{y}$.
This disrupts the contribution of each term to the model out-of-sample fit by removing temporal alignment between predictor and data. 
From this, we derive the change in goodness-of-fit that can be uniquely ascribed to each term:
 $\Delta RMSE^{term} = RMSE^{term} - RMSE^{full}$.
 This value is higher when a particular term has a greater impact on out-of-sample goodness-of-fit in a particular region.

We recognize as ``statistically reliable'' results where the following two conditions hold: 
    (i) the 99\% credibility interval (CI) of the $\beta$ coefficient posterior distribution excludes zero, 
    and (ii) the 99\% CI of the {\deltarmse} distribution excludes zero.
{\deltarmse} also offers a means to directly compare which of two terms offer a ``better fit'' to a given region. 
For a pair of terms {\deltarmse[target]} and {\deltarmse[control]} we evaluate whether the posterior distribution of one is reliably greater than the the other.  
This condition is met when the 99\% credibility interval of {\deltarmse[target]} $-$ {\deltarmse[control]} excludes zero and the above conditions (i--ii) hold for the target term.
This quantifies where held-out data are better captured when one term is added to the model over-and-above another.

\section{Results} \label{sec:results}

\subsection{Validation Checks}

\begin{figure}
    \centering
    \includegraphics[width=\textwidth]{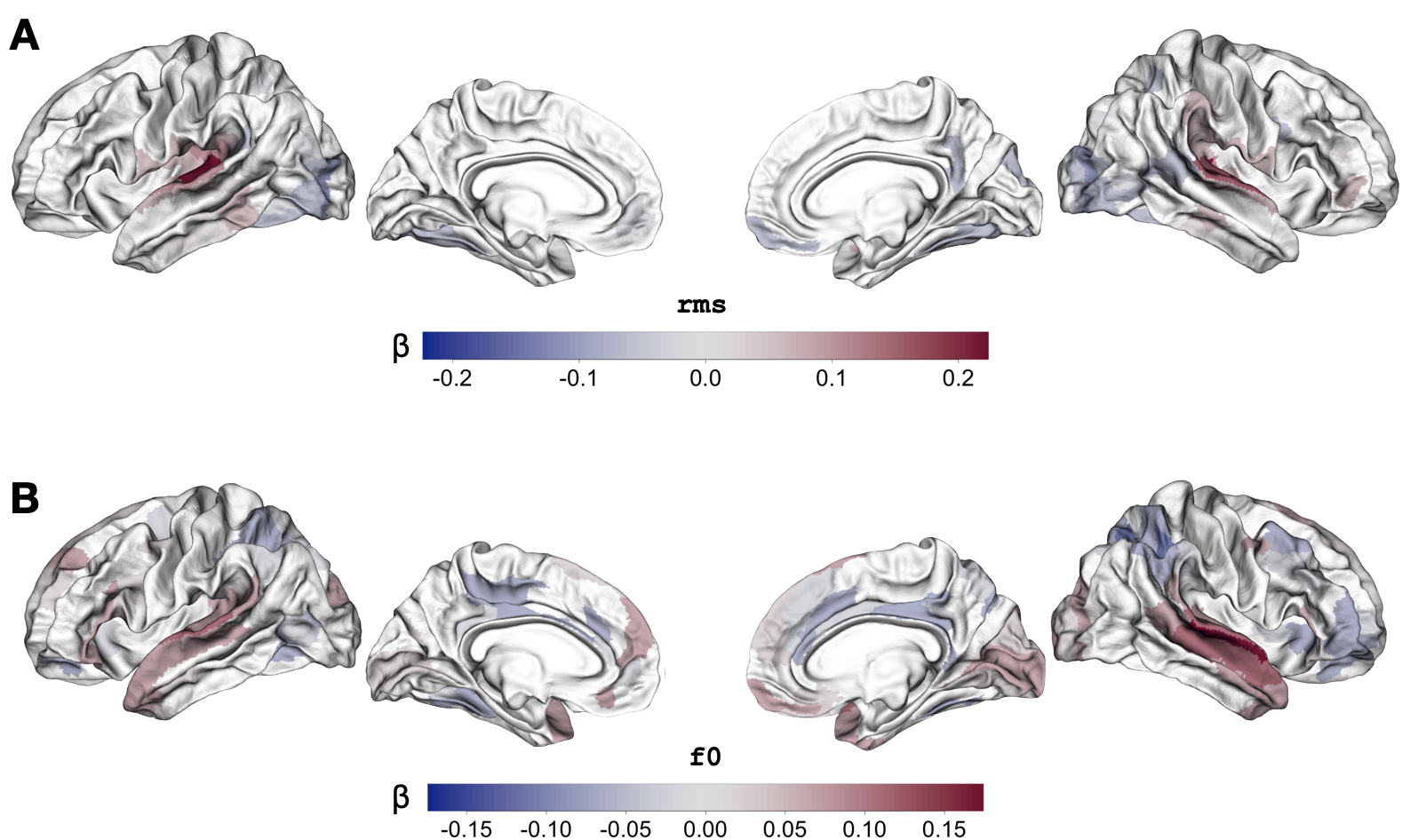}
    \caption{Posterior mean of the regression coefficients ($\beta$) for several control variables; regions are colored where both {\cibeta} and {\cirmse} exclude zero.
    (A) Root-mean square sound power localizes to the primary auditory cortex bilaterally. 
    (B) fundamental frequency, or F0, correlates with superior temporal regions and shows a right hemisphere bias.
    Together, these results show a concordance between our approach and several familiar effects.}
    \label{fig:results-controls}
\end{figure}

Before turning to our principle questions we first validate our procedures by checking that control predictors yield familiar results.
The sound power of the audiobook stimulus, {\rms}, should drive neural activity in primary auditory regions, and that is exactly what we observe in the results shown in Figure \ref{fig:results-controls}A.
We also see similar confirmatory results for sentence prosody via the \fzero{} control predictor (Figure \ref{fig:results-controls}B) which shows a reliable positive correlation with superior temporal activity bilaterally with a right-hemisphere bias; this matches other fMRI investigations of sentence prosody \citep[e.g.][]{Colin-Humphries:2005nx}.

\subsection{CCG-based parser metrics outperform CFG metrics}

\begin{figure}
    \centering
    
    \includegraphics[width=\textwidth]{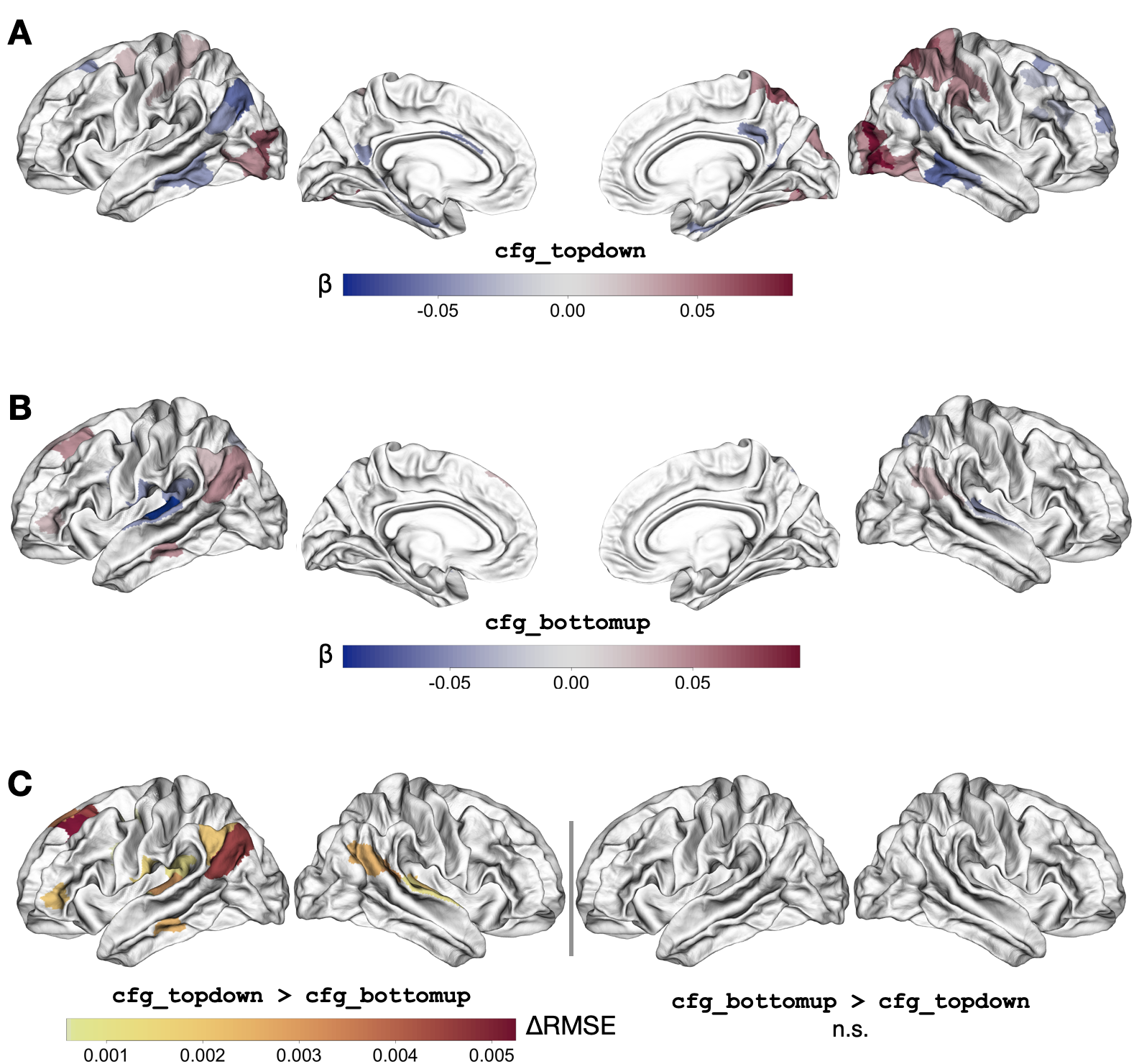}

    \caption{Posterior mean of regression coefficients ($\beta$) for parser-step predictors derived from a CFG; 
    panel (A) shows results for a top-down strategy and panel (B) for a bottom-up parser strategy.
    Regions are colored where both {\cibeta} and {\cirmse} exclude zero. 
    Panel (C) shows the direct comparison of these two predictors in terms of  {\deltarmse}; mean differences between predictors are plotted where $\geq 99\%$ of the posterior distribution is above zero and {\cibeta} for the first term excludes zero.
    Top-down parser steps shows improved fit to the data compared to  bottom-up predictor in left middle temporal, temporo-parietal and superior frontal regions.
    There are no areas where the bottom-up predictor out-performs the top-down predictor.
    }
    \label{fig:results-cfg}
\end{figure}

We now turn to the primary analyses of interest, beginning with the question of whether predictors derived from the more human-like CCG parser, operationalized with {\ccgleft} and {\ccgreveal}, better capture neural signals compared to the CFG parser steps used in prior work, operationalized as {\bottomup} and {\topdown}.
Figure \ref{fig:results-cfg} shows the fitted values for the two CFG predictors along with a comparison between them. 
On their own terms, these results accord with previous studies of parser steps (e.g. \citealp{Bhattasali:2018fk,Brennan:2016yu}), including the statistical comparison favoring the more eager of the two strategies (Figure \ref{fig:results-cfg}C; but cf. the electrophysiological results of \citealp{Nelson:2017kq}).
Note that the two comparisons of \deltarmse{} plotted in panel C and in subsequent figures are not symmetrical: each comparison plot is masked based on the posterior distribution of the target regression coefficient.
For example, \deltarmse{} values shown on left-hand side of Figure \ref{fig:results-cfg} are only plotted for regions where the 99\% of the posterior distribution for the \topdown{} coefficient excludes zero.

\begin{figure}
    \centering
    
    \includegraphics[width=\textwidth]{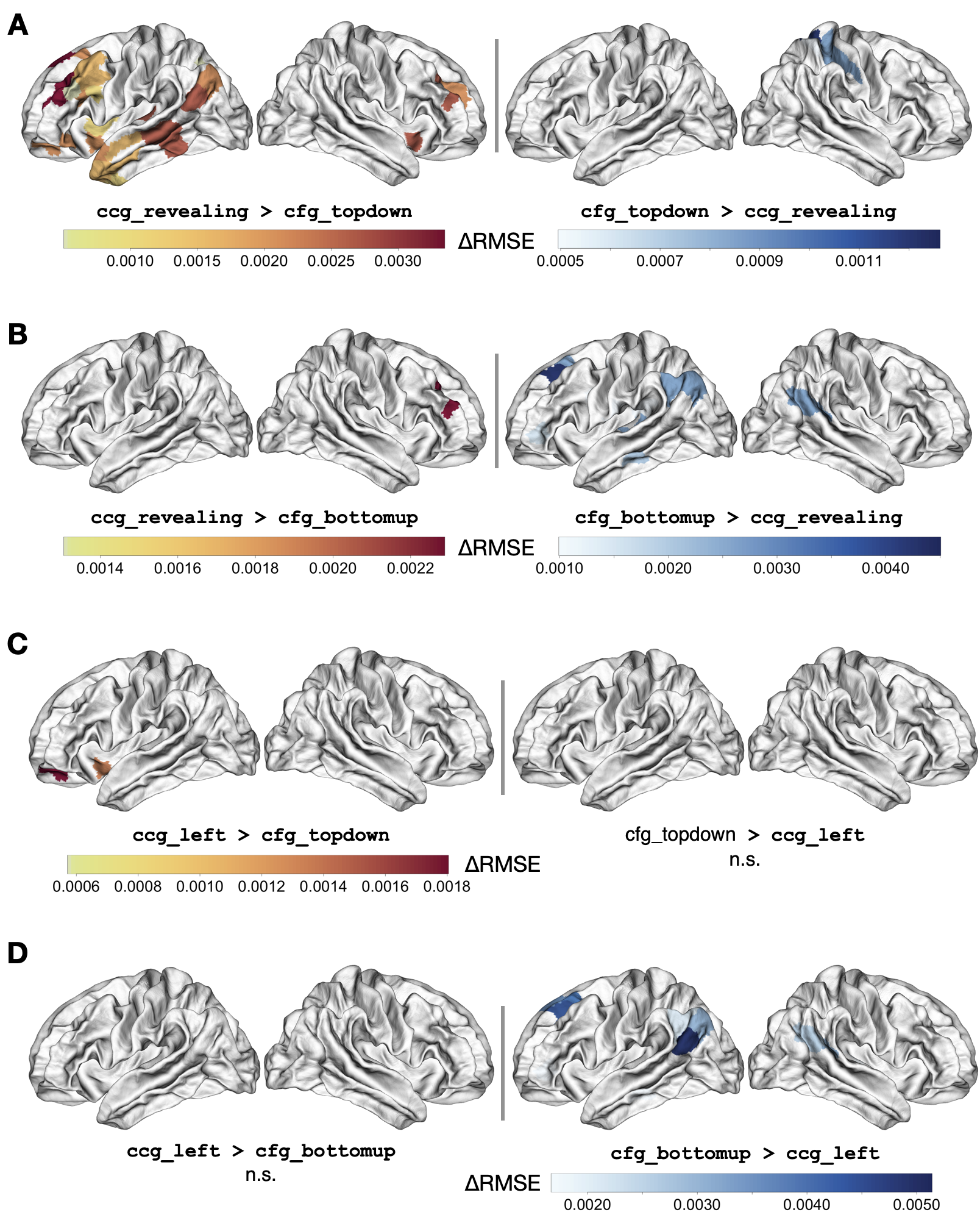}

    \caption{Comparisons between CCG and CFG-derived poser step predictors in terms of the relative sample goodness-of-fit quantified with {\deltarmse}. 
    Each panel plots the pairwise mean differences between predictors where $\geq 99\%$ of the posterior distribution is above zero and {\cibeta} for the first term excludes zero.
    CCG, especially with the {\reveal} operation, outperforms CFG-based models along anterior and posterior regions of the left temporal lobe and the frontal lobes bilaterally.
    Bottom-up and top-down CFG-based predictors show relative improvements compared to CCG in temporo-parietal regions bilaterally and along the left superior frontal gyrus.
    The CCG-based predictors show an overall greater improvement in model fit than the CFG models (see main text).}
    \label{fig:results-ccg-vs-cfg}
\end{figure}

Our focus, however, is on the comparison of these predictors to those derived from the CCG parsers; those fitted values are summarized in Figure \ref{fig:results-ccg} and the comparison between CCG and CFG predictors is shown in Figure \ref{fig:results-ccg-vs-cfg}.
CCG predictors, especially with the \reveal{} operation, show a strong statistical fit with left perisylvian regions extending to the temporo-parietal junction (Figure \ref{fig:results-ccg}). 
This pattern is also observed when comparing the relative contribution to out-of-sample model fit for each pair of CCG and CFG predictors (Figure \ref{fig:results-ccg-vs-cfg}, especially panel A). 
When they are ablated, CCG-derived predictors show a more reliable impact on model-fit in fronto-temporal regions in comparison to the {\topdown} predictor. 
On the other hand, we see some evidence that the {\bottomup} predictor has a greater impact on the fit with fMRI activity in bilateral temporo-parietal regions (see right-hand sides of Figure \ref{fig:results-ccg-vs-cfg}B,D.)

This complex pattern of results points to an important nuance in our modeling effort: processing strategies need not be mutually exclusive. 
This recognizes that interpreting sentence-structure is multifaceted and draws on a range of neural circuits for which the models we are considering are partial estimators (see \citealp{COGS:COGS12445} for some discussion.)
Still, to offer a more global comparison of these two classes of models, we sum \deltarmse{} across all cortical regions for the two CFG terms and two CCG terms together. 
This yields a distribution estimating the effect of the family of terms on overall model fit.
Overall, CCG predictors have a greater impact on quality of fit (M = 0.489, Range = [0.408 0.563] than CFG terms (M = 0.36, Range = [0.299 0.399]).

\subsection{The {\reveal} operation improves the match between CCG and the brain}

\begin{figure}
    \centering
    
    \includegraphics[width=\textwidth]{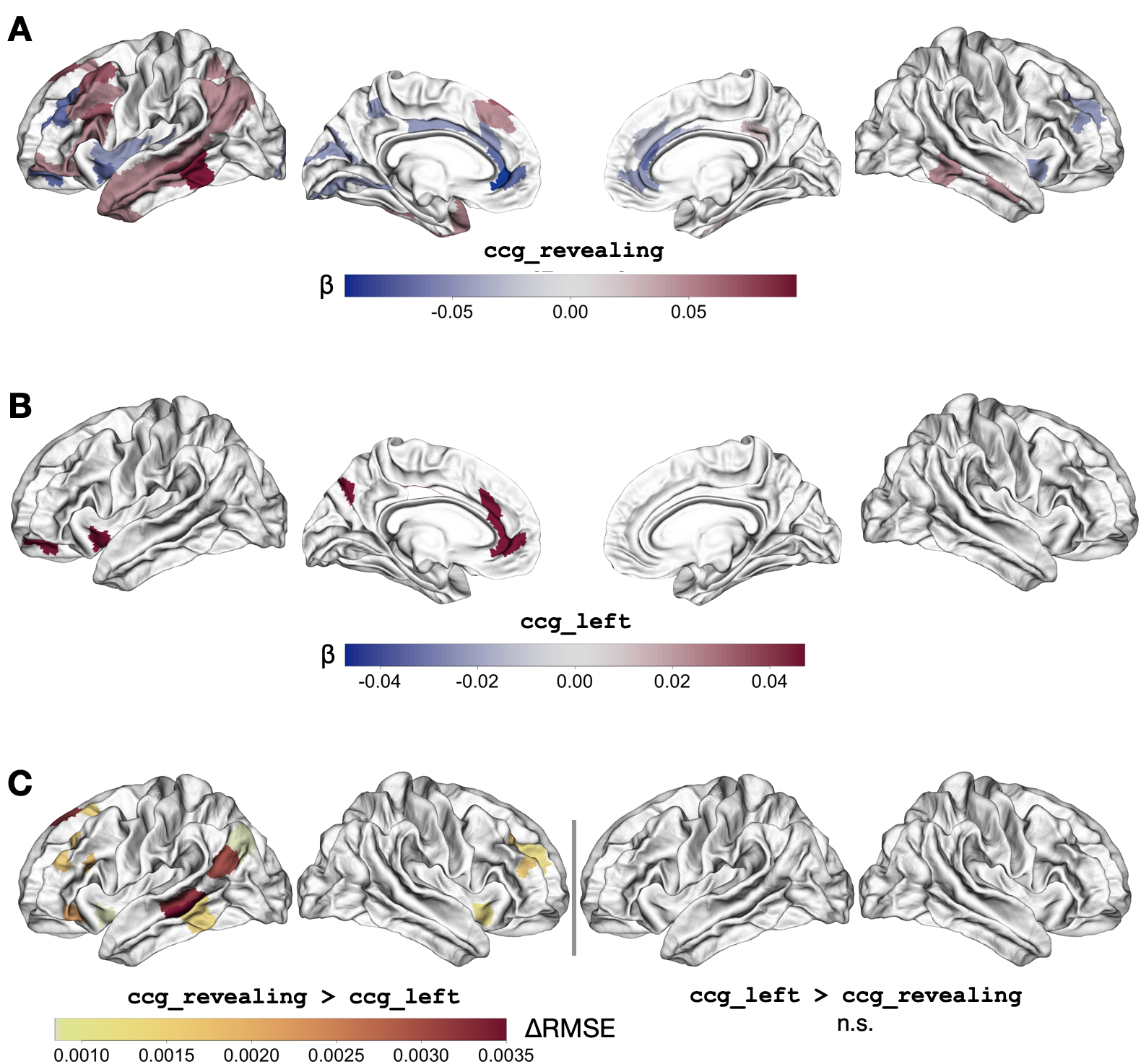}

    \caption{Posterior mean of the regression coefficient ($\beta$) where both {\cibeta} and {\cirmse} exclude zero for the CCG-derived parser step predictors; panel (A) shows results for a predictor derived with the {\reveal} operation, and panel (B) shows results for a left-branching strategy without that operation. 
    Panel (C) shows the direct comparison of these two predictors in terms of  {\deltarmse}; mean differences between predictors are plotted where $\geq 99\%$ of the posterior distribution is above zero and {\cibeta} for the first term excludes zero.
    The CCG-based predictor incorporating the {\reveal} operation leads to improved model fits in left posterior temporal, and bilateral frontal regions; there are no regions where the absence of {\reveal} reliably improves model fit.}

    \label{fig:results-ccg}
\end{figure}

Next we turn to which formulation of the CCG parser offers the best fit to these fMRI data.
The measure of parser steps derived when the {\reveal} operation is included, {\ccgreveal}, reliably correlates with a range of left temporal, inferior frontal, and temporo-parietal regions, as shown in Figure \ref{fig:results-ccg}A.
The pattern of results is qualitatively different than what we observe with the {\ccgleft} predictor that is derived without the {\reveal} operation, which shows instead reliable fits with ventral and medial frontal regions (Figure \ref{fig:results-ccg}B).

A direct comparison of how each term affects goodness-of-fit against out-of-sample data shows that ablating {\ccgreveal} has the largest impact on posterior temporal regions, more so than {\ccgleft}.
There are no areas where the reverse pattern holds.
These comparisons are shown Figure \ref{fig:results-ccg}C and they favor the {\reveal} operation of \citet{stanojevicCCGParsingAlgorithm2019} in as much as it delivers a sequence of parser steps that better matches human left frontal and temporal hemodynamic signals recorded during sentence processing.

\subsection{CCG-based parser steps capture fMRI data independently of effects for predictability }

The above results are all observed in a statistical model that ``controls'' for predictability by including surprisal from the \chinchilla{} LLM as a co-regressor.
We now turn to a direct comparison of the goodness-of-fit offered by surprisal in comparison to CCG-derived parser steps.

\begin{figure}
    \centering
    \includegraphics[width=\textwidth]{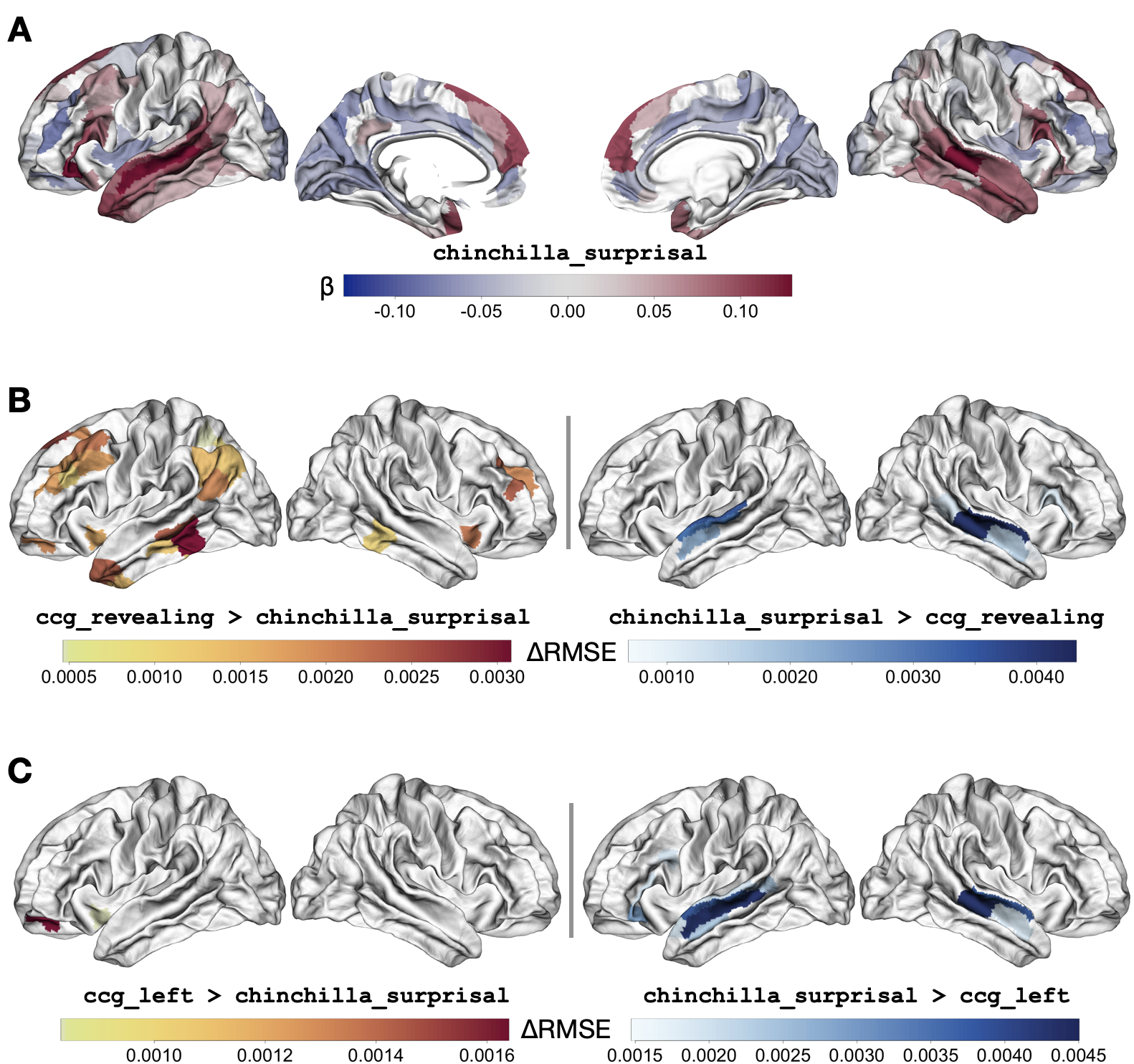}
    \caption{
    (A) Posterior mean of the regression coefficient ($\beta$) for the surprisal predictor derived from the {\chinchilla} LLM; regions are colored where both {\cibeta} and {\cirmse} exclude zero.
    Direct comparisons between surprisal and CCG-derived parser steps are given for 
    {\ccgleft} (B) and 
    {\ccgreveal} (C) 
    in terms of {\deltarmse}; mean differences between predictors are plotted where $\geq 99\%$ of the posterior distribution is above zero and {\cibeta} for the first term excludes zero.
    On its own terms surprisal from {\chinchilla} correlates with temporal and frontal activity bilaterally.
    In comparison, a CCG-derived predictor with {\reveal} shows improved model fits in posterior middle temporal, anterior temporal, superior frontal, and temporo-parietal areas. 
    On the other hand, {\chinchilla} shows improvements in model fit in the superior temporal gyrus bilaterally compared to CCG-derived predictors.
    }
    \label{fig:results-surprisal}
\end{figure}

Surprisal from a LLM, \surprisal, reliably correlates with a broad range of bilateral fonto-temporal activity, with the strongest patterns observed in the superior temporal gyri (Figure \ref{fig:results-surprisal}A).
This result matches previous efforts that specifically compare surprisal from
large language models with human neural~signals \citep{heilbronHierarchyLinguisticPredictions2022, caucheteuxEvidencePredictiveCoding2023}, and also studies drawing on alternative methods for quantifying word-by-word expectations \citep{Lopopolo:2017kq,Henderson:2016yq,Shain:2020qq,Willems:2015yu}.

We check for linearly independent contributions of \surprisal{} in comparison to the \ccgleft{} and \ccgreveal{} terms by evaluating how removing each affects out-of-sample prediction via \deltarmse{}. 
Indeed, we see that removing \ccgreveal{} reliably impacts goodness-of-fit in several regions including the left posterior temporal cortex, left anterior temporal pole, and left temporo-parietal regions to a degree that is reliably greater than any impact of \surprisal{}.
This comparison is illustrated on the left-hand side of Figure \ref{fig:results-surprisal}B.
The reverse comparison, shown on the right-hand side of Figure \ref{fig:results-surprisal}B, reveals unique contributions of \surprisal{} across the superior temporal gyri and superior temporal sulci bilaterally.
The impact of predictability on superior temporal regions is also evident in the comparison to \ccgleft{} shown on the right side of Figure \ref{fig:results-surprisal}C.
The latter CCG predictor shows some evidence of independent contributions in ventral frontal regions, but this result should be interpreted cautiously given the overall better performance of \ccgreveal{} discussed above.

The broader pattern evident in these comparisons reinforces the complementary nature of expectation-based measures like \surprisal{} and measures tied more directly to structural complexity like \ccgreveal{}.
Both classes of predictors reliably and independently correlate with the fMRI signals and their effects are spatially distinct: the former is most prominent on bilateral superior temporal regions, while the latter is associated with inferior frontal, posterior temporal, and anterior temporal foci.

\section{Discussion} \label{sec:discussion}

In this study we examine how well computational~models of parsing~effort capture variance in hemodynamic neural signals collected while participants listened to an audiobook story. 
The models we deploy span dimensions that have been debated in prior~work in computational psycholinguistics and neurolinguistics: 
    We compare parsing models based on a context-free grammar (CFG) with those based on a mildly-context sensitive grammar (CCG), 
    and we also compare parsing models that differ in how eagerly they postulate new structure. 
For the CCG parsers, this includes testing the \reveal{} operation as to whether it yields a more ``human-like'' account of incremental parsing.
These comparisons are evaluated in the context of control covariates which include a state-of-the-art estimate of next-word predictability from the \chinchilla{} large language model (LLM). 
The comparisons reported here carry particular significance for cognitive theories of language processing based on artificial neural networks,
whose operation relate directly to human brain signals \citep[e.g.][]{schrimpfNeuralArchitectureLanguage2021,caucheteuxBrainsAlgorithmsPartially2022}.
They also speak to the theoretical possibility that structural~complexity might only impact incremental processing through the ``bottleneck'' of predictability \citep[e.g.][]{Levy:2008vh}.

\subsection{General considerations in reasoning about computational models of language processing}

Before discussing the results in detail, we first briefly consider the kind of reasoning about cognition that is supported by the modeling approach we adopt.
The overall %
approach
fits~into a tradition that seeks to address %
which computational~model of language~processing best~describes the human~system \cite[e.g.][]{kaplan72,kimball73,berwick84,frazier96,Steedman:2000yq}.
\citet{brennanNaturalisticSentenceComprehension2016} %
 lays~out in very general terms how this traditional approach applies to neural~signals that have been recorded during naturalistic~listening.
In the terminology of that review, the present~work is an ``L-study.''  L-studies test linguistic and/or psycholinguistic theories by
comparing different~models against the same empirical~measure of brain~activity. 
All models meet a basic adequacy condition
of ``processing,'' ``generating'' or in some way ``analyzing'' the same text stimulus so the meaning of the stimulus is held constant.

In fact the present work reports two closely related L-studies, each examining a different dimension.
The first dimension is the grammar i.e. Combinatory Categorial Grammar (CCG), na\"ive phrase structure (CFG) or no explicit grammar at all (LLM).
The second is the parsing~strategy, which only makes~sense in the first two cases but not the third. 
These alternatives are detailed above in subsection~\ref{sec:computationalmodeling}.
The form of the claims to be discussed in this section is always the same:
because model $M'$ fits the brain~data better than $M$, the linguistic or psycholinguistic idea reified in $M'$ but not $M$ is supported.
Obviously, this sort of inference is limited to models whose consequences on the stimulus text can actually be calculated.
Perhaps less-obvious is the absence of any sort of exhaustivity claim.  Rather than excluding classes of possible models,
the methodology delivers relative rankings of actual models. For this reason it is important to prioritize,
as we have done here,
models that already enjoy some degree of plausibility in subdisciplines such as Linguistics and Artificial~Intelligence.

\subsection{Mildly-context sensitive CCG and the \reveal{} operation improve fit to neural signals}

We turn first to the comparison of different parsing models in terms of their fit to hemodynamic data. 
In this effort, we build on prior work using fMRI \citep{Brennan:2016yu,Brennan:2020ku,Henderson:2016yq,Shain:2020qq,Wehbe:2014la,reddyCanFMRIReveal2021,Bhattasali:2018fk} and electrophysiology \citep{Nelson:2017kq,Brennan:2017db,P18-1254}; that work itself is situated within a broader literature that aims to narrow down the neural subsystems uniquely engaged in sentence-level combinatoric processing \cite[for reviews, see][]{Zaccarella:2017sy,matchinCorticalOrganizationSyntax2020,pylkkanenNeuralBasisCombinatory2019}.
The key theoretical issue is whether the more~expressive CCG, which better-matches
human languages, derives improved~fits to fMRI~signals relative to a less~expressive CFG predominantly used in prior~work.

Models using the mildly context-sensitive CCG grammar show an overall better fit to neural signals, especially spanning the left frontal and temporal lobes, in comparison to models based on a CFG. 
This result is perhaps best illustrated in Figure \ref{fig:results-ccg-vs-cfg}A which shows the comparison between \ccgreveal{} and \topdown{}.
This particular comparison is valuable, as those two models each showed the best fit to neural signals in comparison to others based on the same grammar (see Figures \ref{fig:results-cfg} and \ref{fig:results-ccg}).

The improvement in fit for the CCG parsers obtains in spite of the fact that the accuracy of the CCG parser, while high on its on terms, is lower than that of the CFG parser that we use (CCG accuracy: 90.8\% \citealp[Table 4]{stanojevicCCGParsingAlgorithm2019}; CFG accuracy 95.9\% \citealp[Table 1]{kitaev-etal-2019-multilingual}.)

This result affirms the ``hidden consensus'' that human language syntax is built on a formal system with such expressivity \citep{Joshi:1985fk,Stabler:2013zp}.
It also is consistent with one prior~study comparing na\"ive phrase~structure to X-bar structures derived from a more expressive Minimalist~Grammar \citep{Brennan:2016yu}.
That earlier work was limited insofar as the grammar was not broad-coverage.
The current~effort takes advantage of CCGbank~\citep{hockenmaierCCGbankCorpusCCG2007}.
The implicit CCGbank grammar is broad-coverage and seems to generalize in a psychologically-realistic way to our stimulus text, an English translation of \\ \underline{The Little Prince}.
\citet{Shain:2020qq} and \citet{Brennan:2020ku} also deploy grammars that might in~principle be as expressive as CCG, but in~practice both of those efforts ultimately rely on CFG-equivalent variants.

We note, again, that the conclusions we draw from the present~data are limited to the ``small world'' of the models under consideration. 
Other instantiations of both context-free or mildly context-sensitive formalisms may yield different fits to neural time-series.
The present effort does not allow for statistical generalization beyond the specific models under analysis.
However, we contend that the present comparisons are well motivated: 
The bulk of prior neurolinguistic work in this domain has relied on CFGs of the type we deploy, and CCG is especially useful in our~effort because its design-principles reflect direct and incremental composition which accords with human sentence processing \citep{Steedman:2000yq}.

Comparison between models also addressed the dimension of eagerness: how readily is structure postulated based on partial input?
Of special interest here is the utility of the \reveal{} operation for CCG parsing developed by \citet{stanojevicCCGParsingAlgorithm2019}.
That operation was introduced to resolve a challenge in efficiently incremental parsing for optional elements such as right adjuncts. 
Our analysis suggests that the sequence of parsing steps when this operation is included in the model yields a better match to the dynamics of neural activity than when this operation is not included. 
This pattern was most striking in left posterior temporal regions (Figure \ref{fig:results-ccg}), in accordance with theoretical accounts positing a special role for that region in combinatoric processing \citep{matchinCorticalOrganizationSyntax2020} and recent intracranial recordings that test for sensitivity to linguistic phrase~composition \citep{murphyMinimalPhraseComposition2022}.

\subsection{Structural complexity and next-word predictability capture complementary neural signals}

This study also aims to tease apart structural processing, operationalized here as parser~steps, from predictability. 
The import of latter has been studied extensively in the cognitive sciences \citep[][offer a broad perspective]{delangeReconstructingPredictiveArchitecture2022} and has been of central importance in the study of sentence comprehension \citep[see][for a review]{Traxler:2014qf}.
Crucially, structural complexity has been argued to modulate usage and, in turn, linguistic expectations derived from usage; 
under a strong formulation, structural complexity might only affect processing via the ``bottleneck'' of predictability (\citealp{Levy:2008vh} \citealp[cf.][]{Hawkins2014}).
Thus, a theoretical question is whether direct indices of structural complexity can be teased apart from word-by-word expectations within a naturalistic language comprehension task. 

To meet that question we use the strongest-to-date estimator for next word predictability, derived from the \chinchilla{} neural network LLM \citep{chinchilla:arxiv}. 
Against that baseline, CCG-derived structural~complexity reliably correlates with activity in the left middle temporal gyrus,  left temporal pole, angular gyrus and superior frontal gyrus (see especially Figure \ref{fig:results-surprisal}B.)

We observe complementary activation for predictability, operationalized as \\ \surprisal{}. 
That activation is observed, linearly independent of any effects for structural complexity, in the superior temporal gyrus bilaterally (right-hand panels in Figures \ref{fig:results-surprisal}B,C).
This result accords with the striking match between LLM-estimated predictability and neural signals reported by \citet{heilbronHierarchyLinguisticPredictions2022} and \citet{caucheteuxEvidencePredictiveCoding2023} for single-sentence comprehension. 
The bilateral temporal localization also matches reports from other efforts that examine more naturalistic comprehension using alternative methods to estimate predictability such as $n$-gram language models \citep{Willems:2015yu,Lopopolo:2017kq,Brennan:2016yu,Shain:2020qq}.

While these comparisons between complexity and predictability are consistent with predictability having a (relatively) focal effect in comprehension, we caution against drawing an overly strong conclusion here. 
For one, we expect based on both theoretical and empirical arguments that next-word predictions will correlate, at least to some extent, with structural~complexity and other factors (such as word-frequency).
Within this context, the particular analysis just mentioned aims to isolate only those spatial correlates that are linearly independent among the effects included in our models.
Indeed, areas where fMRI activity shows a reliable correlation with \surprisal{} along-side other factors include the temporal lobes broadly, temporal parietal junction, inferior frontal gyrus and posterior superior frontal areas (Figure \ref{fig:results-surprisal}A). 
At the least, our data are consistent with predictability have a large-scale modulatory effect on multiple, spatially distinct, facets of language comprehension.

\subsection{On parsimony}

Our analyses do not include predictors derived from the internal state of a LLM \citep[cf.][]{schrimpfNeuralArchitectureLanguage2021,kumarReconstructingCascadeLanguage2022,caucheteuxBrainsAlgorithmsPartially2022}.
That choice is driven by a focus on interpretability and parsimony in our modeling efforts; estimators for structural complexity are derived from the internal states of psycholinguistically plausible parsing models by counting parser steps.
Deep neural networks play a role in instantiating the parsing models, both for the CCG parser used here \citep{stanojevicCCGParsingAlgorithm2019}
and the estimates of predictability from \chinchilla{} \citep{chinchilla:arxiv}.
To pursue a more direct comparison between interpretable parsing models and the internal states of LLMs opens up the question of how to balance the statistical fit between a model to neural signals against the complexity or parsimony of that model.

LLMs are, famously, not parsimonious. 
\chinchilla{} has 70 billion parameters, which is substantially smaller than other state-of-the-art LLMs of the same class (GPT-3 has 175 billion; Gopher 280 billion etc.)
The size of these models is only exceeded by the size of the training data used to fit them: \chinchilla{} is trained with 1.4 trillion language tokens.%
\footnote{To put this in perspective, to have equivalent linguistic experience a 10 year old child would need to be exposed to over 4,000 words per second every hour of every day.
\citep[see][for large-sample and cross-cultural study of naturalistic linguistic input]{bergelsonEverydayLanguageInput2022}.}
The CCG parser used here has $\approx 0.1$ million trainable neural network parameters and is trained on $\approx 0.9$ million tokens \citep[CCGbank]{hockenmaierCCGbankCorpusCCG2007}.%
\footnote{\citet{stanojevicCCGParsingAlgorithm2019} also contains non-trainable parameters (i.e. they are fixed during parser training) that come from pretrained ELMo embeddings \citep{elmo}. But even if ELMo data and parameters are included in the CCG model size it is still many orders of magnitude smaller than Chinchilla both in data and parameter size. Similar numbers to CCG parser also hold for the Benepar constituency parser that uses pretrained BERT embeddings \citep{devlin-etal-2019-bert}.}
However, these counts over-state the parameter space spanned by the interpretable parsing model because they are used only for the disambiguation/predictability of the parse trees and do not participate directly in the complexity metrics used for our analysis. 
Our CCG parsing effort is measured only by the number of parsing actions taken per word, not by their predictability.
In principle, the CCG parser makes one of a finite set of choices per step (shift, reduce, reveal, etc.). 
From that perspective, the size of the CCG model is on the order of $445$ possible choices.%
\footnote{The parser has $425$ different shift transitions, one for each possible lexical category, and $20$ different reduce transitions ($5$ for binary combinators, $2$ for unary type-raising combinators, $12$ for unary type-changing rules and $1$ for revealing operation).}

The upshot of this discussion is twofold. 
First, independent effects of structural~complexity from the CCG parser, in comparison to \chinchilla{}, stand out against the massive difference in complexity between those two models. 
Second, connecting our work with efforts that align brain data directly to the internal states of LLMs is made difficult by the lack of parsimony in those efforts.

\section{Conclusion} \label{sec:conclusion}

We find evidence that CCG captures neural activity above-and-beyond that which correlates with predictability of large language model and parsing steps of context-free grammar parser.
Moreover, we find improved fits when CCG is augmented by the \textsc{reveal}~operation, allowing for more human-like parsing of right-adjunction. 
We observe the strongest effects for parsing~steps in the posterior temporal lobe, around the middle temporal gyrus. This is in close~alignment with recent models for the neural bases of something akin to ``structure-building''.

\bibliography{BIB}
\bibliographystyle{apalike}

\clearpage
\onecolumn
\appendix

\end{document}


\maketitle

\clearpage

\begin{figure}[h!]
    \centering
    \subcaptionbox{$\beta$}{
    \includegraphics[width=\textwidth]{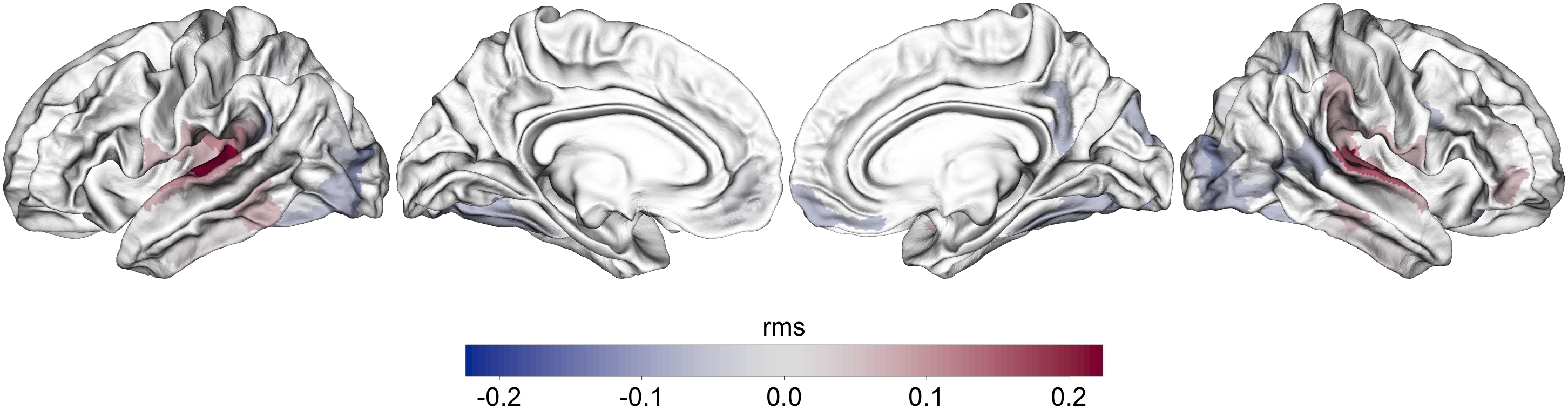}
    }
    \subcaptionbox{$\beta$}{
    \includegraphics[width=\textwidth]{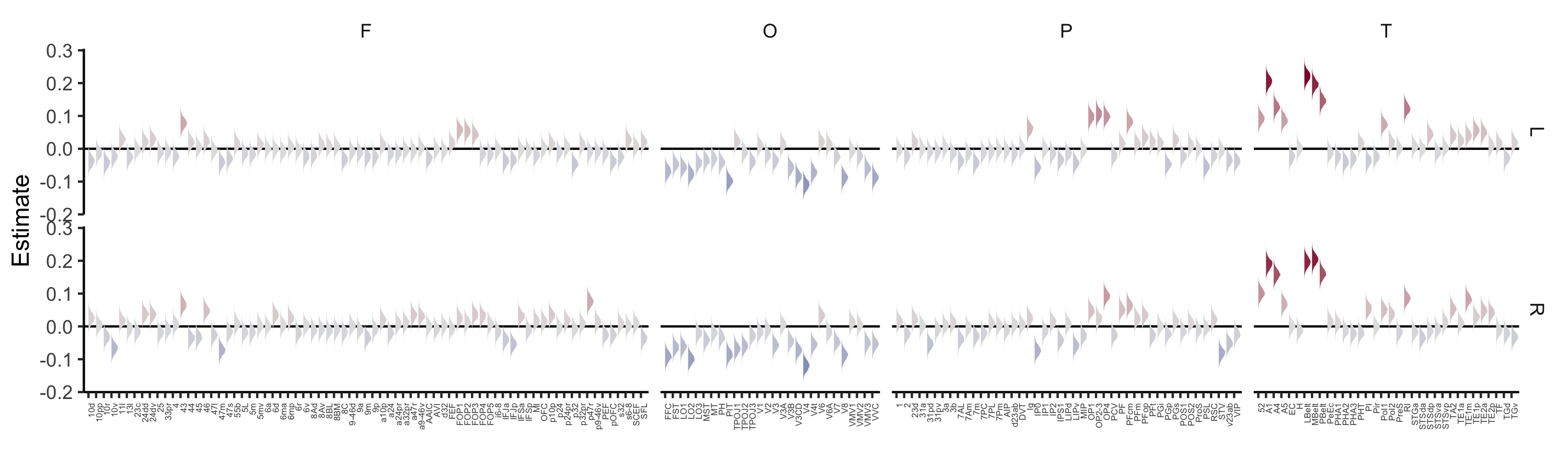}
    }
    \subcaptionbox{\deltarmse}{
    \includegraphics[width=\textwidth]{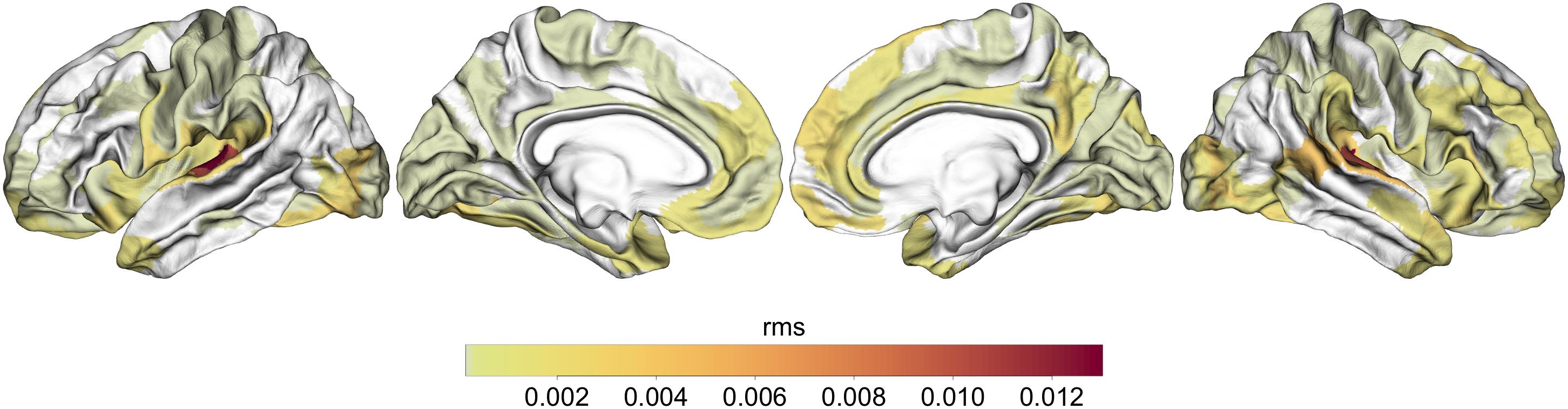}
    }
    \subcaptionbox{\deltarmse}{
    \includegraphics[width=\textwidth]{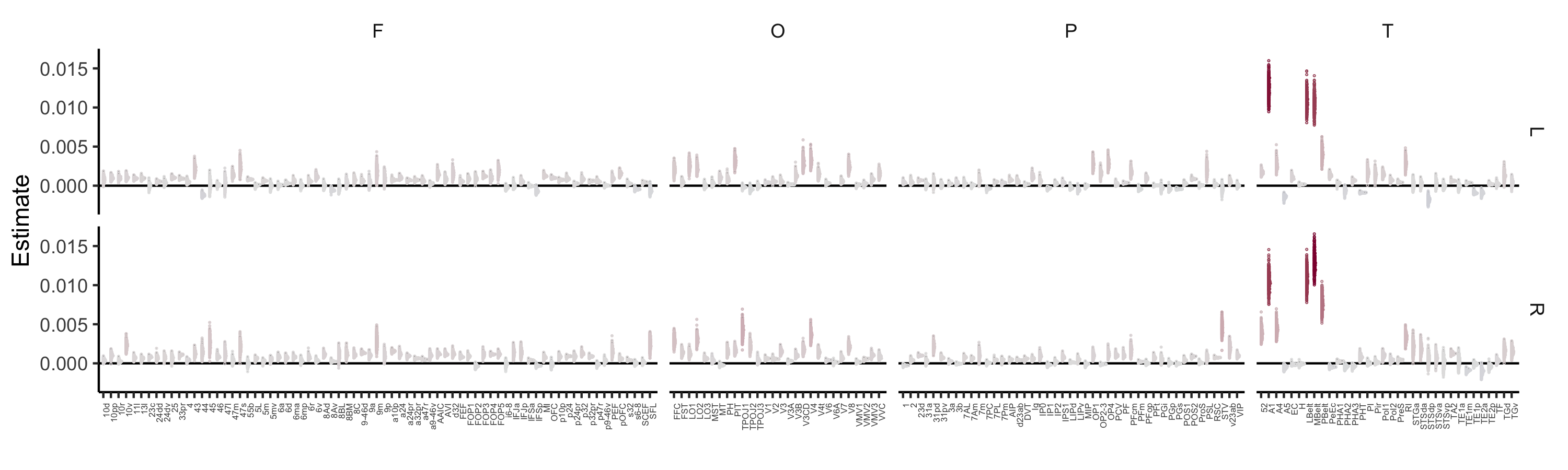}
    }
    \caption{Full results for \rms. 
    (A) Posterior mean of $\beta$ where {\cibeta} excludes zero.
    (B) Posterior distributions of $\beta$ for each region separated by lobe (\underline{F}rontal, \underline{O}ccipital, \underline{P}arietal, \underline{T}emporal) and hemisphere (\underline{L}eft, \underline{R}ight).
    (C) Posterior mean of \deltarmse{} where {\cirmse} excludes zero.
    (D) posterior distributions of \deltarmse{} organized as in (B).
    }
    \label{fig:supp:rms}
\end{figure}

\begin{figure}[h!]
    \centering
    \subcaptionbox{$\beta$}{
    \includegraphics[width=\textwidth]{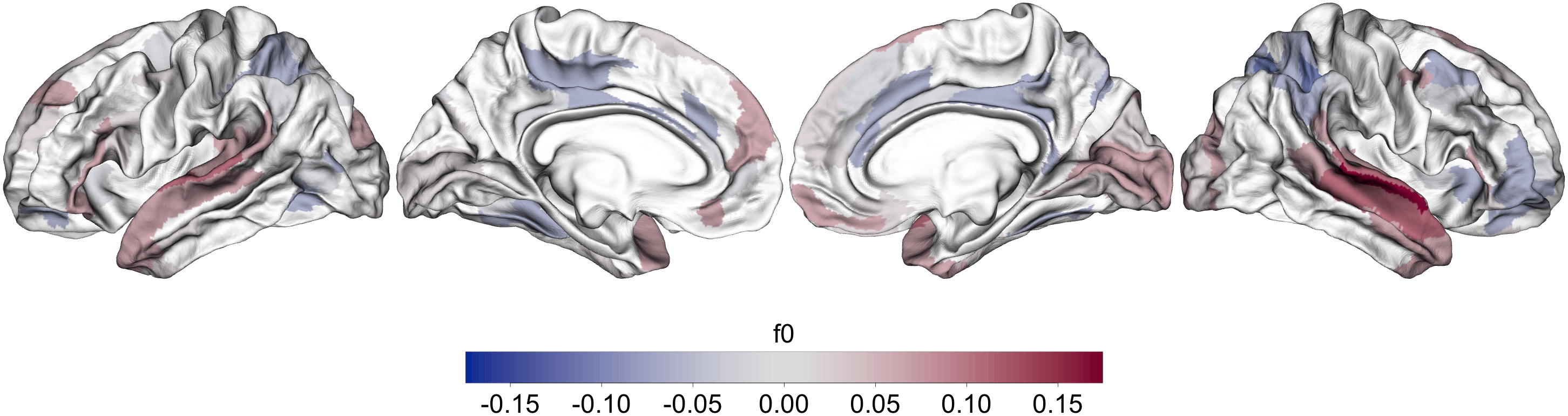}
    }
    \subcaptionbox{$\beta$}{
    \includegraphics[width=\textwidth]{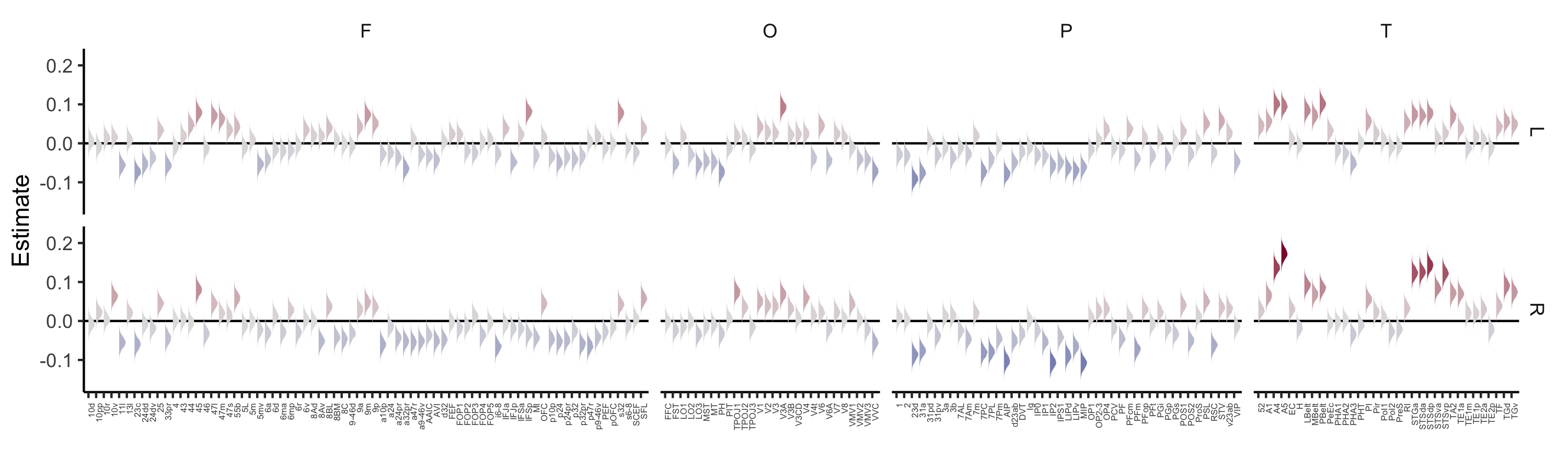}
    }
    \subcaptionbox{\deltarmse}{
    \includegraphics[width=\textwidth]{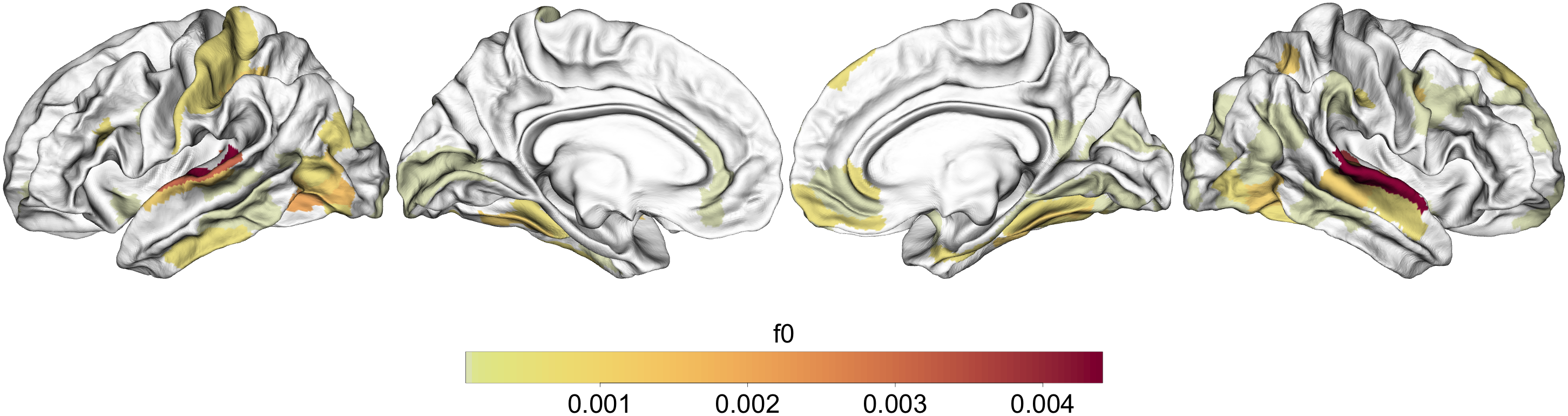}
    }
    \subcaptionbox{\deltarmse}{
    \includegraphics[width=\textwidth]{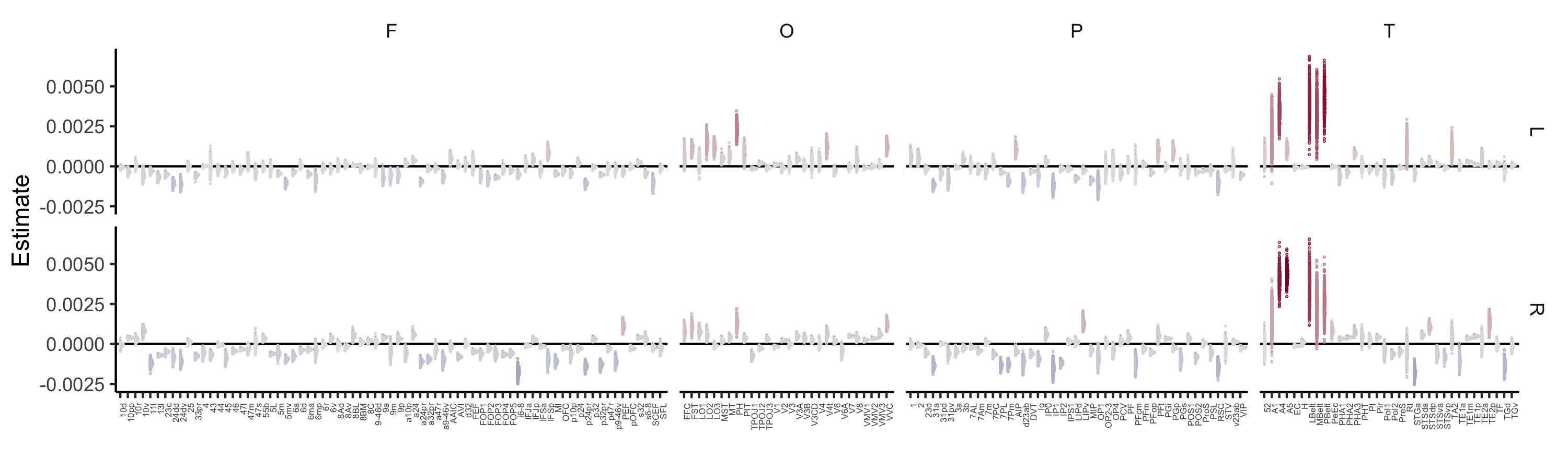}
    }
    \caption{Full results for \fzero.
    See Figure \ref{fig:supp:rms} caption for details.
    }
    \label{fig:supp:f0}
\end{figure}

\begin{figure}[h!]
    \centering
    \subcaptionbox{$\beta$}{
    \includegraphics[width=\textwidth]{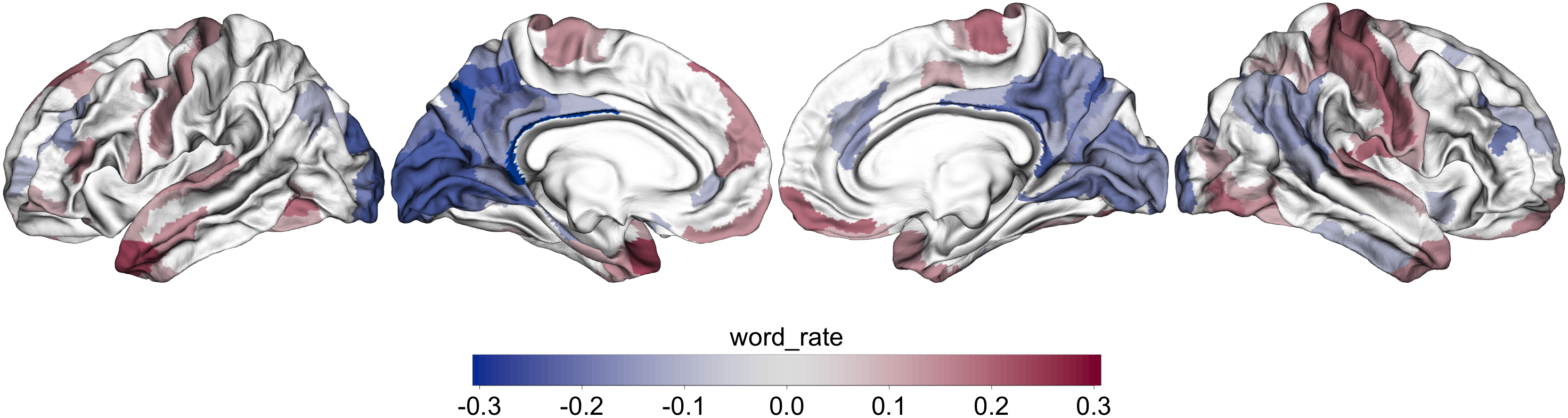}
    }
    \subcaptionbox{$\beta$}{
    \includegraphics[width=\textwidth]{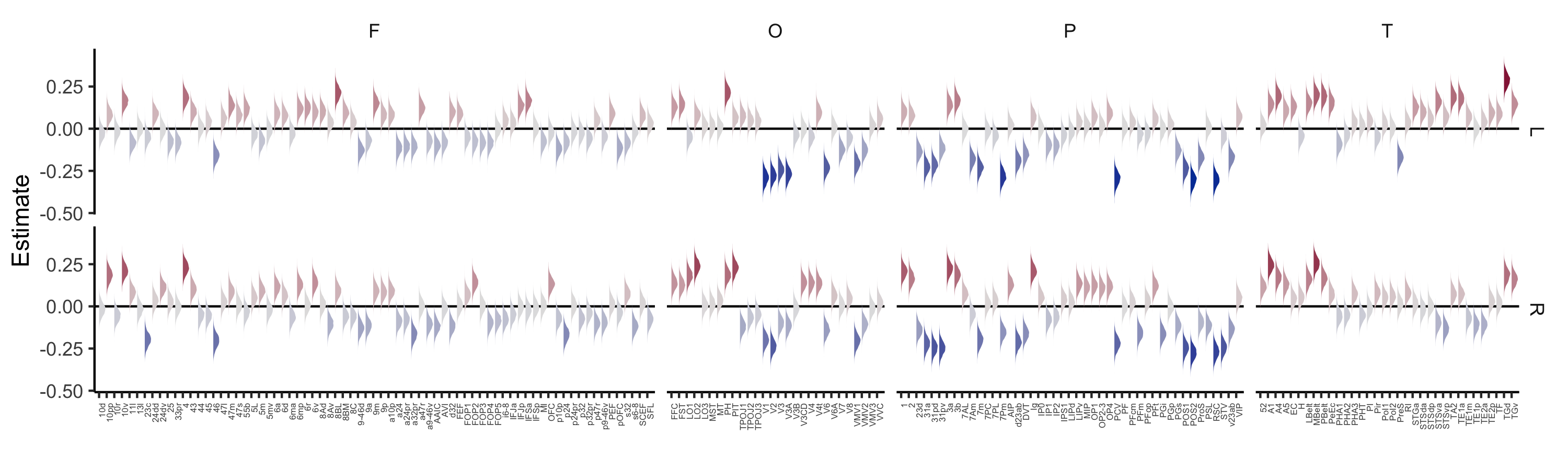}
    }
    \subcaptionbox{\deltarmse}{
    \includegraphics[width=\textwidth]{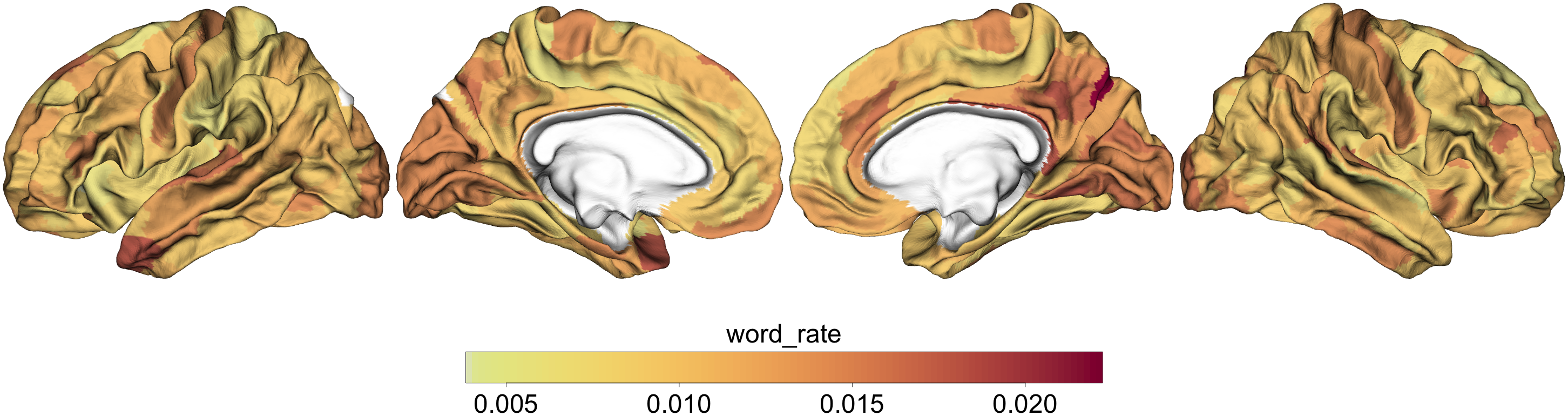}
    }
    \subcaptionbox{\deltarmse}{
    \includegraphics[width=\textwidth]{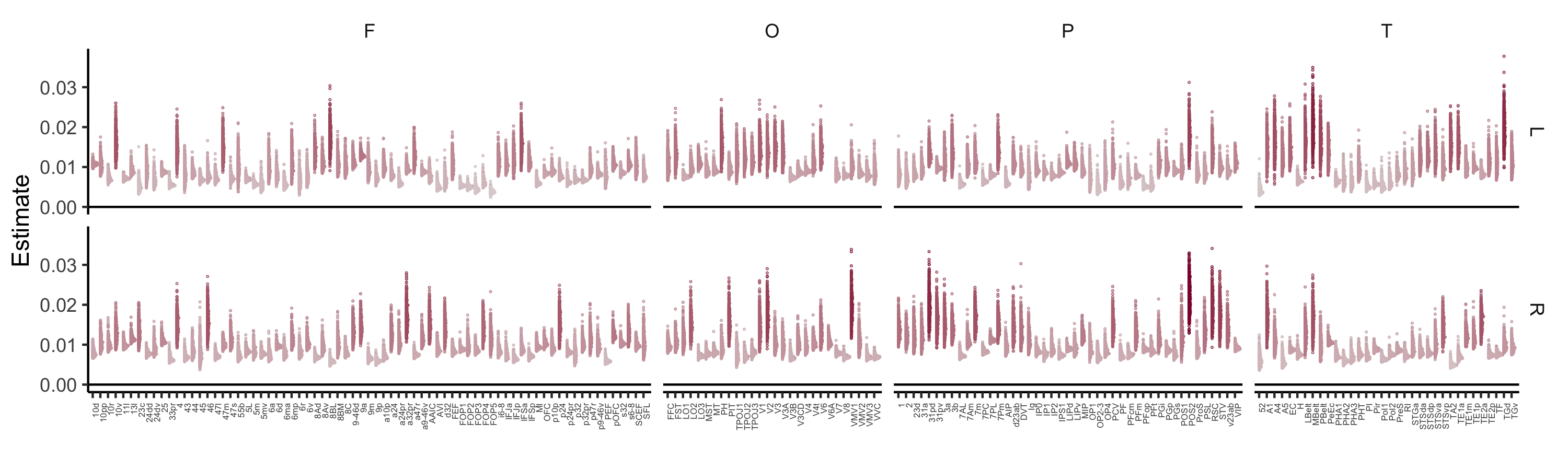}
    }
    \caption{Full results for \wordrate. 
    See Figure \ref{fig:supp:rms} caption for details.
    }
    \label{fig:supp:wordrate}
\end{figure}

\begin{figure}[h!]
    \centering
    \subcaptionbox{$\beta$}{
    \includegraphics[width=\textwidth]{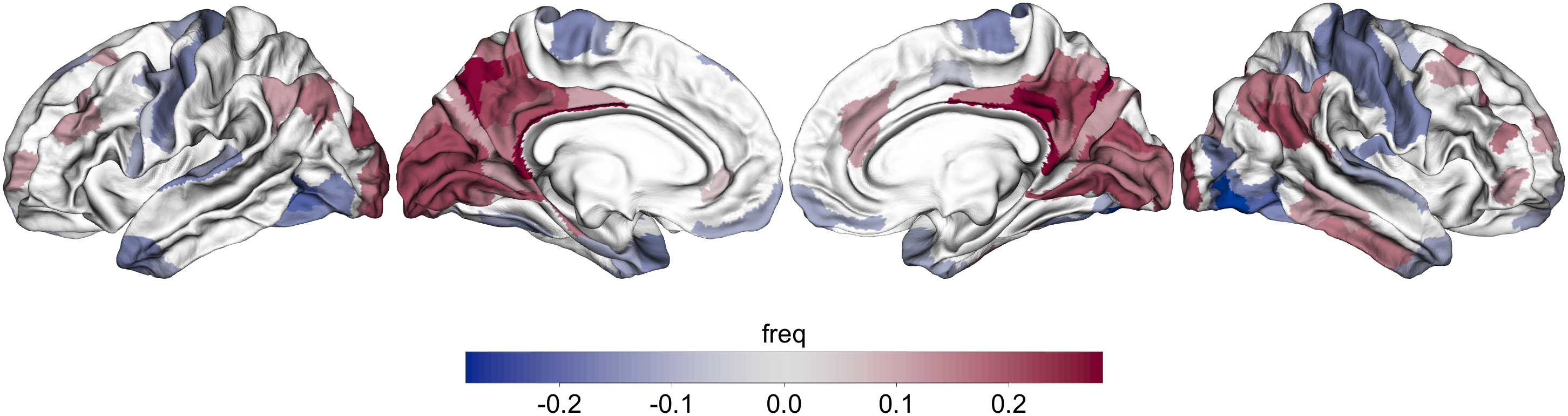}
    }
    \subcaptionbox{$\beta$}{
    \includegraphics[width=\textwidth]{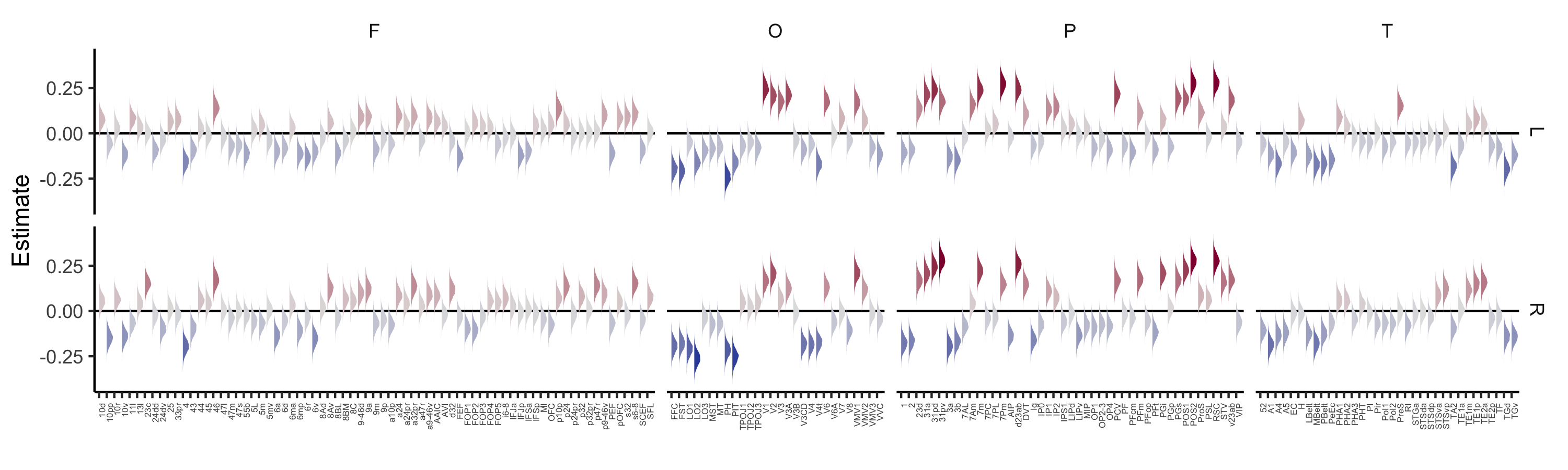}
    }
    \subcaptionbox{\deltarmse}{
    \includegraphics[width=\textwidth]{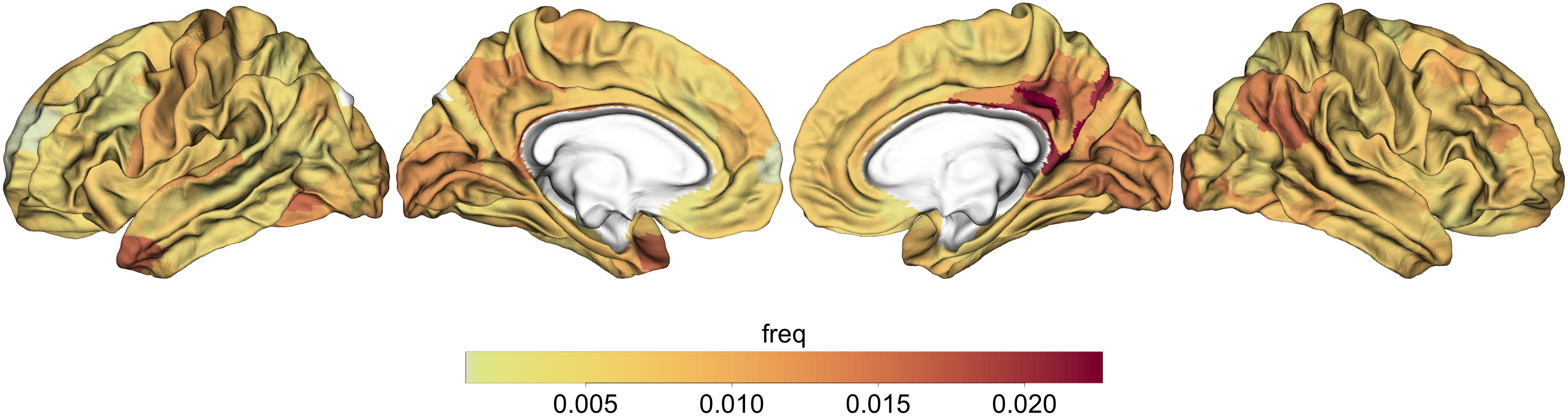}
    }
    \subcaptionbox{\deltarmse}{
    \includegraphics[width=\textwidth]{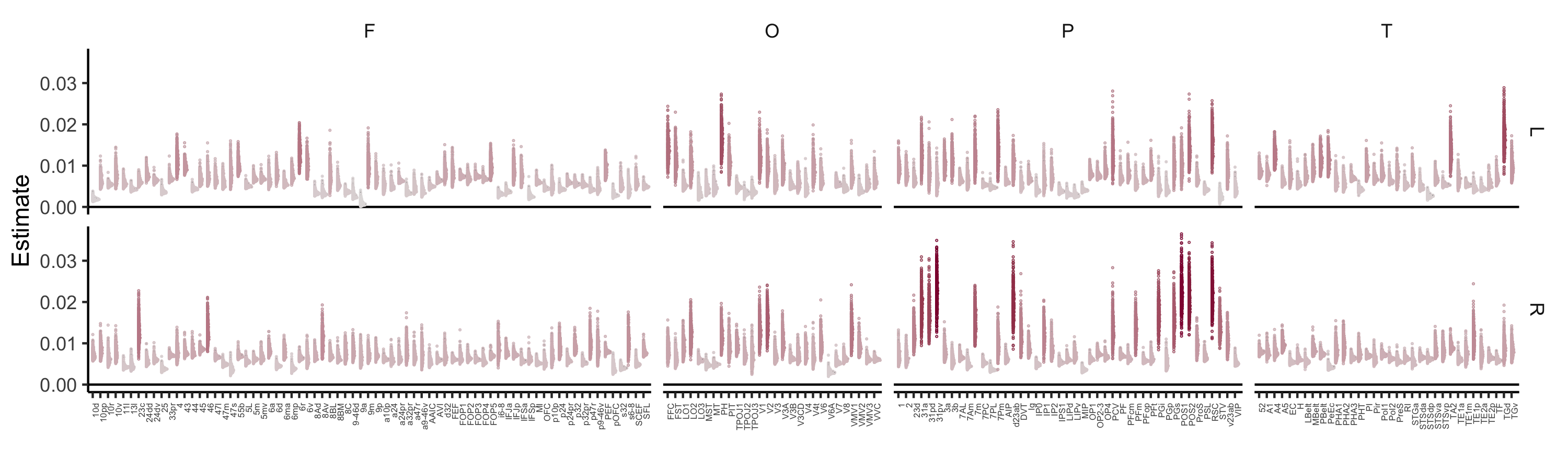}
    }
    \caption{Full results for \freq. 
    See Figure \ref{fig:supp:rms} caption for details.
    }
    \label{fig:supp:freq}
\end{figure}

\begin{figure}[h!]
    \centering
    \subcaptionbox{$\beta$}{
    \includegraphics[width=\textwidth]{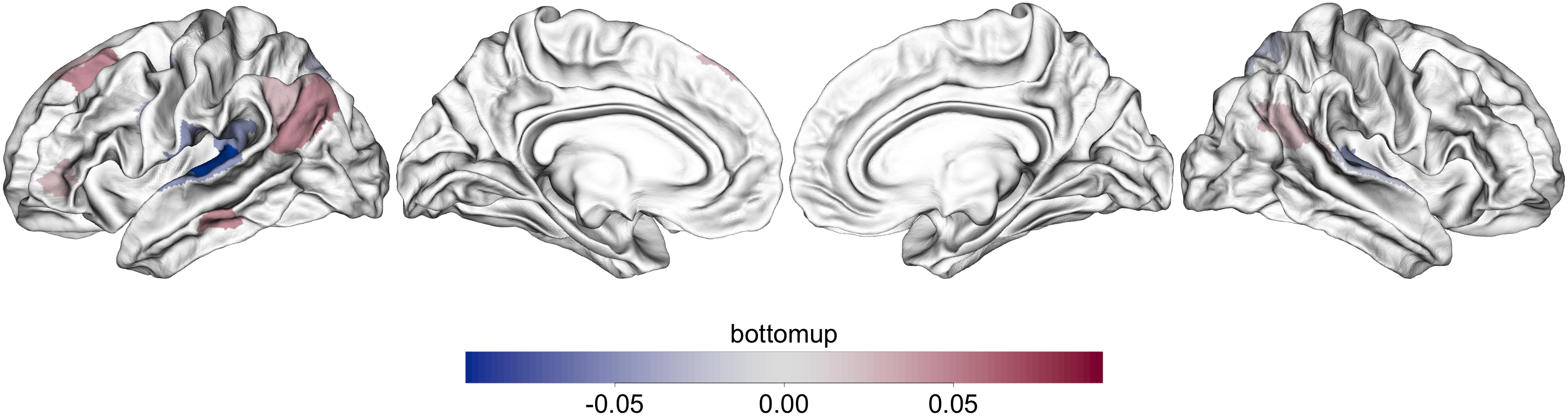}
    }
    \subcaptionbox{$\beta$}{
    \includegraphics[width=\textwidth]{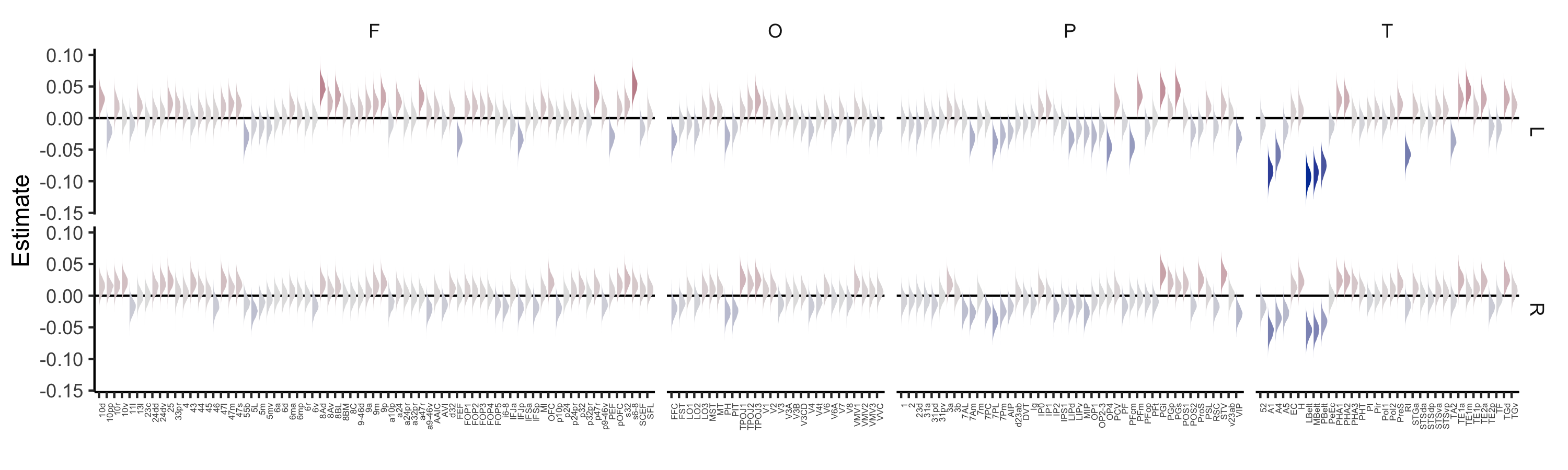}
    }
    \subcaptionbox{\deltarmse}{
    \includegraphics[width=\textwidth]{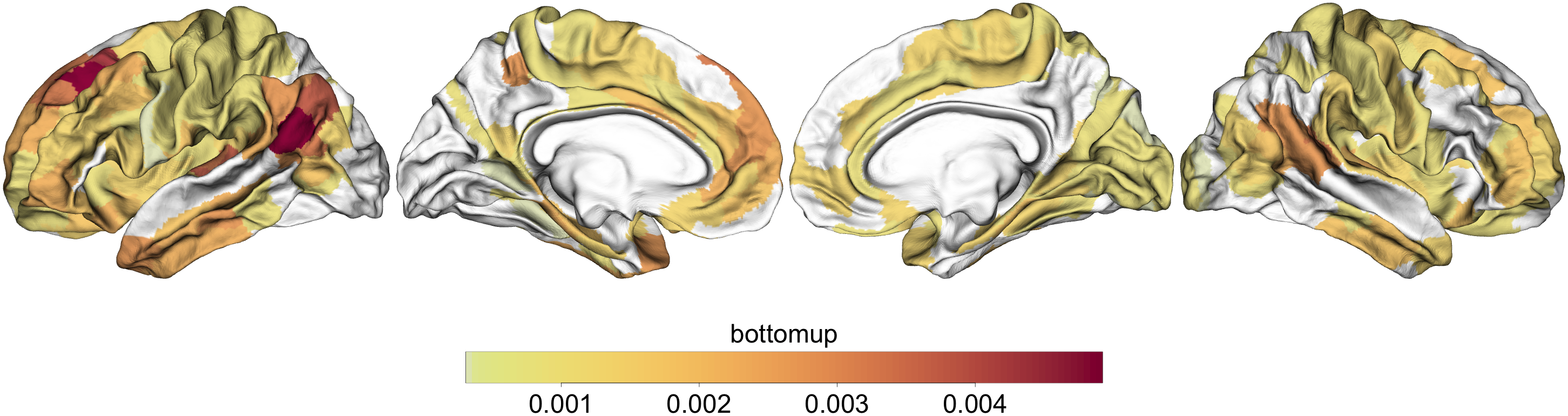}
    }
    \subcaptionbox{\deltarmse}{
    \includegraphics[width=\textwidth]{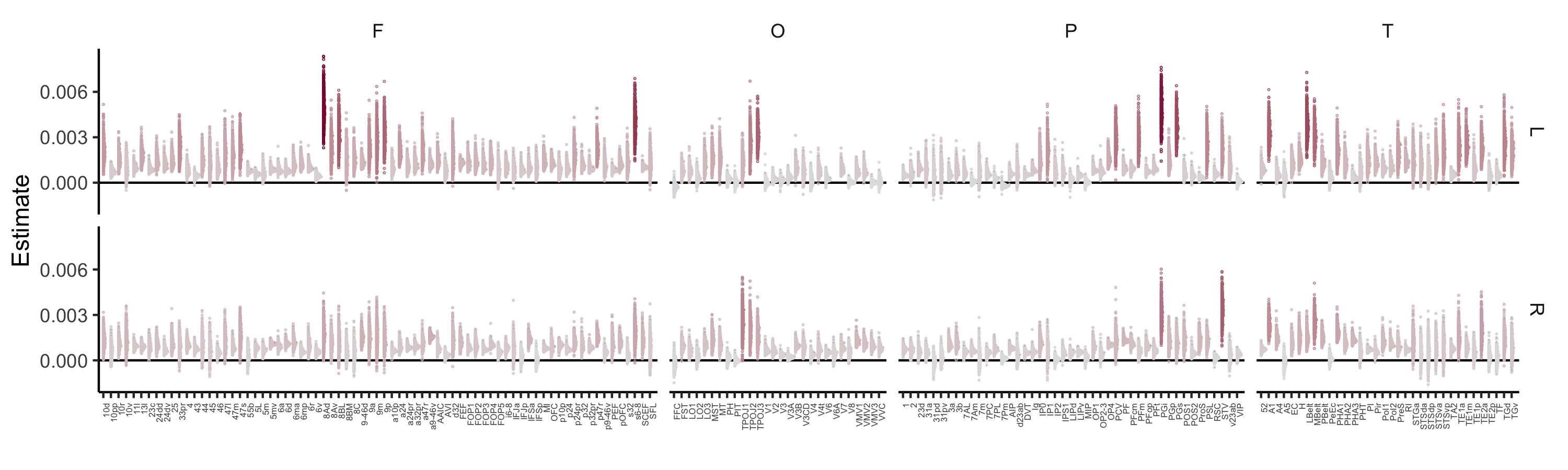}
    }
    \caption{Full results for \bottomup. 
    See Figure \ref{fig:supp:rms} caption for details.
    }
    \label{fig:supp:bottomup}
\end{figure}

\begin{figure}[h!]
    \centering
    \subcaptionbox{$\beta$}{
    \includegraphics[width=\textwidth]{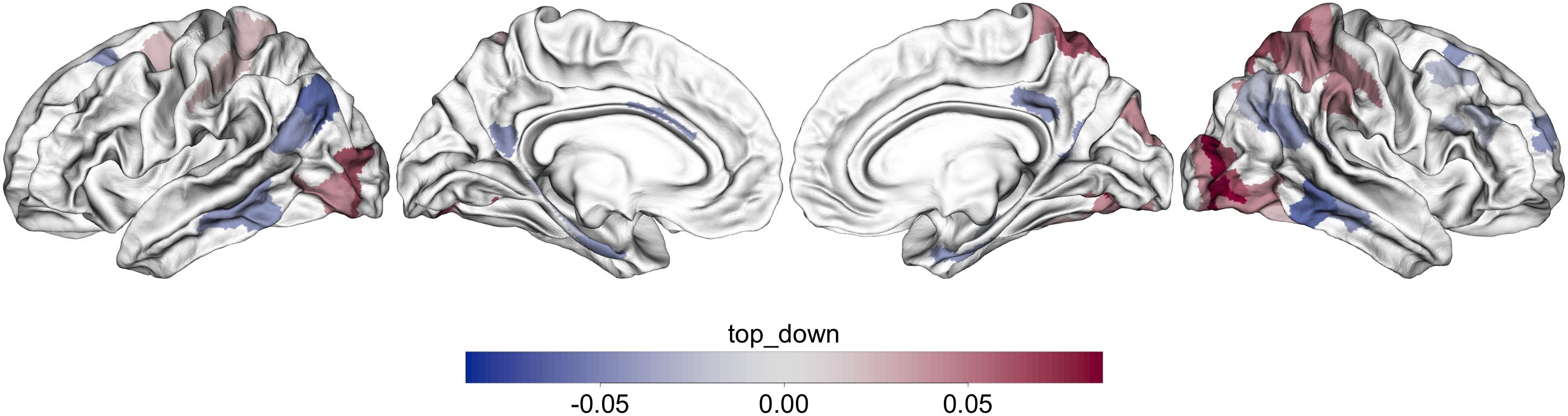}
    }
    \subcaptionbox{$\beta$}{
    \includegraphics[width=\textwidth]{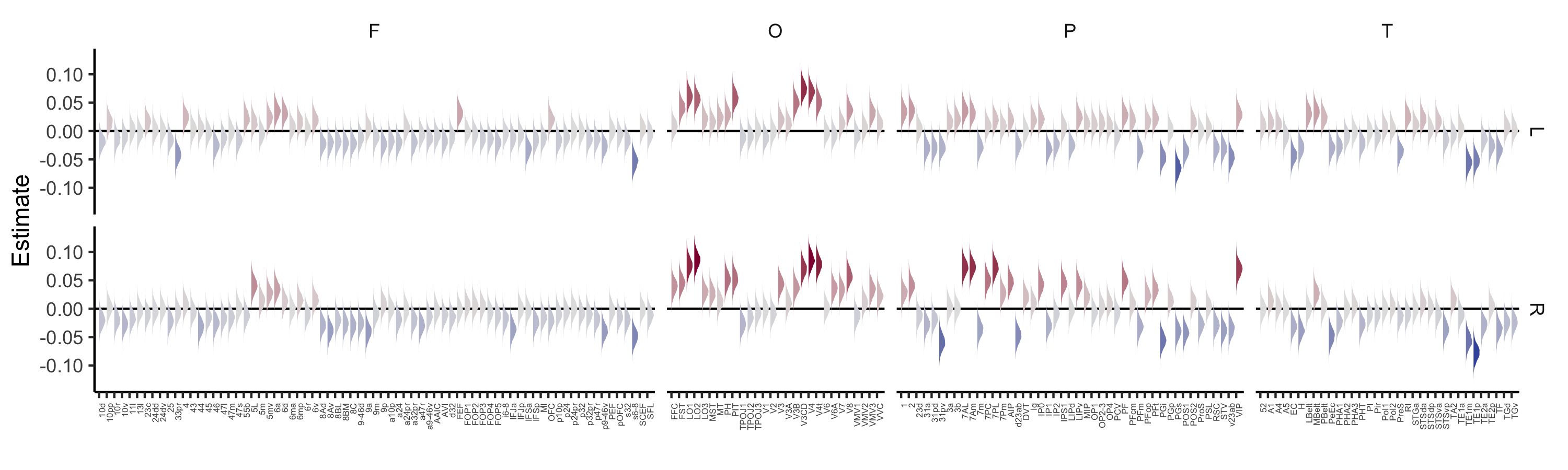}
    }
    \subcaptionbox{\deltarmse}{
    \includegraphics[width=\textwidth]{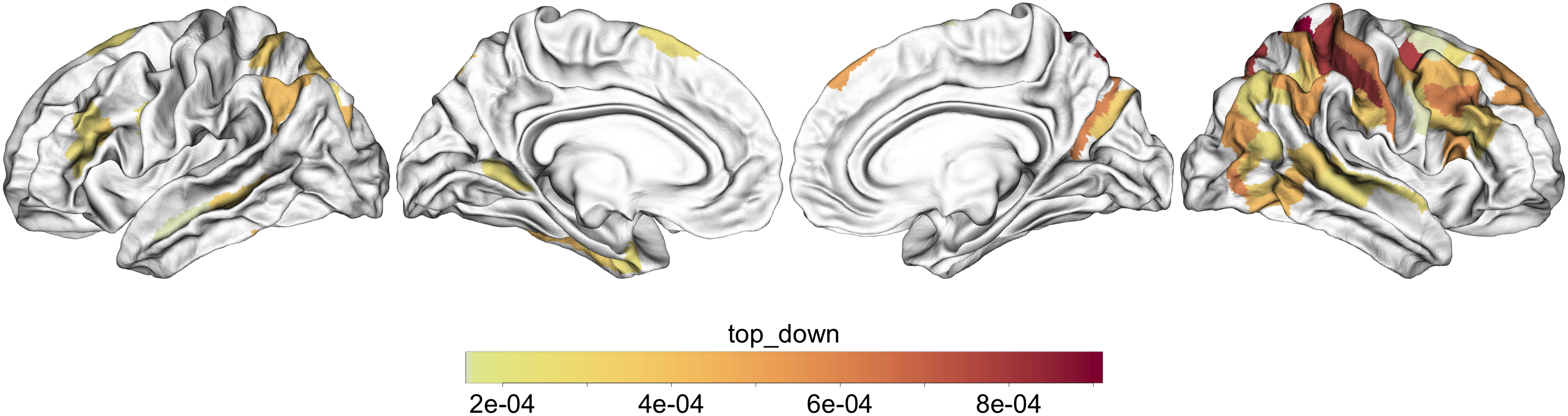}
    }
    \subcaptionbox{\deltarmse}{
    \includegraphics[width=\textwidth]{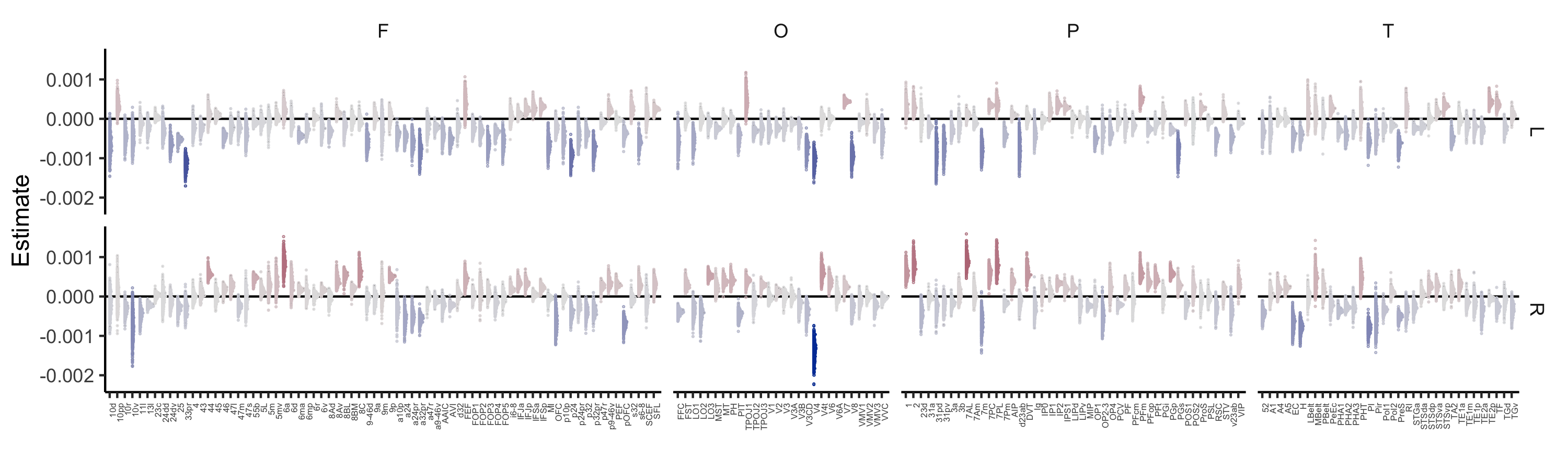}
    }
    \caption{Full results for \topdown. 
    See Figure \ref{fig:supp:rms} caption for details.
    }
    \label{fig:supp:topdown}
\end{figure}

\begin{figure}[h!]
    \centering
    \subcaptionbox{$\beta$}{
    \includegraphics[width=\textwidth]{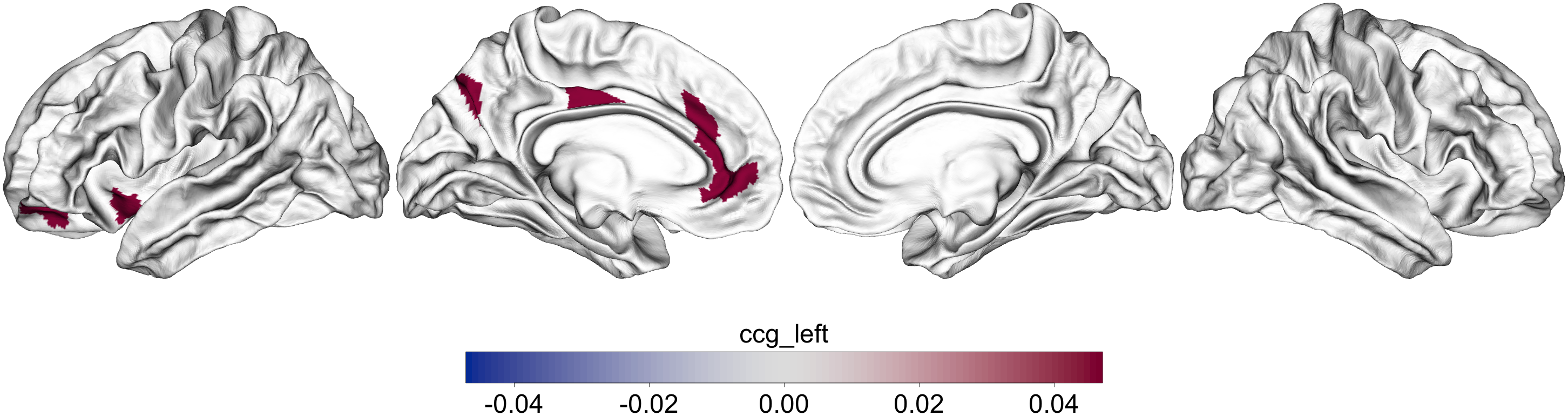}
    }
    \subcaptionbox{$\beta$}{
    \includegraphics[width=\textwidth]{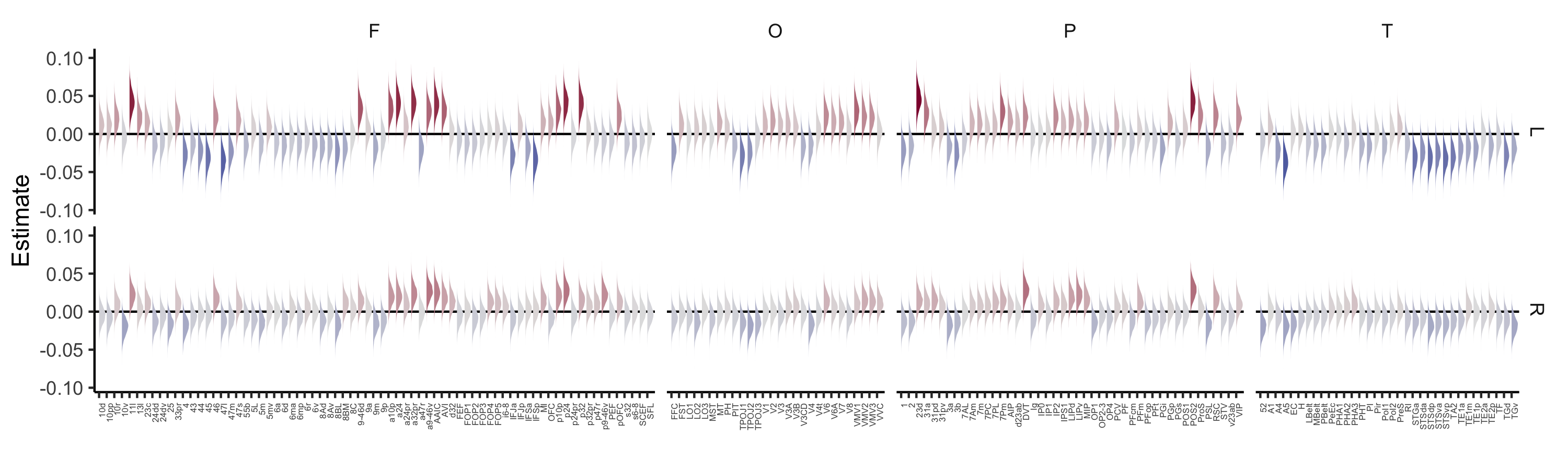}
    }
    \subcaptionbox{\deltarmse}{
    \includegraphics[width=\textwidth]{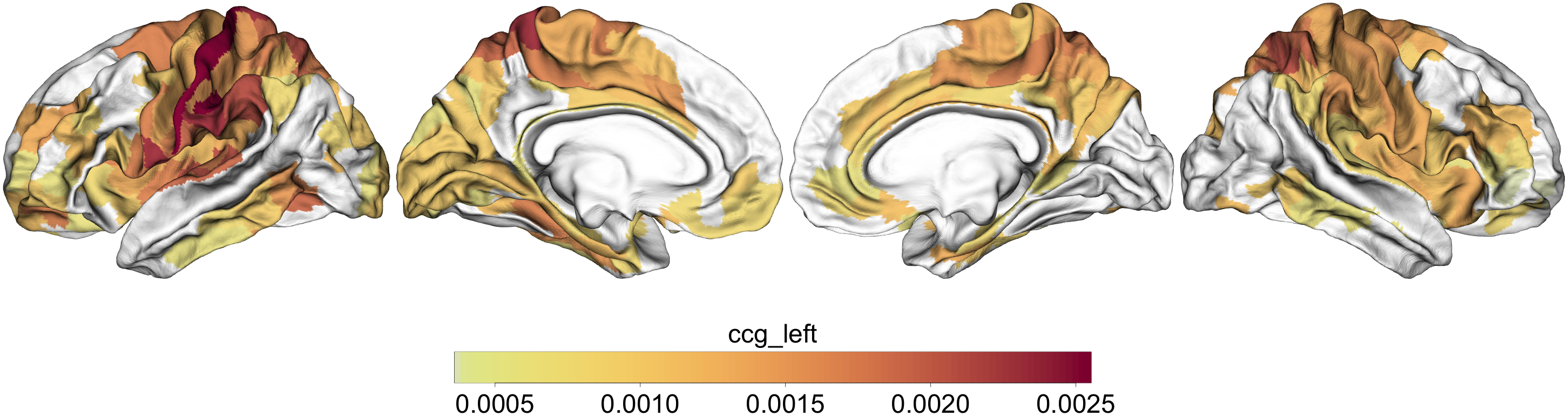}
    }
    \subcaptionbox{\deltarmse}{
    \includegraphics[width=\textwidth]{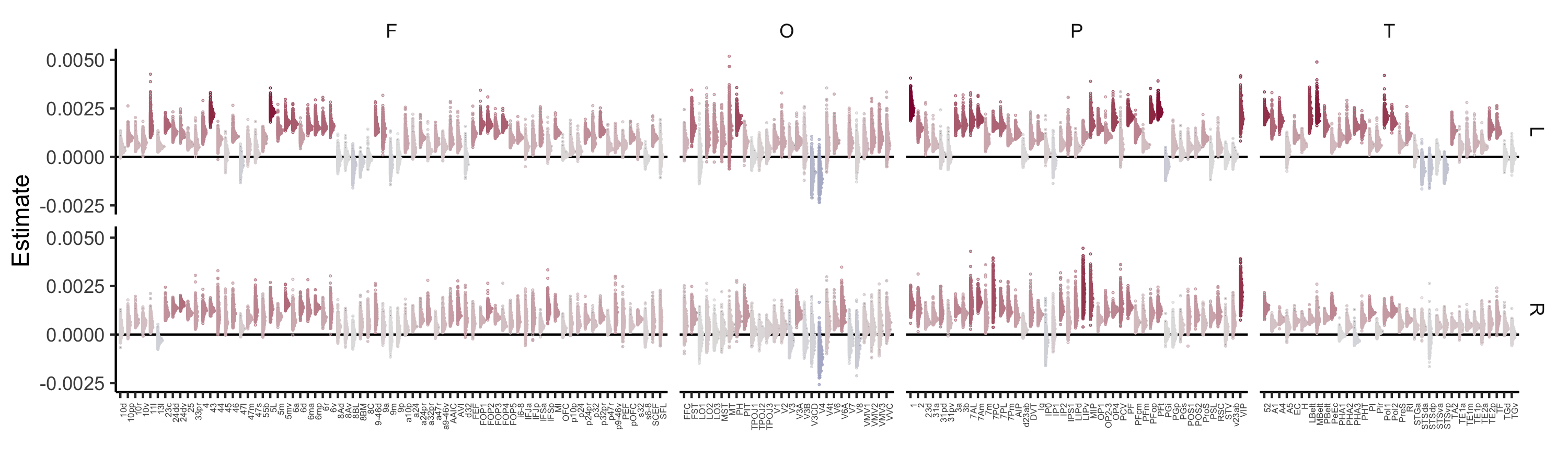}
    }
    \caption{Full results for \ccgleft. 
    See Figure \ref{fig:supp:rms} caption for details.
    }
    \label{fig:supp:ccgleft}
\end{figure}

\begin{figure}[h!]
    \centering
    \subcaptionbox{$\beta$}{
    \includegraphics[width=\textwidth]{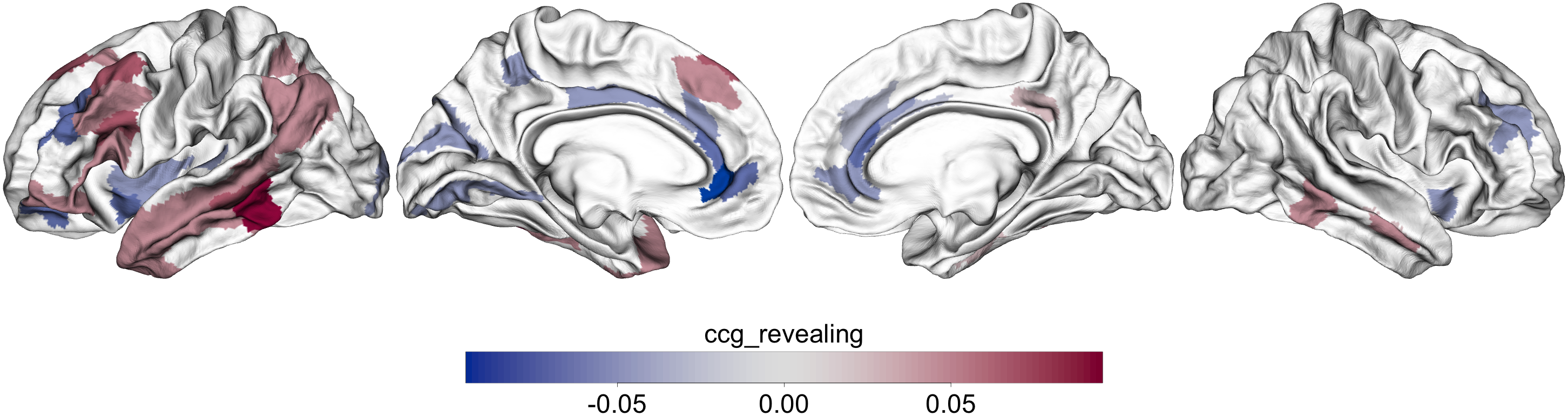}
    }
    \subcaptionbox{$\beta$}{
    \includegraphics[width=\textwidth]{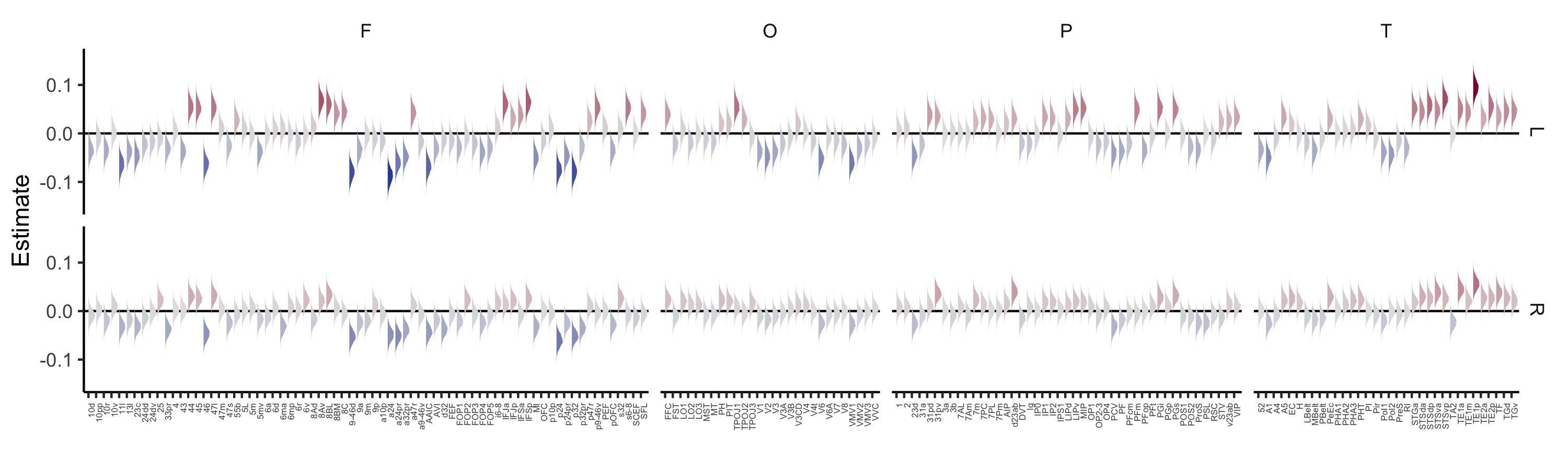}
    }
    \subcaptionbox{\deltarmse}{
    \includegraphics[width=\textwidth]{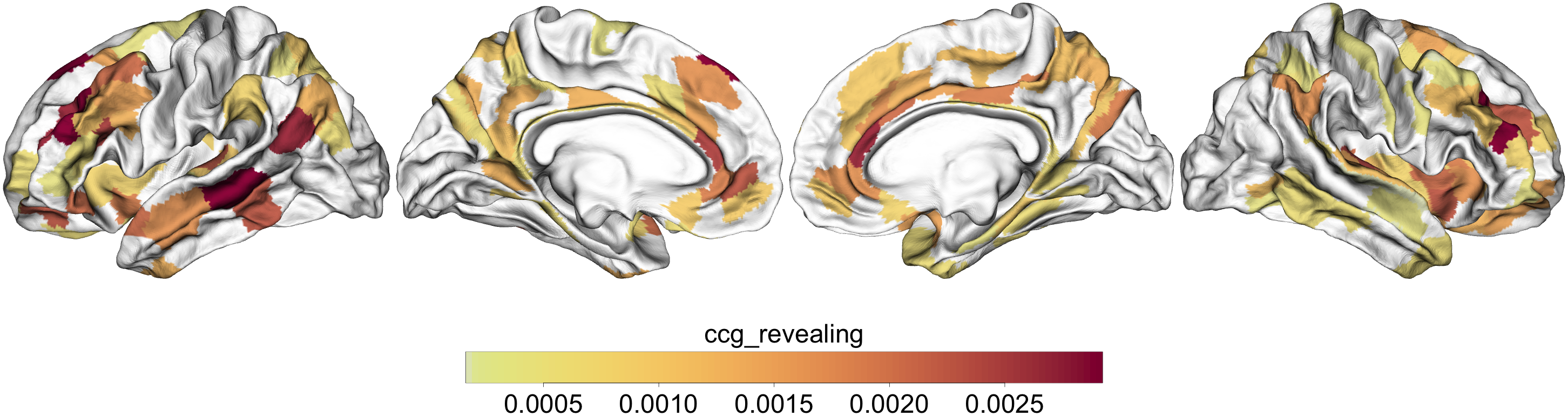}
    }
    \subcaptionbox{\deltarmse}{
    \includegraphics[width=\textwidth]{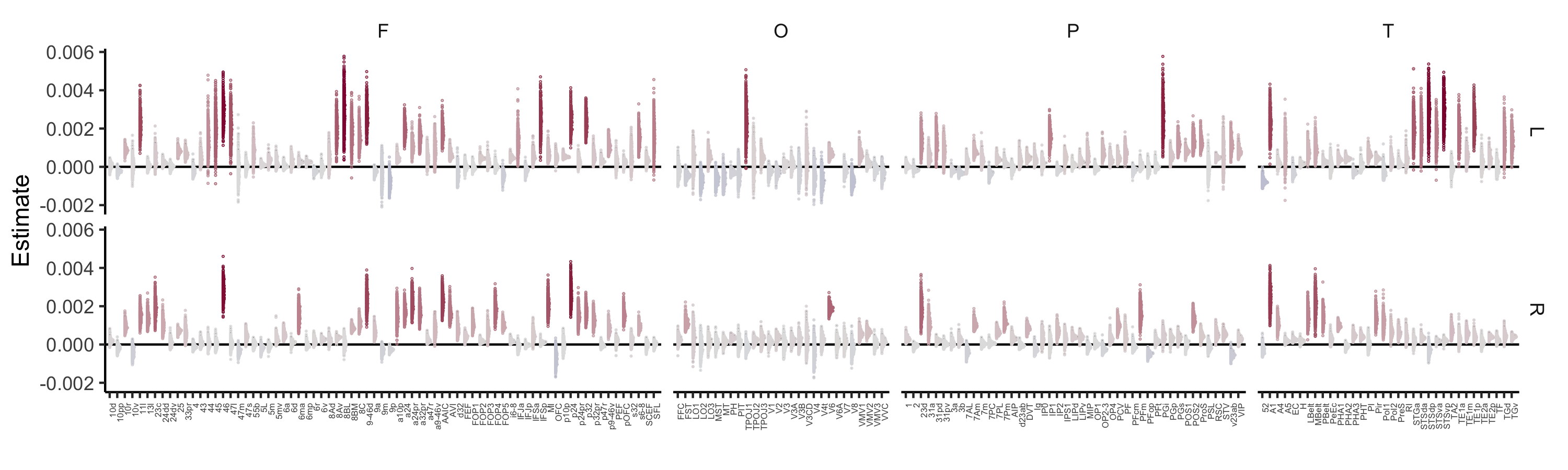}
    }
    \caption{Full results for \ccgreveal. 
    See Figure \ref{fig:supp:rms} caption for details.
    }
    \label{fig:supp:ccgrevealing}
\end{figure}

\begin{figure}[h!]
    \centering
    \subcaptionbox{$\beta$}{
    \includegraphics[width=\textwidth]{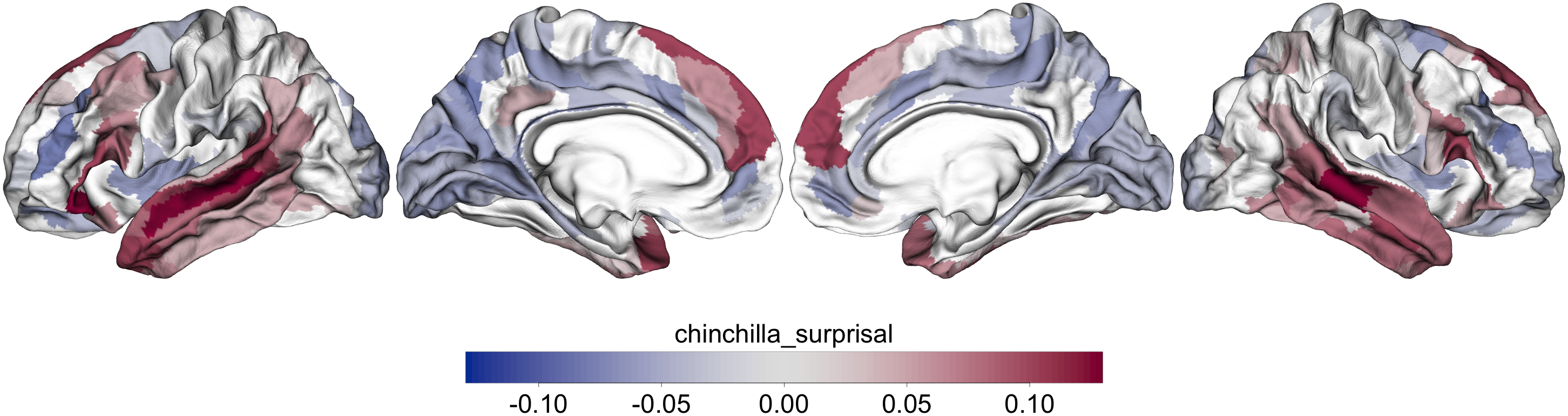}
    }
    \subcaptionbox{$\beta$}{
    \includegraphics[width=\textwidth]{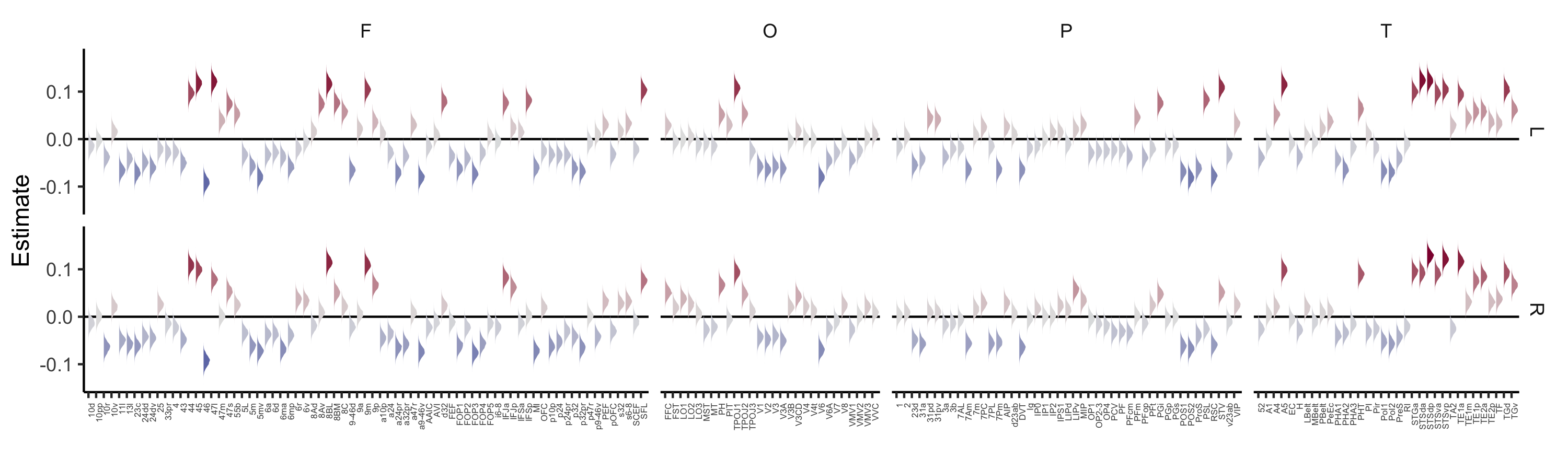}
    }
    \subcaptionbox{\deltarmse}{
    \includegraphics[width=\textwidth]{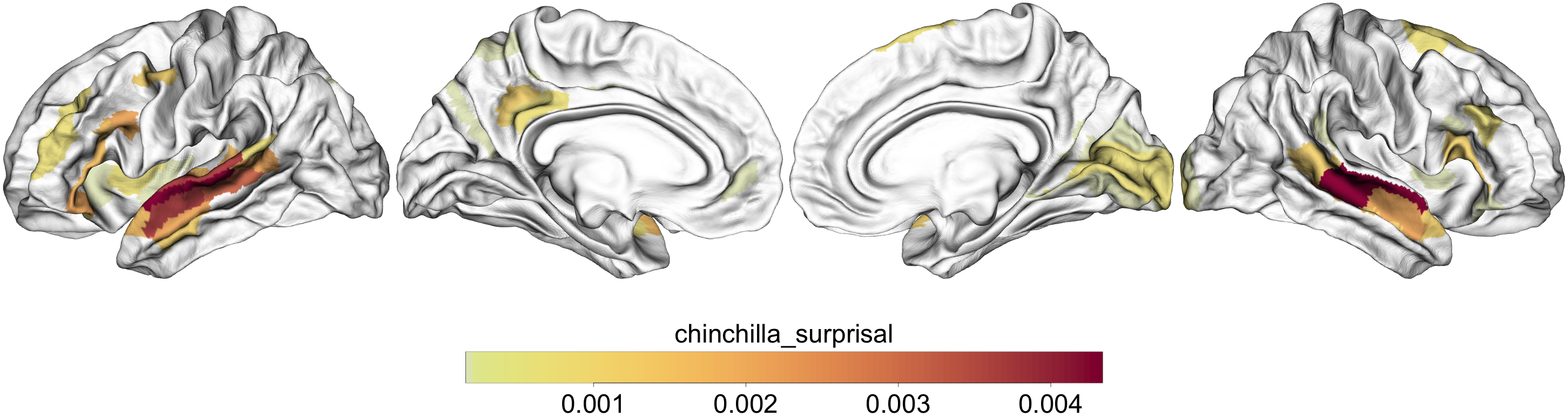}
    }
    \subcaptionbox{\deltarmse}{
    \includegraphics[width=\textwidth]{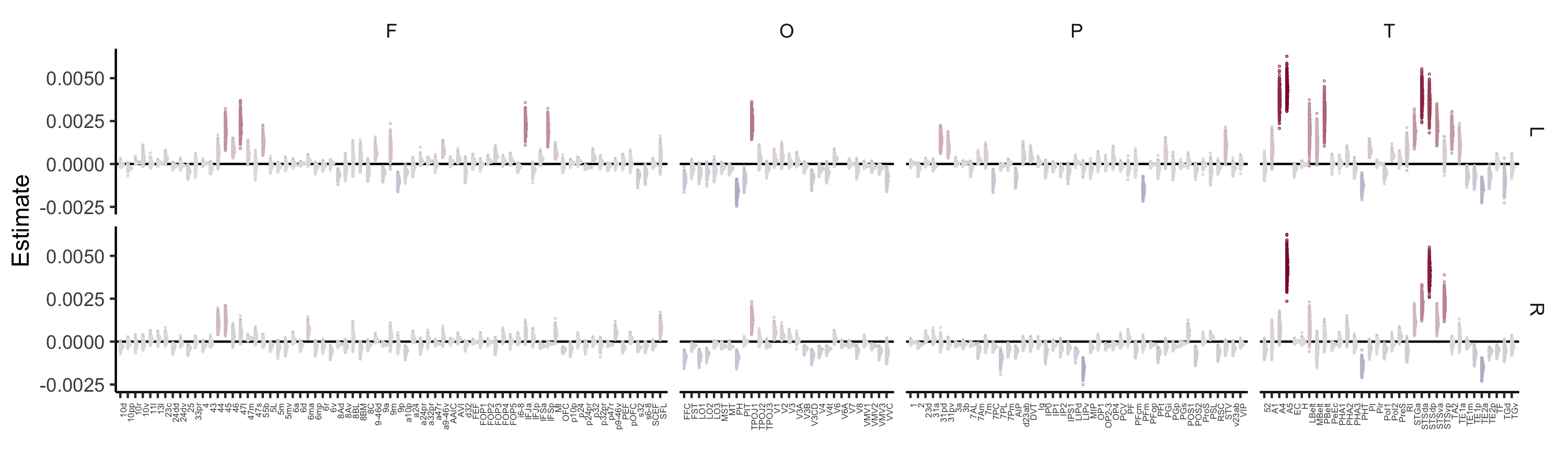}
    }
    \caption{Full results for \surprisal. 
    See Figure \ref{fig:supp:rms} caption for details.
    }
    \label{fig:supp:surprisal}
\end{figure}